\documentclass{article} 
\usepackage{PRIMEarxiv}
{}
{}
{}
\usepackage[utf8]{inputenc}
\usepackage{multirow}
\usepackage{array}
\usepackage{booktabs}
\usepackage{booktabs,tabularx}
\usepackage{mathtools}
\usepackage{amsmath,amssymb,amsfonts}
\usepackage{algorithmic}
\usepackage{graphicx}
\usepackage{textcomp}
\usepackage{xcolor}
\usepackage{algorithmic,comment}
\usepackage[ruled,vlined]{algorithm2e}
\usepackage[T1]{fontenc}
\usepackage[utf8]{inputenc}
\usepackage{verbatim}
\usepackage{hyperref}
\usepackage{soul} 
\usepackage[hyphenbreaks]{breakurl}
\usepackage{adjustbox}
\usepackage{booktabs}
\usepackage{tikz}
\usepackage{subcaption}
\usepackage{caption}
\usepackage{float}

\usepackage[symbol]{footmisc}
\newcommand{\blr}[0]{\color{black}}
\newcommand{\ar}[1]{\textcolor{black}{#1}}

\newcommand{\xc}[1]{\textcolor{olive}{#1}}

\newcommand{\arr}[1]{\textcolor{red}{\textit{Razi: #1}}}

\usepackage{enumerate}
\usepackage{colortbl}
\usepackage{array}
\usepackage{booktabs}

\makeatletter
\def\UrlAlphabet{%
      \do\a\do\b\do\c\do\d\do\e\do\f\do\g\do\h\do\i\do\j%
      \do\k\do\l\do\m\do\n\do\o\do\p\do\q\do\r\do\s\do\t%
      \do\u\do\v\do\w\do\x\do\y\do\z\do\A\do\B\do\C\do\D%
      \do\E\do\F\do\G\do\H\do\I\do\J\do\K\do\L\do\M\do\N%
      \do\O\do\P\do\Q\do\R\do\S\do\T\do\U\do\V\do\W\do\X%
      \do\Y\do\Z}
\def\UrlDigits{\do\1\do\2\do\3\do\4\do\5\do\6\do\7\do\8\do\9\do\0}
\g@addto@macro{\UrlBreaks}{\UrlOrds}
\g@addto@macro{\UrlBreaks}{\UrlAlphabet}
\g@addto@macro{\UrlBreaks}{\UrlDigits}
\makeatother

\def \myVer{1}
\def \newVer{1}
\def \oldVer{2}

\def\myVer{1}

\begin{document}

\title{Deep Learning Serves Traffic Safety Analysis: A Forward-looking Review}

\author{
   Abofazl Razi $^*$\footnotemark[2], Xiwen Chen\footnotemark[2], Hao Wang,  \\
  School of Computing \\
  Clemson University \\
  \And
    Huayu Li\\
    Department of Electrical and Computer Engineering\\
    University of Arizona\\
  \And
    Brendan Russo\\
    Department of Civil Engineering\\
    Northern Arizona University\\
    \And
    Yan Chen \\
    The Polytechnic School\\
    Arizona State University
    \And
    Hongbin Yu\\
    School of Electrical, Computer and Energy Engineering\\
     Arizona State University
}



\maketitle
\begin{abstract}
This paper explores Deep Learning (DL) methods that are used or have the potential to be used for traffic video analysis, emphasizing driving safety for both Autonomous Vehicles (AVs) and human-operated vehicles. We present a typical processing pipeline, which can be used to understand and interpret traffic videos by extracting operational safety metrics and providing general hints and guidelines to improve traffic safety. This processing framework includes several steps, including video enhancement, video stabilization, semantic and incident segmentation, object detection and classification, trajectory extraction, speed estimation, event analysis, modeling and anomaly detection. Our main goal is to guide traffic analysts to develop their own custom-built processing frameworks by selecting the best choices for each step and offering new designs for the lacking modules by providing a comparative analysis of the most successful conventional and DL-based algorithms proposed for each step. We also review existing open-source tools and public datasets that can help train DL models.  To be more specific, we review exemplary traffic problems and mentioned requires steps for each problem. Besides, we investigate connections to the closely related research areas of drivers' cognition evaluation, Crowd-sourcing-based monitoring systems, Edge Computing in roadside infrastructures, Automated Driving Systems (ADS)-equipped vehicles, and highlight the missing gaps. Finally, we review commercial implementations of traffic monitoring systems, their future outlook, and open problems and remaining challenges for widespread use of such systems\footnotemark[3].

\end{abstract}

\footnotetext[1]{Corresponding author. email: arazi@clemson.edu}
\footnotetext[2]{These authors contributed equally to this work.}
\footnotetext[3]{This material is based upon the work supported by the National Science Foundation under Grant No. 2008784 and the Arizona Commerce Authority under the Institute of Automated Mobility (IAM) project.}

\section{Introduction}  \label{sec:intro}
Despite recent advances in vehicles' operational safety features, computer-assisted control, and technology-based traffic management systems, traffic safety remains one of the main challenges in today's life. 
Every year, traffic crashes account for 20-50 million causalities and 1.35 million fatalities worldwide, making it one of the top-10 causes of death. Indeed, traffic crashes is the leading cause of death for people aged 5 to 29 \cite{accidentKiller}.

\begin{figure*}
\begin{center}
\centerline{\includegraphics[width=1\textwidth]{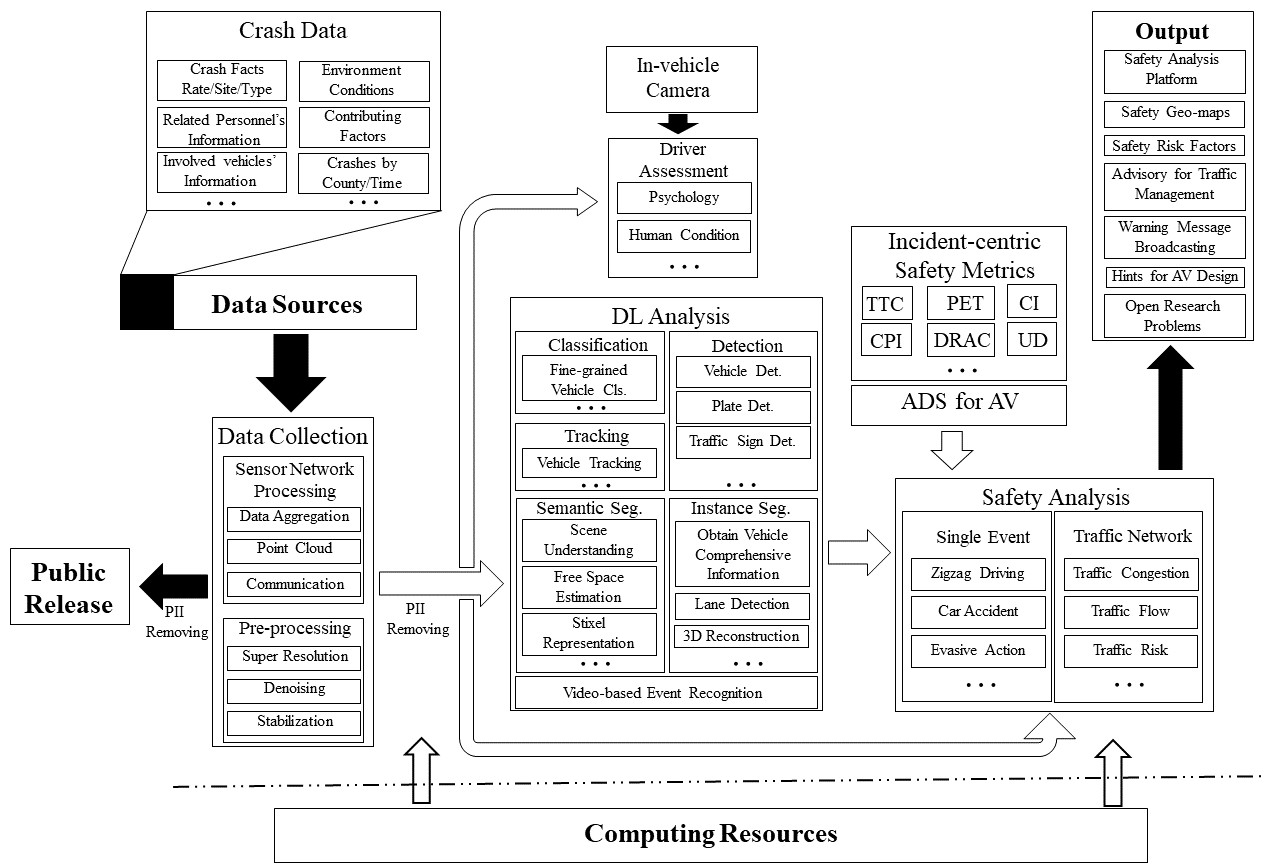}}
\caption{A framework of Computer Vision (CV)-based traffic safety analysis pipeline.}  
\label{fig:network1}
\end{center}
\end{figure*}
\begin{figure*}
\begin{center}
\centerline{\includegraphics[width=1\textwidth]{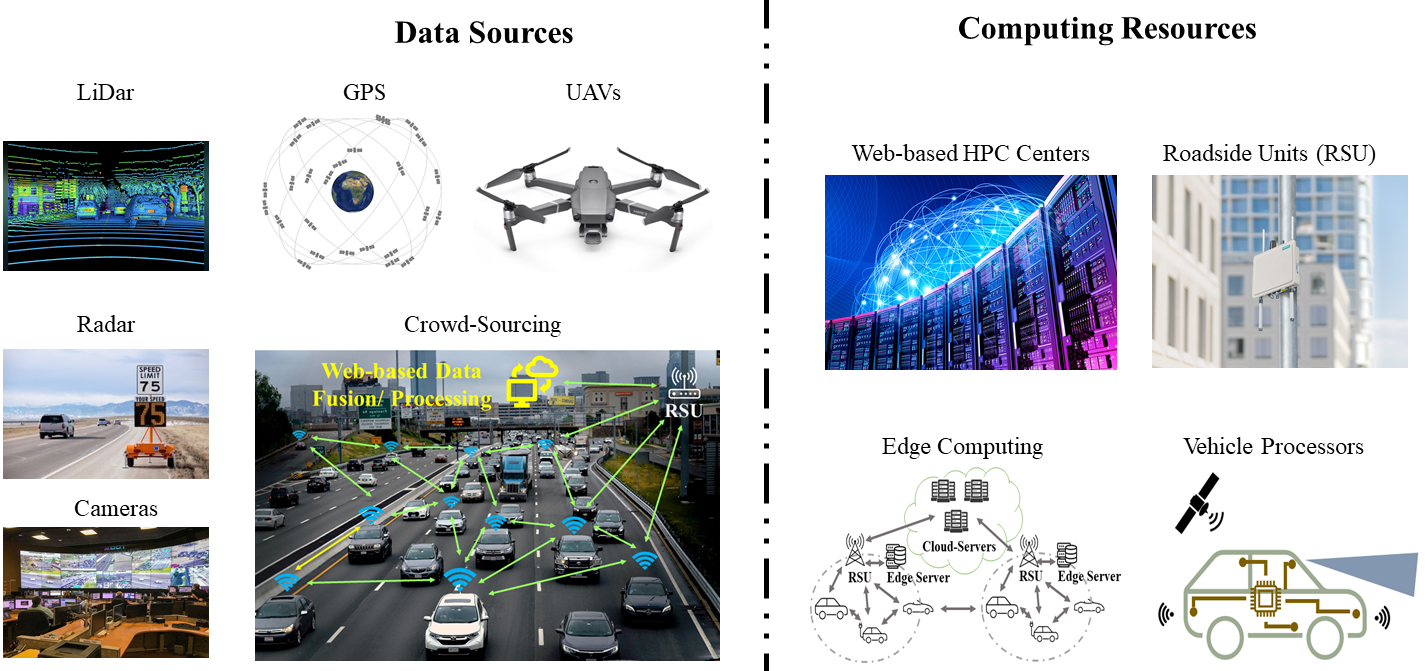}}
\caption{The data sources and computing resources used in video-based traffic analysis.}
\label{fig:network2}
\end{center}
\end{figure*}

Using computational intelligence and computer tools for enhancing traffic safety has gained a lot of attention in recent years. Mainstream technological trends include (i) implementing vehicle safety features such as forward collision warning, blind-spot detection, lane departure warning, backup camera, and autonomous emergency braking \cite{state_farm}, (ii) simulation-based road infrastructure design such as Site3D, RoadEng \cite{dinita_2019}, and OpenRoads Designer \cite{software_solutions}, and (iii) intelligent traffic flow management systems such as Global Positioning System (GPS)-based navigation tools. An example of the last category is  Google's road-user interpretive software that can infer the common road behavior of other drivers that allows Engine Control Units (ECUs) to make better route decisions \cite{kirkland_2019}. 



\ar{Many car manufacturers have developed Artificial Intelligence (AI)-based safety features \cite{kirkland_2019}. Tesla, Audi, and BWM have already developed sensor-based and vision-based perception systems to help the drivers judge road conditions more accurately and use partial autonomy \cite{how_industry_2020,tesla2019}. 
Particularly,} the recent advances in Deep Learning (DL) methods for video processing backed by low-cost and high-speed computational platforms such as Graphics/Tensor Processing Units (GPUs/TPUs) have accelerated the pace of developing AI-based features both at vehicle and infrastructure levels \cite{zheng2014traffic}.









\ar{
Commercial transportation has witnessed unprecedented growth in utilizing autonomous driving systems, in recent years. Waymo launched a self-driving taxi service in Phoenix, Arizona in 2020 \cite{waymo2020}. Almost the same time, General Motors (GM) presented their 
prototype for autonomous buses \cite{gm2022origin}. Meanwhile, Amazon revealed 
their future public transportation plan, which offers passengers-only compact autonomous shuttles \cite{zoox2020}. Starship robots \cite{starship2020} and Nuro's autonomous delivery bots \cite{nuro2022} are other examples of AI-based commercial traffic platforms. The autonomous semi-truck concept, independently presented by Tesla and Waymo \cite{tesla2022semi, waymo2022via}, brings tons of benefits to logistic and cold chain monitoring, building a solid fundamental for industry traffic.}

\ar{In the industry domain, camera-based traffic monitoring systems have been broadly used to detect incidents and apply congestion control \cite{flir2020incident,trafficvision2016}. Sensor-based adaptive traffic light control systems are used to reduce traffic jams \cite{flir2020signal,notraffic2020,lee2020design}. Modern systems use Vehicle-To-Vehicle (V2V) and Vehicle-To-Infrastructure (V2I) communications to enhance the overall driving safety. For instance, Roadside Unit (RSU)-based radar systems with real-time data analysis are used to monitor pedestrians on the road and minimize crash rates by sending alerts to connected vehicles \cite{smartmicro2019}. These technologies have optimized existing infrastructures and greatly increased traffic safety.}

\subsection{Architectural View}
Traffic Safety analysis can be viewed as a modular and multi-faceted problem that involves many aspects. As shown in Figs. \ref{fig:network1} and \ref{fig:network2}, the overall analysis platform can be viewed as a software pipeline where the collected information undergoes different processing steps until it is translated to navigation commands, advisory messages, or overall guidelines for improving traffic safety. 

This paper provides a comprehensive review of the popular methods, tools, software packages, and datasets developed by the scientific community for each \ar{vision-based} sub-problem. We also highlight the open problems and future challenges on each frontier. Our main emphasis is on the role of \ar{vision-based} DL methods in enhancing traffic flow (e.g., improving efficiency)  and mitigating traffic safety risks.

\begin{figure}[h]
\begin{center}
\centerline{\includegraphics[width=0.4\columnwidth]{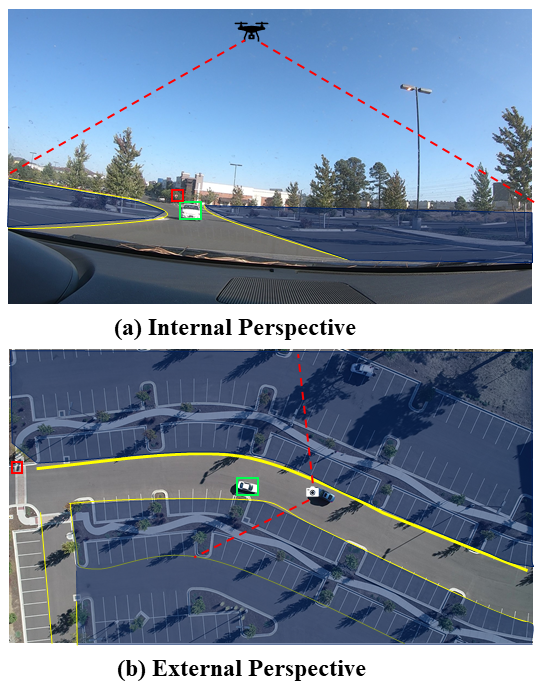}}
\caption{The example of Perspectives: (a) Internal Perspective and (b) External Perspective.}
\label{fig:Inter_exter}
\end{center}
\end{figure}

In contrast to the primary trend of processing individual events from an involved vehicle's perspective, we consider traffic safety at both the vehicle-level and network-level by processing videos captured by an external observer. The main information source is captured video by vehicle-mounted cameras and roadside cameras, but we also will review other sensor information that can be used for traffic management. 
We believe that enabling advanced traffic safety analysis and monitoring platforms will play a crucial role in future smart cities.  Fig.\ref{fig:Inter_exter} compares video-analysis from two different perspectives, from an external point of view as well as an internal node's perspective.

\begin{table*}[htbp]
\centering
\caption{Summary of related review papers. The paper with '*' means although this paper is related but outdated. $L\surd$ denotes a topic is covered in fewer details. }
\resizebox{\textwidth}{!}{
\begin{tabular}{cccccccccccccc} 
\toprule \textbf{Paper}                                                                                                     & \cite{hu2020review}Hu et al.& \cite{mozaffari2020deep}Mozaffari et al. & \cite{grigorescu2020survey}Grigorescu et al. & \cite{yurtsever2020survey}Yurtsever et al. & \cite{janai2017computer}Janai et al. & \cite{badue2020self}Badue et al. & \cite{kumaran2019anomaly}kumaran et al. & \cite{wang2019enhancing}Wang et al. & \cite{nguyen2018deep}Nguyen et al.& *\cite{shirazi2016looking}shirazi et al.& *\cite{mukhtar2015vehicle}Mukhtar et al. & *\cite{morris2013understanding} Morris et al.& Ours         \\ 
\midrule 
\textbf{Year}                                                                          & 2020                  & 2020                  & 2020                  & 2020                  & 2020                  & 2020                  & 2019                  & 2019                  & 2018                  & 2016                  & 2015                  & 2013                  &                              \\ 
\midrule
\begin{tabular}[c]{@{}c@{}}\textbf{Human-driven}\\\textbf{ Vehicle}\end{tabular}       & $\surd$               & $\surd$               &                       & $\surd$               &                       &                       & $\surd$               & $\surd$               & $\surd$               & $\surd$               & $\surd$               & $\surd$               & $\surd$                      \\ 
\midrule
\textbf{AVs}                                                                           & $\surd$               & $\surd$               & $\surd$               & $\surd$               & $\surd$               & $\surd$               & $\surd$               &                       & $L\surd$              &                       &                       &                       & $\surd$                      \\ 
\midrule
\begin{tabular}[c]{@{}c@{}}\textbf{Safety Assessment}\\\textbf{ Analysis}\end{tabular} & $\surd$               & $\surd$               & $\surd$               & $\surd$               &                       &                       & $\surd$               &                       &                       & $\surd$               &                       & $\surd$               & $\surd$                      \\ 
\midrule
\begin{tabular}[c]{@{}c@{}}\textbf{CV-based}\\\textbf{ Method}\end{tabular}            & $\surd$               & $L\surd$              & $\surd$               & $\surd$               & $\surd$               & $\surd$               & $\surd$               & $\surd$               & $L\surd$              & $\surd$               & $\surd$               & $\surd$               & $\surd$                      \\ 
\midrule
\begin{tabular}[c]{@{}c@{}}\textbf{Deep Learning}\\\textbf{ Method}\end{tabular}       & $L\surd$              & $\surd$               & $\surd$               & $\surd$               & $\surd$               & $L\surd$              & $\surd$               & $\surd$               & $L\surd$              &                       &                       &                       & $\surd$                      \\ 
\midrule
\textbf{Sensors}                                                                       & $\surd$               &                       & $\surd$               & $\surd$               & $\surd$               & $\surd$               &                       &                       &                       & $\surd$               & $\surd$               &                       &         $\surd$                     \\ 
\midrule
\textbf{Datasets}                                                                      &                       &                       & $\surd$               & $\surd$               & $\surd$               &                       & $\surd$               &                       &                       & $\surd$               &                       & $\surd$               &  $\surd$                            \\ 
\midrule
\begin{tabular}[c]{@{}c@{}}\textbf{Network}\\\textbf{ Analysis}\end{tabular}           & $\surd$               & $\surd$               &                       &                       &                       &                       & $\surd$               & $\surd$               & $\surd$               &                       &                       & $\surd$               & $\surd$                      \\ 
\midrule
\begin{tabular}[c]{@{}c@{}}\textbf{Vehicular }\\\textbf{Edge Computing}\end{tabular}   &                       &                       &                       &                       &                       &                       &                       &                       &                       &                       &                       &                       & $\surd$                      \\ 
\midrule
\begin{tabular}[c]{@{}c@{}}\textbf{Behavioral \&}\\\textbf{Driver Cognition}\end{tabular}
& \multicolumn{1}{l}{} & \multicolumn{1}{l}{} & \multicolumn{1}{l}{} & \multicolumn{1}{l}{} & \multicolumn{1}{l}{} & \multicolumn{1}{l}{} & \multicolumn{1}{l}{} & \multicolumn{1}{l}{} & \multicolumn{1}{l}{} & \multicolumn{1}{l}{} & \multicolumn{1}{l}{} & \multicolumn{1}{l}{} & {$\surd$}  \\
\bottomrule
\end{tabular}}\label{tab:surveys}
\end{table*}

\begin{table*}[h]
\caption{Specific details on the shortcomings of the most recent survey papers in CV-based traffic analysis. }
\centering
\resizebox{\textwidth}{!}{
\begin{tabular}{cllcc} 
\toprule
\textbf{Survey}                                    & \textbf{Content}                                                                                                                                                                                                                                                                                           & \textbf{Drawback}                                                                                                                                                                                                                                                                                    & \textbf{Perspective} & \textbf{Application}   \\ 
\midrule
\cite{hu2020review}(2020)         & \begin{tabular}[c]{@{}l@{}}Review existing research works on traffic conflicts based on perception\\technologies and communication technologies.\end{tabular}                                                                                                                                                   & \begin{tabular}[c]{@{}l@{}}Discusses very limited DL methods (covers less than 10 DL methods);\\covers in fewer details the CV tasks\\(including detection, trajectory extraction, and video analysis).\end{tabular}                                                                                   & Internal             & AVs                    \\ 
\midrule
\cite{mozaffari2020deep}(2020)    & \begin{tabular}[c]{@{}l@{}}Reviews DL approaches for Trajectory prediction based on \\the input representation, output type, and prediction method.\end{tabular}                                                                                                                                                & \begin{tabular}[c]{@{}l@{}}Because of the limited works focusing on CV data (e.g. BEV), \\it only discusses 6 (CNN)+4 (CNN+RNN) methods.\end{tabular}                                                                                                                                                 & External             & Avs;
traffic analysis  \\ 
\midrule
\cite{grigorescu2020survey}(2020) & \begin{tabular}[c]{@{}l@{}}Reviews DL approaches (mainly as CNN, RNN, \\Deep Reinforcement Learning) for AVs; \\partially discusses safety assessment.\end{tabular}                                                                                                                                      & \begin{tabular}[c]{@{}l@{}}shallowly covers sensor information; \\less details are provided for performance comparison \\(only covers 10 detection methods on a general dataset and \\4 semantic segmentation methods on CityScapes).\end{tabular}                                             & Internal             & AVs                    \\ 
\midrule
\cite{yurtsever2020survey}(2020)  & \begin{tabular}[c]{@{}l@{}}Reviews practical challenges and solutions for Avs; \\the main tasks involve localization, mapping,\\perception, planning, and human-machine interfaces;\\emphasizes the whole driving system architecture.\end{tabular}                                                & \begin{tabular}[c]{@{}l@{}}Excludes video pre-processing methods (e.g., Video stabilization);\\provides little details on the performance analysis \\(9 methods on ImageNet and 7 methods on KITTI 3D).\end{tabular}                                                                                  & Internal             & AVs                    \\ 
\midrule
\cite{janai2017computer}(2020)    & \begin{tabular}[c]{@{}l@{}}Reviews CV-based approaches for AVs;\\discusses open problems and current challenges.\end{tabular}                                                                                                                                                                          & \begin{tabular}[c]{@{}l@{}}Lacks the safety assessment analysis; \\does not give detailed performance comparisons.\end{tabular}                                                                                                                                                                         & Internal             & AVs                    \\ 
\midrule
\cite{badue2020self}(2020)        & \begin{tabular}[c]{@{}l@{}}Reviews technologies for the perception and \\decision-making system of Avs; \\discusses the AV industry milestones.\end{tabular}                                                                                                                                        & \begin{tabular}[c]{@{}l@{}}Covers limited DL methods (sign and light detection, \\pavement marking detection,\\MOT, excluding segmentation methods); \\does not give detailed performance comparisons.\end{tabular}                                                                                          & Internal          & AVs                    \\ 
\midrule
\cite{kumaran2019anomaly}(2019)   & \begin{tabular}[c]{@{}l@{}}Reviews approaches for \\anomaly detection for surveillance videos.\end{tabular}                                                                                                                                                                                             & \begin{tabular}[c]{@{}l@{}}Only includes tasks related to anomaly detection;\\does not include other CV tasks;\\does not provide comprehensive performance analysis.\end{tabular}                                                                                                                                   & External             & Traffic analysis       \\ 
\midrule
\cite{wang2019enhancing}(2019)    & \begin{tabular}[c]{@{}l@{}}Reviews DL approaches for multiple traffic applications, \\including time series prediction,\\classification, and optimization.\end{tabular}                                                                                                                  & \begin{tabular}[c]{@{}l@{}}Covers a few CV-based problems (including traffic sign recognition,\\vehicle detection, and pedestrian detection,\\excluding MOT and segmentation);\\offers a limitted comparative analysis\\(including only the traffic sign and traffic flow prediction).\end{tabular} & External             & AVs;
traffic analysis  \\ 
\midrule
\cite{nguyen2018deep}(2018)       & Reviews DL approaches for processing traffic data.                                                                                                                                                                                                                                                          & \begin{tabular}[c]{@{}l@{}}Covers a few CV-based problems (including vehicle detection,\\perception on Avs, etc,.) without sufficient detail.\end{tabular}                                                                                                                                                & External             & Traffic analysis       \\ 
\bottomrule
\end{tabular}}\label{tab:surveys2}
\end{table*}
It is noteworthy that several review papers have been published to review methods and tools used for video-based traffic analysis. A summary of these papers is provided in Table \ref{tab:surveys}. However, most review papers have limitations in certain aspects. 

\ifx \myVer \newVer

Some survey papers (e.g., \cite{yurtsever2020survey,janai2017computer,badue2020self,wang2019enhancing,nguyen2018deep}) focused merely on solving traffic-related tasks (such as perception) while not covering safety assessment methods. Some other papers (e.g., \cite{hu2020review,badue2020self}) do not provide a comprehensive summary of DL methods, which recently has become the dominant approach in both industry and academia research. Other papers (e.g., \cite{yurtsever2020survey,janai2017computer,badue2020self}), which review driving techniques for vehicles equipped with Automated Driving Systems (ADS), keep their attention solely on the DL methods developed for AVs while not investigating the practicality of these methods on human-driven vehicles, which still are the most widely used vehicles. A few papers (e.g., \cite{kumaran2019anomaly,hu2020review}) put their primary focus on DL methods but without special emphasis on vision-related tasks particularly useful for traffic analysis. A different set of papers (e.g., \cite{mozaffari2020deep}) limit their analysis to the perspective of internal nodes, also known as the first-person perspective. Although Computer Vision (CV)-based tasks are covered by several papers \cite{wang2019enhancing,nguyen2018deep}, they do not provide sufficient details on this subject from different perspectives. Most of the papers, including \cite{grigorescu2020survey,janai2017computer,badue2020self,kumaran2019anomaly,nguyen2018deep}, although very informative, only review the DL algorithms that are used or can be used for traffic analysis and do not provide any sort of comparative analysis, which does not help choosing the right method for different real-world traffic problems. The relevant surveys we include are from 2013 and after; however, we want to mention some papers, including \cite{shirazi2016looking,mukhtar2015vehicle,morris2013understanding} are outdated, but highly related to our survey with the topics of CV-based traffic safety analysis. 
Table \ref{tab:surveys2} provides more specific details on the main focus and the shortcoming of the recently published survey papers. 

In addition to covering newly published \ar{CV-based} methods, our paper covers the shortcomings of previous surveys and considers the \ar{video-based} driving safety problem from different perspectives. More specifically, we list exemplary problems in \ar{video-based} driving safety analysis; we review requirements and challenges from an external observer's perspective; we review datasets and important industrial developments; we make connections to closely related areas of utilizing crowd-sourcing, and edge and cloud computing for bulk processing; we highlight connections to behavioral science, insurance industry, and other policy maker entities. 
\else
\ar{Old ver: }
For instance, some papers (e.g., \cite{sivaraman2013looking,khan2017uav,wali2015comparative,mukhtar2015vehicle}) which focused on visual-based methods for regular vehicles often consider basic visual-related tasks (such as car detection) while not covering safety assessment methods. Some papers (e.g., \cite{nguyen2018deep}) put their primary focus on DL methods but without special emphasis on vision-related tasks particularly useful for traffic analysis. Some other papers (e.g., \cite{mukhtar2015vehicle,wali2015comparative}) are limited to the perspective of internal nodes (also known as the first-person perspective). 
Furthermore, most of these survey papers do not provide a comprehensive summary of DL methods, which recently has become the dominant research direction in both industry and academia. A few survey papers (e.g.,\cite{janai2017computer,badue2020self,yurtsever2020survey}) that review automated driving system (ADS)-equipped vehicles driving techniques keep their attention solely on the advanced DL methods developed for AVs while not covering the practicality of these methods on human-driven vehicles. It is important to consider both AVs and human-driven vehicles as there is likely to be a mixed traffic stream for some time moving forward. Table \ref{tab:perspective} and Fig.\ref{fig:Inter_exter} compare video-analysis from two different perspectives, from an external point of view as well as an internal node's perspective.
\fi

The rest of this paper is organized as follows. 
\ar{Section \ref{sec:dl} reviews mainstream lines of DL methods used for vision-based driving safety analysis.} In Section \ref{sec:hardware}, data acquisition equipment and methods are reviewed. 
\ifx \myVer \newVer
\fi
Section \ref{sec:video-preproc} includes discussions about different stages of video pre-processing for safety analysis by highlighting historical milestones, successful methods, current trends, and remaining challenges for each category. Section \ref{sec:video-proc} reviews DL methods for video processing with application to traffic safety analysis. A short list of sample problems in driving safety analysis is provided in Section \ref{sec:sProblem}. \ar{Section \ref{sec:new-trends} reviews recent trends in deep learning that can influence the field if video-based driving safety analysis.} 
A list of commonly used datasets with applications to traffic monitoring and traffic safety analysis is provided in Section \ref{sec:dataset}. 
\ifx \myVer \newVer
Section \ref{sec:safety-metrics} list a set of key safety metrics used for 
assessing the potential for crash occurrence and crash severity.
\else
In Section \ref{sec:safety-metrics}, a comprehensive review of key safety metrics is provided, followed by a set of newly proposed network-level metrics for future studies. 
In Section \ref{sec:safety-metrics}, a comprehensive review of key safety metrics is provided, followed by a set of newly proposed network-level metrics for future studies. 
\fi
\ar{
Section \ref{sec:points} provides different key points such as connection to other fields and potential applications of safety analysis methods.
A roadmap of this technology is offered in Section \ref{sec:roadmap}. Remaining challenges and issues are discussed in section \ref{sec:challenges}.}


\section{Deep Learning Methods}  \label{sec:dl}

\ar{The common core of vision-based driving safety analyses is using deep learning methods for image/video processing. We skip the details of DL methods here for the sake of brevity and refer the interested reader to previous reviews \cite{lecun2015deep,goodfellow2016deep}.}

\ar{It is noteworthy that most of the recent developments in DL have been driven by two applications, Computer Vision (CV), and Natural Language Processing (NLP), as two key representatives of visual and sequential processing problems. 
Most elegantly designed DL platforms utilize Multilayer Perceptron (MLP), Convolutional Neural Networks (CNN), Recurrent Neural Networks (RNN), and transformers as their building blocks.  
Table \ref{tab:compare_dl} provides a compact and informative comparison of these methods. We will provide a more detailed analysis of custom-built DL methods used in the context of driving video analysis and safety control. } 

\begin{table*}[hb]
\centering
\caption{A brief comparison among typical DL architectures.}
\label{tab:compare_dl}
\resizebox{\textwidth}{!}{%
\blr
\begin{tabular}{lllll} \toprule
\textbf{Architecture} & \textbf{MLP} & \textbf{CNN} & \textbf{RNN/LSTM} & \textbf{Transformer} \\ \midrule
Pros & Straightforward to design & \begin{tabular}[c]{@{}l@{}} Appropriate for high-dimensional  data\\      Learns features with locality\\      Shift-invariance\\      Fewer parameters\\      Powerful on CV tasks\end{tabular} & \begin{tabular}[c]{@{}l@{}}Appropriate for sequential data\\      Can learn long-term dependencies\end{tabular} & \begin{tabular}[c]{@{}l@{}}Processes sequences in parallel\\      Appropriate for long sequences\\Captures long-term dependencies\\     Enables self-attention mechanism\\      Powerful on both CV and NLP tasks\end{tabular} \\  \midrule
Cons & \begin{tabular}[c]{@{}l@{}} Many parameters (dense connections)\\      Limited capability\end{tabular} & \begin{tabular}[c]{@{}l@{}}Do not encode the position of   object\\      Gradient vanishing in deep architectures \\      (solved by ResNet and auxiliary output)\\    Weak on long sequential data\end{tabular} & \begin{tabular}[c]{@{}l@{}}Hard to train for gradient issues\\      Handles only serially fed data\\      Powerless on extreme long-term dependencies\end{tabular} & \begin{tabular}[c]{@{}l@{}}Many parameters (long training time) \\      Does not consider locality\\ Over complicated for 
short sequences\end{tabular} \\ \midrule
Inductive Bias & Weak & Locality/spatial invariance & Sequentiality/time invariance & Weak \\
\bottomrule
\end{tabular}%
}
\end{table*}

\ar{Training a network is not always straightforward and can be impacted by many factors, such as hyper-parameters (mini-batch size and learning rate). Due to the time sensitivity of safety-related applications, fast training is a key requirement. Table \ref{tab:DL_tricks} lists some common tricks (e.g., dropout, regularization, easy-derivative Rectified Linear Unit (ReLU) and leaky ReLU activation function, data augmentation, and soft labeling) to accelerate the training process and boost the recognition performance. 
For example, word embedding uses low-dimensional space to present high-dimensional data allowing words with similar meanings to be closer in the low-dimensional space. This often outperforms the one-hot encoding. Such tricks can be translated from the NLP context to driving behavior analysis. An exemplary scenario would be the personalized driving anomaly detection. Suppose there are 20 collected features for each driver. If there are 100 drivers, by one-hot encoding, the input dimensions of data should be $100+20=120$ per time point, 
which leads to the \textit{curse of dimensionality}. In fact, difference among different drivers (or driving styles) can be well represented by as few as 20 dimensions.}

 \begin{table}[]
\centering
\caption{\ar{Common tricks to accelerate DL training.}}
\label{tab:DL_tricks}
\resizebox{0.49\columnwidth}{!}{%
\blr
\begin{tabular}{ll}\toprule
\textbf{Name} & \textbf{Description} \\ \midrule
Data Normalization & Stabilize the training \\ \midrule
Xavier/Kaiming Initi. & Initialize based on the magnitudes    of signals \\ \midrule
\begin{tabular}[c]{@{}l@{}}Batch/Layer/Instance \\ Normalization\end{tabular} & \begin{tabular}[c]{@{}l@{}}Local response   normalization\\      Regularize\\      Avoid gradient vanishing\end{tabular} \\ \midrule
Dropout & \begin{tabular}[c]{@{}l@{}} Reduce number of parameters \\ Avoid overfitting \end{tabular}\\ \midrule
L1/L2 regularization & Avoid overfitting \\ \midrule 
Weight decay & \begin{tabular}[c]{@{}l@{}}Similar to L2   regularization\\      but different in some optimizers (e.g. Adam)\end{tabular} \\ \midrule
LeakyRelu & Avoid dead neurons (as for Relu) \\ \midrule
Word Embedding & Manage high-space categorical data \\ \midrule
Soft Labeling & \begin{tabular}[c]{@{}l@{}}State the relations between   different classes;\\      Improve generalization\end{tabular} \\ \midrule
Data Augmentation & \begin{tabular}[c]{@{}l@{}}Improve the variety of   data\\      Improve generalization\end{tabular} \\ \midrule
Residual Connection & \begin{tabular}[c]{@{}l@{}}Avoid gradient vanishing\\      Allow building very deep networks\end{tabular} \\ \midrule
$1\times 1$ Conv & Combine channels \\ \midrule
Gradient Clipping & Avoid gradient explosion \\ \bottomrule
\end{tabular}%
}
\end{table}

\section{Data Acquisition}   \label{sec:hardware}


In this section, we investigate the role of data acquisition in developing safety-related algorithms. First, we review data modalities and hardware used for information acquisition. Next, we list the most commonly used datasets for testing the developed algorithms.


Most traffic analysis platforms rely on data collected by different types of sensors, including cameras, Global Positioning Systems (GPS), Radio Detection And Ranging (Radar), and Light Detection and Ranging (LiDAR). These sensors (except GPS) can be used in vehicles or on external observer systems such as roadside infrastructures, drone-based aerial platforms, etc. 
The following is a short description of sensors followed by a summary provided in Table \ref{tab:sensors}.

\textbf{Video Cameras} are the most widely used means of information collection. Modern cars are heavily equipped with cameras in various parts to capture imagery for processing. Using thermographic cameras for night vision has become more common than ever. 
Camera feeds can be used by the driver (e.g., backup camera) to minimize safety risks, or by the control computer in an AV for automated driving. The imagery can also be used by more advanced AI platforms for driver's cognition assessment in real-time mode (e.g., driver drowsiness assessment \cite{affectiva}, and distraction awareness \cite{mejia_2020}). Volume collections of imagery can be stored for further analysis for transportation infrastructure revisions.

\textbf{GPS} systems provide an accurate position of the vehicle by communicating to GPS satellites. 
The accuracy of GPS depends on different factors, including satellite geometry, signal blockage, atmospheric conditions, and receiver design features/quality. The global average User Range Error (URE) of GPS can be as low as 2.3 ft\cite{gpsIMAGE}. With recent advances in 5G wireless networking, even preciser positioning is available for vehicles. For instance, Verizon launched its Hyper Precise Location (HPL) using Real-Time Kinematics (RTK), with an unprecedented accuracy of 1-2 centimeters\cite{radio_resource_2028}.

\textbf{Radar} operates based on Doppler frequency shift in the reflected wave \cite{munoz2008traffic}. Radar units are often installed on modern cars, roadside poles, police vehicles, and portable speedometer guns to measure the absolute or the relative speed of other vehicles. Radars are low-cost and relatively robust devices appropriate for different weather conditions and illumination intensities. Some radar units may have a narrow Field of View (FoV) but are capable of long-range detection \cite{shirazi2016looking}. A traditional radar is a single source-detector, which does not have the spatial resolution required for precise scene/environment description; however, in recent years, imaging radars have been developed by adopting multiple-input, multiple-output (MIMO), and radar-on-chip technologies. Although such advanced technology is still expensive, attempts are made to reduce the cost of broader adoption\cite{bilik2019rise}.

\textbf{LiDAR} uses laser beam reflection to enable accurate positioning down to centimeter's scale \cite{tan2016weakly}. The 3D scanning of multiple laser beams provides a 3D point cloud image (3D map) of the surrounding obstacles with accuracy much higher than regular radars \cite{ma2018mobile}. LiDARs send out a near-infrared laser beam and detect reflections from the object; thus, it can still operate in dark conditions, in contrast to visual sensors. Its use is less common than radars for a few reasons, including its higher cost, relatively sparse spatial resolution, especially for a limited number of scanning laser beams, extremely narrow FoV, and computational complexity of $360^\circ$ scanning, noting that low-complexity point cloud methods are still under development. Solid-state LiDARs have been developed to reduce the cost, while enhancing the spatial resolution\cite{erichsen2020semantic}.

\textbf{Drones} are commonly used nowadays to implement aerial monitoring systems. Most external observer systems utilize sensors in roadside infrastructures. However, the use of drones, also known as Unmanned Aerial Vehicles (UAVs), is gaining more attention in different applications to enable fast, low-cost, and on-demand monitoring \cite{khan2017uav}, and traffic analysis is not an exception. Particularly, drones can provide top-view and clearer occlusion-free images of the traffic flow when needed \cite{ke2018real}. A network of cooperative drones can collectively cover relatively large areas \cite{barmpounakis2020new}. Modern drones may be equipped with advanced sensing platforms, high-resolution cameras, and more importantly advanced features such as learning-based image processing, AI-based autonomous control, collision avoidance and auto lading, auto-calibration, real-time transmission, object tracking, and image restoration. The key challenges of drones are their limited payload, flight time, and communication range under study by several research teams. Aerial images pose new challenges to the research community, such as tackling image stabilization, small object recognition, and developing lightweight ML algorithms customized for top-view images. 

\textbf{Other Sensors:} 
In addition to the aforementioned broadly-used sensors, there exist some custom-built advanced sensors that can be used for traffic monitoring and driving safety analysis. Passive Infrared sensors\cite{ahmed1994active}, Inductive Loops\cite{mimbela2007summary}, and Piezoelectric sensors\cite{zhang2015new} can enable measuring basic flow parameters such as vehicle count, speed, and flow volume, noting that Piezoelectric sensors are still used for weigh-in-motion measurement. Environmental Sensor Stations (ESS)\cite{ESS} on the roadway can collect atmospheric data, including air temperature and humidity, visibility distance, wind speed, wind direction, etc., as side parameters to be used for traffic analysis. 

 \begin{table*}[ht]
\centering\caption{Comparison of sensors. Here “$\surd$” means this kind of measure has been implemented or available to complete.}
\resizebox{0.8\textwidth}{!}{
\begin{tabular}{lccccccccc} 
\toprule
\multirow{2}{*}{\textbf{Sensor Type}} &
  \multicolumn{5}{l}{~~~~~~~~~~~~~~~~~~~~~~~~~~~~~~~~~~~~~~~~~~~~\textbf{Measures Type}} &
  \multirow{2}{*}{\textbf{Cost}} &
  \multirow{2}{*}{\begin{tabular}[c]{@{}c@{}}\textbf{Shape}\\ \textbf{Modeling}\end{tabular}}&
  \multirow{2}{*}{\begin{tabular}[c]{@{}c@{}}\textbf{Robust to}\\ \textbf{Bad Environment}\footnotemark[4]\end{tabular}}
  \\\cmidrule{2-6}
 &
  \begin{tabular}[c]{@{}c@{}}Vehicle\\ Count\end{tabular}  &
  \begin{tabular}[c]{@{}c@{}}Speed/Distance\\ Estimation\end{tabular} &
  \begin{tabular}[c]{@{}c@{}}Vehicle\\ Classification\end{tabular}&
  \begin{tabular}[c]{@{}c@{}}\ar{Pedestrian}\\ \ar{Detection}\end{tabular}&

  Road feature &
  &
  \\ \midrule
Camera (Visible) & $\surd$ & Hard & $\surd$ &\ar{$\surd$}  & $\surd$ & Low    & \ar{2D} & \ar{Low}  \\ \midrule
\ar{Camera (Infrared)} & \ar{$\surd$} & \ar{Hard}   & \ar{$\surd$} &\ar{$\surd$} &  \ar{$\surd$} & \ar{Low}    & \ar{2D} & \ar{High}   \\ \midrule
LiDAR  & $\surd$ & Easy & $\surd$ &\ar{$\surd$}  & $\surd$ & High   & \ar{3D}  & \ar{Middle}   \\ \midrule
Radar  & $\surd$ & Easy  &   &\ar{$\surd$} &     & Moderate & \ar{weak} & \ar{High}  \\ \midrule
UAV Camera & $\surd$ & Moderate  & $\surd$  &\ar{$\surd$}  &   $\surd$   & Low & \ar{2D}& \ar{Low } \\ \bottomrule
\end{tabular}
}
\label{tab:sensors}
\end{table*}
 
\ar{We recognize that the properties of sensors are varied. For example, visual cameras require clean and high-visible environments and offer a richer set of information in terms of color space and visual geometry, hence are more appropriate for detection and classification applications. However, interpreting imagery may require more computation powers. LiDARs, on the other hand, provide more accurate precision for object detection and depth and speed estimation but are expensive and less energy efficient. Some manufacturers like Tesla prefer pure vision-based perception \cite{tesla_2021}. However, fusing multiple sensors can be advantageous from the safety perspective and is adopted by more vendors, as discussed in \cite{fayyad2020deep}.} 

\footnotetext[4]{\ar{Noting that visual cameras may perform poorly under low illumination, and both visual cameras and LiDARs are limited by the bad weather, such as fog, dust, rain, or snow.}}



\section{Video and Image Pre-processing}   \label{sec:video-preproc}

In this section, we review different stages of a typical video-based traffic analysis framework and highlight key developments, historical milestones, current trends, and existing challenges. 

\textbf{Super Resolution}: The video and image super-resolution aims to reconstruct a Higher Resolution (HR) result from a Low Resolution (LR) observation. Super-resolution is a typical stage in image pre-processing and can be applied to traffic imagery to enhance the performance of the subsequent learning tasks, such as vehicle classification and license plate detection. One popular supervised learning method is the DL-based Single Image Super-Resolution (SISR) method which creates a mapping between the low and high-resolution images by training a deep CNN. Most of the existing learning-based SISR methods are trained and evaluated using simulated datasets~\cite{agustsson2017ntire, fujimoto2016manga109, blau20182018, huang2015single}, where the LR images are generated by applying a hand-crafted degradation process into the HR samples. For instance, one may apply bi-cubic down-sampling to the original HR samples to obtain LR results. 
Recently, more advanced SISR~\cite{cai2019toward} methods are developed for real-world applications with unknown and more complicated degradation processes, which can be used as a benchmark method for traffic image analysis as well.

In contrast to the simple spatial interpolation used in the SISR family for image processing, Video Super-Resolution (VSR) methods utilize both spatial and temporal relationships between consecutive frames to improve the quality of the reconstructed videos\cite{wang2020deep}. 
These methods are essential in processing roadside traffic videos, especially under poor visibility in foggy, rainy, and cloudy weather conditions.

An important application of SR methods is license plate detection for vehicle identification. Early works often focused on conventional signal processing methods. For instance, \cite{4220667,6135797} deployed a Markov random fields-based method for plate detection. \cite{5604296} proposed a Gaussian Mixture Model (GMM) to enhance the plate location and SR reconstruction. Compressed Sensing (CS)-based methods\cite{yang2010image,dong2011image,zhang2013super} can also be used to address this task by enforcing the sparsity of images in the frequency domain, which is equivalent to smoothness in the spatial domain. Recently, DL-based SR algorithms are proposed, which perform more accurately and efficiently. \cite{vasek2018license} is an example of such methods which use a CNN architecture to convert a low-resolution license plate into a high-resolution version. Some recent works  \cite{liu2017beyond,8552121,lee2019practical,zhang2018joint} tend to use Generative Adversarial Networks (GAN) as their processing framework, which achieves a higher performance using a more reasonable real-time loss along with an adversarial loss, when inferring. 

\textbf{Denoising}: This is another critical pre-processing task to compensate for imaging artifacts and obtain clear and noise-free images before feeding them into the subsequent learning modules. This is a critical step in processing traffic imagery, especially when taken in motion or under low illumination and poor environmental conditions like rainy, cloudy, and foggy weather.   

Conventional methods typically use filtering, interpolation, and smoothing methods either in time, frequency, or wavelet domains to remove noise from the captured images. In contrast, newer methods use more advanced concepts such as sparsity in the frequency domain, dictionary learning to model common noise patterns, prior knowledge about the noise model, and noise pattern discovery to more elegantly remove the noise from the captured images. 

Traditional methods suffer from several shortcomings, including (1) involving complex optimization methods in some cases, (2) the need for manual parameter setting (e.g., the scale factor of Gaussian spatial filtering), (3) and using a fixed model which deems inflexible in tackling different noise patterns and ignore the learnability of some noises. 

DL-based video and image denoising algorithms take advantage of the neural networks to learn the spatial or temporal dependency between pixels to reconstruct clean samples by end2end training and inferring. Therefore, DL methods provide sufficient flexibility in adapting to different conditions. In most research works on image denoising, a synthetic Additive white Gaussian Noise (AWGN) model is adopted to simulate the noise and evaluate the algorithm. Using a synthetic AWGN noise model has the clear advantage of simplifying the testing phase and quantifying the noise impact. However, it might oversimplify the problem since the real-world noise models can be more complicated depending on the noise source. Further, one may benefit from exploiting common noise patterns for more structured noises. For instance, the noise caused by rainy conditions may need a different treatment than a noise caused by the camera lens scratch. For instance, \cite{sidd, dnd} generates noisy and clean image pairs by controlling the ISO (sensitivity to light) of the cameras. By these approaches, the collected data could be used to emulate the camera-related noise under real-world conditions. 

Similar to SR methods, denoising methods are typically used for generating clean traffic images that could be used to improve the precision of higher-level tasks. For example, in~\cite{yuan2014efficient}, a low-rank decomposition image denoising method is proposed for restoring the noisy traffic image. Likewise, in~\cite{chakraborty2019data}, spatio-temporally denoised images are used to enhance the performance of the traffic incident detection algorithm. 


\begin{table*}[ht]
\centering
\caption{Examples of fine-grained vehicle classification.}
\resizebox{0.8\textwidth}{!}{%
\begin{tabular}{lll}
\toprule
\textbf{Method} &
  \textbf{Paper:[Ref] Authors (year)} &
  \textbf{Performance Accuracy [Dataset]} \\ \midrule
3D box+CNN &
  \cite{sochor2016boxcars}Sochor et al.(2016) &
  83.20\%  on  BoxCars116k(self)\\ \midrule
3D box+Vgg-16 &
  \cite{sochor2018boxcars}Sochor et al.(2018) &
  92.27\% on  BoxCars116k(self) \\ \midrule
DenseNet-161 &
  \cite{ma2019fine}Ma et al.(2020) &
  \begin{tabular}[c]{@{}l@{}}93.81 \% on Stanford Cars\cite{krause20133d}\\      97.89 \% on    COMPCARS\cite{yang2015large}\end{tabular} \\ \midrule
DenseNet-161 &
  \cite{li2019dual}Li et al.(2019) &
  93.51\% on Stanford Cars\cite{krause20133d} \\ \midrule
part-based +DenseNet-264 &
  \cite{xiang2019global}Xiang et al.(2019) &
  \begin{tabular}[c]{@{}l@{}}94.3\% on Stanford Cars\cite{krause20133d}\\      99.6\% on COMPCARS\cite{yang2015large}\end{tabular} \\ \midrule
Part-based+RCNN+SVM &
  \cite{huang2016fine}Huang et al.(2016) &
  87.3\% \\ \midrule
Faster R-CNN+Feature Fusion &
  \cite{zhu2019fine}Zhu et al.(2019) &
  Top 1: 79.1\%, Top 5: 94.1\% on COMPCARS\cite{yang2015large} \\ \midrule
CNN+Feature Fusion &
  \cite{yu2018ff}Yu et al.(2018) &
  98.89\% on COMPCARS\cite{yang2015large} \\ \midrule
ResNet-50 + Attention &
  \cite{yu2020cam}Yu et al.(2020) &
  \begin{tabular}[c]{@{}l@{}}93.1\% on Stanford Cars\cite{krause20133d}\\      95.3\% on COMPCARS\cite{yang2015large}\end{tabular} \\ \midrule
Attention &
  \cite{ke2020fine}Ke et al.(2020) &
  \begin{tabular}[c]{@{}l@{}}94.5\% mAP on Stanford Cars\cite{krause20133d}\\      95.8\% FZU dataset (self)\end{tabular} \\ \midrule
multi-task CNN &
  \cite{hu2017location}Hu et al.(2017) &
  Top 1: 91\%, Top 5: 97.7\% on COMPCARS\cite{yang2015large} \\ \midrule
CNN+filter bank &
  \cite{wang2018learning}Wang et al.(2018) &
  93.8\% on Stanford Cars\cite{krause20133d} \\ \midrule
Fine-tuning Vgg-16 &
  \cite{zhang2018fine}Zhang et al.(2018) &
  98.71\% on COMPCARS\cite{yang2015large} \\ \bottomrule
\end{tabular}%
}
\label{tab:fine-grained}
\end{table*}

\textbf{Video Stabilization}: Traffic videos may contain vibrations, especially when captured by vehicle dashboard-cameras while driving on rough roads. This may underline the performance of the subsequent processing stages (e.g., vehicle detection, speed estimation, etc.). Digital video stabilization techniques are proposed to improve the visual quality of the captured videos~\cite{ling2016feedback, zhao2019fast}. 
The common spirit of most video stabilization methods is extracting trajectory of objects or their representative feature points between consecutive frames, and the re-aligning the frames to smooth out the trajectories, based on the assumption that noise-like high-frequency fluctuations, especially when shared among most image descriptors, are caused by the camera shakes.
We can subdivide these methods into pixel-based and feature-based methods. The pixel-based methods typically use block matching~\cite{tang2011fast,xu2006digital}, phase information~\cite{kwon2005video, zhu2006video}, and optical flow~\cite{liu2014steadyflow, chang2004robust} to estimate the camera motion. Feature point detection methods can be used to convert high-dimensional images into low-dimensional representations to reduce the computation overhead. In~\cite{amisha2015survey}, Scale-Invariant Feature Transform (SIFT)~\cite{lowe1999object}, Speeded Up Robust Features (SURF)~\cite{bay2006surf}, and other feature point extraction methods are compared for evaluating their impacts  on video stabilization. Due to the computational complexity of video stabilization methods, it is often performed offline, which may not satisfy the requirements of real-time video-processing tasks. Recently, DL-based methods \cite{wang2018deep, xu2018deep, lee2019video} enable fast and accurate online stabilization in almost a real-time fashion by instant processing of each incoming video frame with low latency. 

Most DL-based methods utilize CNN and similar architectures for video stabilization using various sources of traffic data. In~\cite{liang2004video,zhang2010central,caraffi2012system}, video stabilization was used for aligning the car-mounted camera captured videos. In~\cite{outay2020applications}, the stabled UAV captured videos were used to obtain higher accuracy for traffic video analysis.

\section{Deep Learning for video processing} \label{sec:video-proc}

Deep learning methods are heavily used for video processing for their outstanding power in solving different problems such as object detection, object recognition, event recognition, and other video understanding tasks in general. DL methods can be considered the brainpower of most AI platforms developed for \ar{video-based} traffic safety analysis.   

\subsection{DL-based Classification Methods in Traffic Analysis} 

\begin{figure}[h]
\begin{center}
\centerline{\includegraphics[width=1\columnwidth]{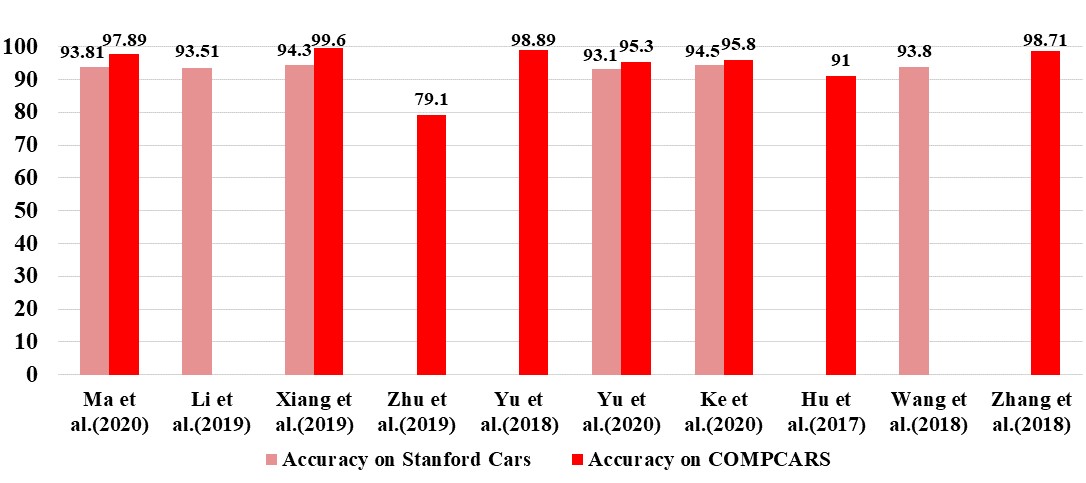}}
\caption{The performance of some fine-grained models on Stanford Cars and COMPCARS.}
\label{fig:VehicleFineGrained_perf}
\end{center}
\end{figure}

\ifx \myver \oldVer
Classification is one of the most fundamental tasks in computer vision. The use of classification in this context can be implemented for object classification (e.g., classifying objects between vehicles, pedestrians, motorcycles, traffic signs, etc.), traffic light and railroad crossing barrier status check, and scene detection (e.g., buildings, roads, road lanes, roadside infrastructures, etc.). It can be developed at different levels, such as the basic level for classifying different object types and fine-grained classification into sub-categories (such as differentiating traffic signs or identifying vehicle classes among sedans, SUVs, trucks, etc.) based on the semantic content of the input.

Due to the generality and flexibility of DL methods, most existing DL methods can be used for traffic safety analysis, like other applications. 
It is noteworthy that various DL network architectures are developed primarily for classification tasks and then further extended for other tasks. Therefore, standard classification problems can be used as a benchmark testing method to evaluate the performance of different network architectures. 
There are popular DL networks whose performance is further boosted by several architectural modifications. 
DL algorithms are nowadays available for different operating systems running on most computer architectures accelerated by GPUs/TPUs, Message Passing Interface (MPI)-based parallel programming, and high-performance computing (HPC). 
Other modifications are implemented to accommodate emerging processing platforms. For instance, lightweight networks are proposed for mobile, FPGA-based, and embedded computation platforms. 

The first Deep Convolutional Neural Networks (D-CNNs) was LeNet~\cite{lecun1998gradient} that achieved state-of-the-art performance on hand-written digit recognition. However, LeNet has a limited performance on large-scale data. 
In 2012, AlexNet~\cite{krizhevsky2012imagenet} realized a breakthrough by dramatically reducing the classification error rate from $26\%$ to $15\%$.
Another breakthrough was the implementation of ImageNet~\cite{russakovsky2015imagenet}, which soon became the new benchmark for image classification tasks. Afterward, a countless number of various deep CNN architectures were developed to tackle large-scale datasets. 
Another notable architecture was VGGnet\cite{simonyan2014very}, which utilized a deeper network architecture to further improve the classification accuracy. But when the network goes deeper, different problems such as computational complexity, error propagation in gradient-based back-propagation for network training, and the need for larger datasets arise by stacking many layers, and the performance of the network saturates with minimal improvement for additional layers.  ResNet~\cite{he2016deep} was proposed in 2016 to address such issues by enabling people to train a much deeper network by applying residual connections to the CNNs. 

Another issue with standard CNNs is the restrictive use of fixed receptive fields in convolutional layers. GoogLeNet~\cite{szegedy2015going} proposed a multi-scale feature fusion architecture that concatenates features extracted by convolution layers with different filter sizes. 
Another remarkable milestone was developing wider than deeper networks. For instance, Wide Residual Networks (WRN)~\cite{zagoruyko2016wide} is proposed to used wider bottlenecks to enhance the performance of ResNet. 

Another research direction is enabling CNNs to execute on mobile and embedded systems. For instance, Depth-Wise Separation Convolution (DWConv) was proposed in~\cite{chollet2017xception}, which achieved similar performance as naive convolution layers with much less computational complexity and fewer parameters. MobileNet~\cite{howard2017mobilenets} and ShuffleNet~\cite{zhang2018shufflenet} are two light-weight network architectures specific for mobile and embedded devices. 

An important issue with the CNNs is that the Convolution layers take all feature channels equally. To make the CNN focus on informative channels, an attention mechanism was introduced in SENet~\cite{hu2018squeeze}. SENet set up channel attention modules to self-guide the networks to filter the less important features that remarkably improved the performance of existed CNNs. Other recent works implement different channel attention mechanisms to improve the representational capacity of convolutional networks \cite{li2019meda}.
The example of fine-grained vehicle model/make classification using CNN-based methods is shown in Fig.\ref{fig:VehicleFineGrained}. 

\begin{figure}[h]
\begin{center}
\centerline{\includegraphics[width=1\columnwidth]{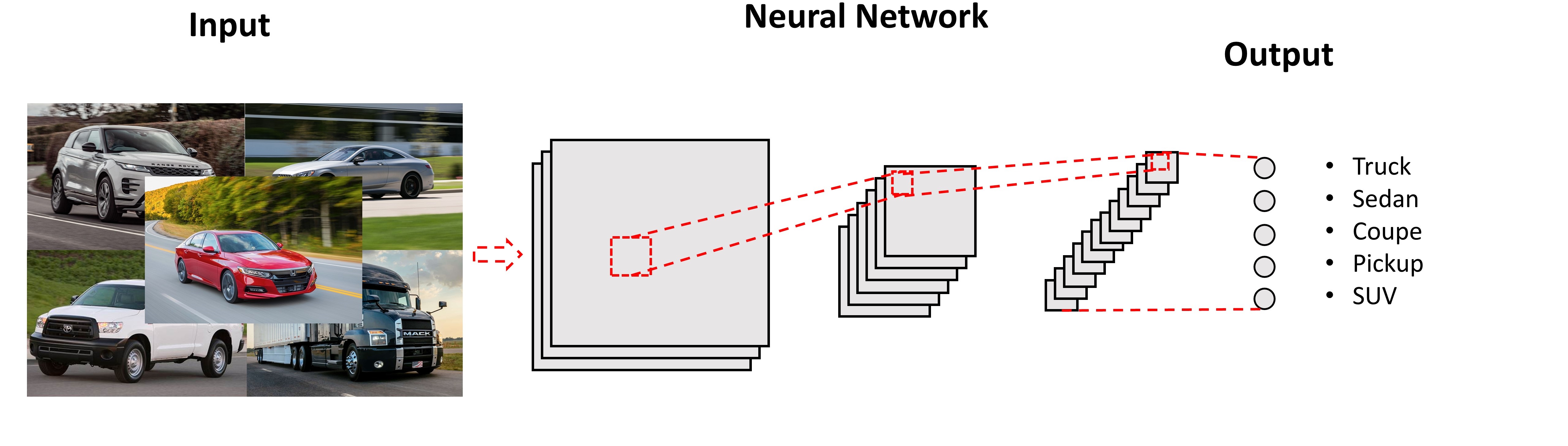}}
\caption{The example of fine-grained vehicle make/model classification.}
\label{fig:VehicleFineGrained}
\end{center}
\end{figure}

\else
Classification is one of the most fundamental tasks in computer vision. The use of classification in this context can be implemented for object classification  (e.g., classifying objects between vehicles, pedestrians, motorcycles, traffic signs, etc.), traffic light and railroad crossing barrier status check, and scene detection (e.g., buildings, roads, road lanes, roadside infrastructures, etc.). Often, it is the backbone or the feature extractor part of the detection networks (e.g., SSD\cite{liu2016ssd}) and segmentation networks (e.g., Mask R-CNN\cite{he2017mask}). It can be developed at different levels, such as the basic level for classifying different object types and fine-grained classification into sub-categories (such as differentiating traffic signs or identifying vehicle classes among sedans, SUVs, trucks, etc.) based on the semantic content of the input. The well-known baseline classification methods include AlexNet(2012) \cite{krizhevsky2012imagenet}, VggNet(2014) \cite{simonyan2014very}, GoogleNet(2015) \cite{szegedy2015going}, ResNet(2016) \cite{he2016deep}, MobileNets(2017) \cite{howard2017mobilenets}, DenseNet(2017)\cite{huang2017densely}, EfficientNet(2019), \cite{tan2019efficientnet} etc. 
\ar{It is noteworthy that ResNet is 
the most highly cited paper in all areas in Google Scholar Metrics 2020, which further proves its extraordinary achievement.} 

\ar{Classification is often considered an upstream task. Therefore, frameworks of the downstream tasks, such as detection, tracking, and segmentation, often use the pre-trained versions of these baselines DL architectures as their backbone to extract the hidden representations. Note that these baseline methods mainly utilized CNN architectures until recently, when a framework named \textit{Transformer} has offered a new breed of deep learning methods with even higher performances, as discussed in Section \ref{sec:transformer}.}

Since the use of classification methods for most traffic-related problems is straightforward and noting the fact that there exist comprehensive reviews on classification methods, we skip the review of classification methods and refer the interested reader to \cite{shen2019survey,schmarje2020survey,zhao2017survey,wang2019development}. Here, we only review fine-grained classification methods that are custom-built or customized for traffic-related problems, as presented in Table \ref{tab:fine-grained} and Fig.\ref{fig:VehicleFineGrained_perf}.

\fi

\begin{figure*}[hb]
\begin{center}
\centerline{\includegraphics[width=1\textwidth]{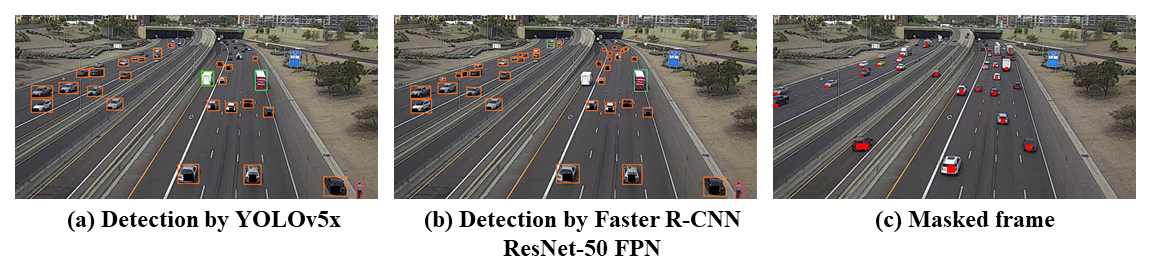}}
\caption{ (a),(b) are the examples of vehicle detection by two alternative methods YOLOv5\cite{glenn_jocher_2021_4679653} and Faster R-CNN\cite{ren2015faster}. They output the location and category of each object with detection confidence. Cars are shown by orange bounding boxes, trucks are shown by lime bounding boxes, buses are shown by green bounding boxes, and persons are shown by red bounding boxes. As shown in (a),(b), YOLOv5 performs worse on small vehicle detection while Faster R-CNN has missed objects in the near zone. (c) is the is PII removed data. 
} 
\label{fig:detection}
\end{center}
\end{figure*}
\subsection{DL-based Object Detection Methods in Traffic Analysis}

Object detection is another key stage in DL-based processing pipelines for driving safety analysis. Object detection simply means locating different objects in images and video frames, potentially with complex backgrounds, by drawing bounding boxes around the objects of interest.It can coexist or be integrated with object classification and labeling. An illustrative example of vehicle detection using 2 benchmark methods is shown in Fig. \ref{fig:detection}(a)(b). 

Notable examples of object detection in the context of driving safety analysis include detecting surrounding vehicles, humans, traffic signs, and obstacles. It also can be part of more complicated tasks such as traffic distribution and composition analysis, improper lane crossing events, trajectory extraction, speed estimation, moving object tracking, path planning, and detecting vehicles on road shoulders, etc., as presented in Table \ref{tab:problem}.

Another use case of object detection is removing Personal Identifiable Information (PII), such as masking human face and license plate numbers before publishing traffic video, as shown in Fig. \ref{fig:detection}(c).



Although there exist some datasets for traffic analysis from the roadside cameras \cite{wang2011automatic,morris2011trajectory,hospedales2012video}, still there is a critical need for larger datasets that cover different zones, urban, suburban, and rural setups, residential and high-risk zones, railroads, and environmental conditions.

Some notable ongoing research problems include solving the trade-off between the algorithm's accuracy and speed, realizing small object detection, distributed and federated learning, model sharing among roadside servers, and  implementing lightweight models and embedded devices appropriate for autonomous vehicles, etc. 

\begin{figure*}[hb]
\begin{center}
\centerline{\includegraphics[width=0.8\textwidth]{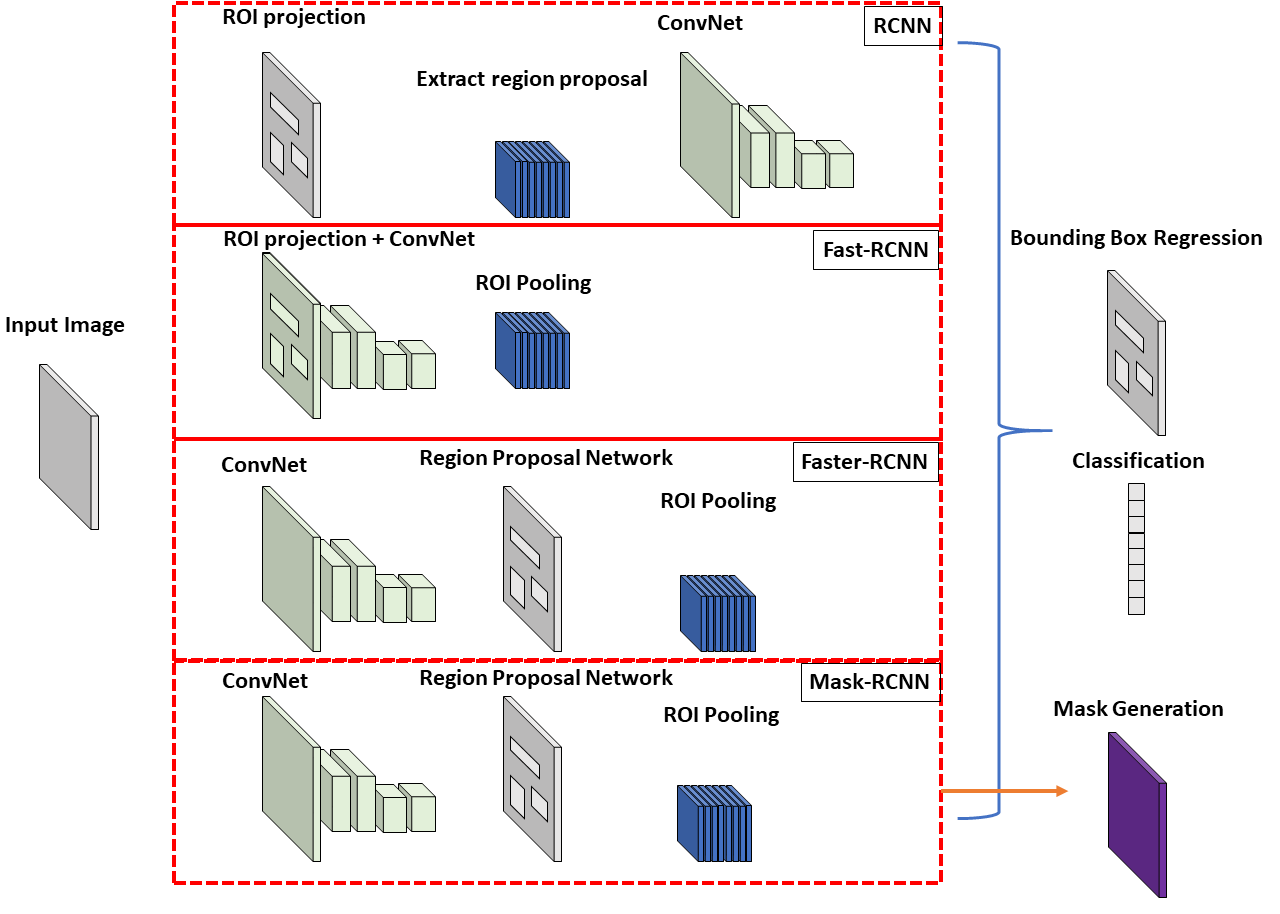}}
\caption{Network architecture for R-CNN family of localization and segmentation.} 
\label{fig:R-CNN family}
\end{center}
\end{figure*}

Compared to the conventional object detection algorithms such as Viola-Jones detector\cite{viola2001rapid}, the Histogram of Oriented Gradients (HOG detector)\cite{dalal2005histograms}, and Deformable Part-based Models (DPM)\cite{felzenszwalb2009object} ), CNN-based methods substantially improve the recognition success rate. 

From the implementation point of view, the DL-based algorithms can be divided into one-stage and two-stage methods. The two-stage detectors first generate Regions of Interests (RoIs) and then send the region proposals down the pipeline for object classification and bounding-box regression. R-CNN series\cite{girshick2014rich,girshick2015fast,ren2015faster,dai2016r,pang2019libra} comprises the most popular two-stage algorithm family. 
\ar{The architectures of R-CNN series are shown in Fig. \ref{fig:R-CNN family}.
Specifically,  R-CNN \cite{girshick2014rich} adopts \textit{Selective Search}\cite{uijlings2013selective} to generate region proposals, and then feed resized (by cropping or warping) proposals into a CNN-based backbone to extract features. Finally, Support Vector Machine (SVM) performs classification for objects' categories and locations. SPPNet \cite{he2015spatial} introduced \textit{Spatial Pyramid Pooling} (SPP) layers, which solved the fixed-size input issue. Integrated by R-CNN and SPPNet, Fast R-CNN \cite{girshick2015fast} realized \textit{RoI pooling} to output fixed size features and adopted multi-task loss to allow single-stage training. Since Fast R-CNN was still bottlenecked by the \textit{Selective Search} heavy computation, Faster R-CNN \cite{ren2015faster} introduced \textit{Region Proposal Network} (RPN), an end-to-end trainable network to generate quality region proposals. Mask R-CNN adds a parallel branch to predict the object mask.}

\begin{figure}[h]
\begin{center}
\centerline{\includegraphics[width=0.5\columnwidth]{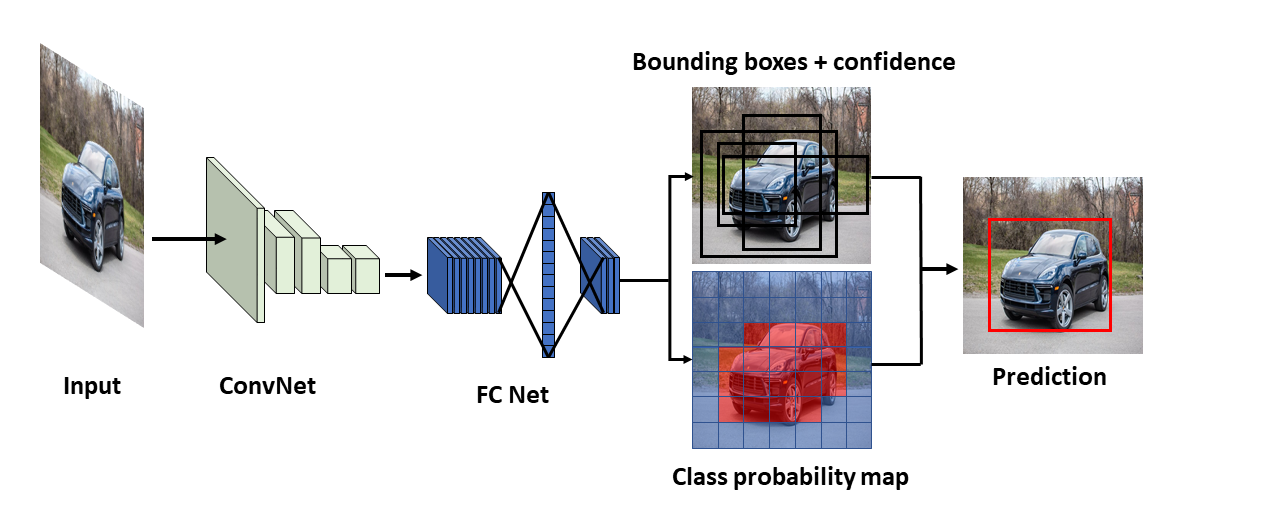}}
\caption{\ar{The demonstration of the basic idea of YOLO.}}
\label{fig:YOLO}
\end{center}
\end{figure}

One-stage detectors directly treat object detection tasks as a regression and classification problem. These methods are divided into anchor-free and anchor-based methods. In anchor-based methods, a set of bounding boxes with different predefined sizes are required to capture the scale and aspect ratio of the objects. Some famous implementations include \textit{You Only Look Once} (YOLO) family (e.g., YOLOv2 \cite{redmon2017yolo9000}, YOLOv3\cite{redmon2018yolov3}, and YOLOv4\cite{bochkovskiy2020yolov4}, and YOLOv5 \cite{glenn_jocher_2021_4679653}\footnote[5]{Note that the authors of YOLOv5 are different from the previous versions}, as well as the Single Shot multi-box Detector (SSD) series\cite{liu2016ssd}). 
\ar{The basic idea of YOLO is demonstrated in Fig. \ref{fig:YOLO}. The network treats the detection as a regression problem and learns the location and class of each bounding box separately. Specifically, an image is split into $S\times S$ patches. In each patch, the network can predict the coordination of $B$ bounding boxes with their confidence levels as well as the class of each patch. After aggregation, the bounding box of an object is localized with its category. Then, non-maximal suppression is applied to filter out the extra bounding boxes for the same object.}

Recently, anchor-free methods are getting more attention to avoid defining anchor-related hyperparameters and to ease complicated computations. The main ideas include implementation by dense prediction (e.g.,  DenseBox\cite{huang2015densebox}, Fully Convolutional One-Stage (FCOS) object detectors \cite{tian2019fcos}, RetinaNet\cite{lin2017focal}) as well as implementation by keypoints and center points (e.g., CornerNet\cite{law2018cornernet,law2019cornernet}, CenterNet\cite{zhou2019objects, duan2019centernet},ExtremeNet\cite{zhou2019bottom}). 

\begin{figure}[hb]
\begin{center}
\centerline{\includegraphics[width=1\columnwidth]{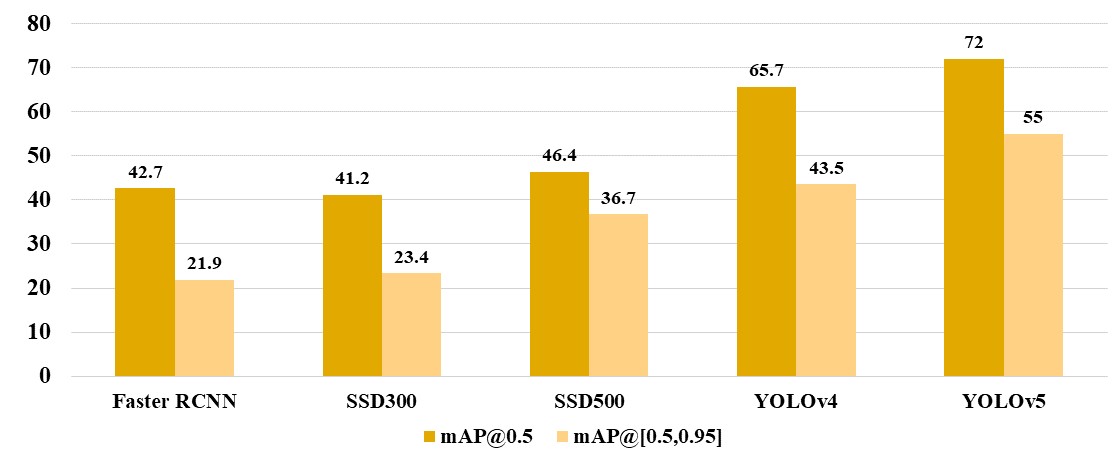}}
\caption{Some object detection models perform on MS COCO test-dev.} 
\label{fig:detection_per}
\end{center}
\end{figure}

Generally speaking, two-stage methods can achieve higher accuracy but at lower speeds than the one-stage methods. Some recent one-stage methods (including YOLO v4\cite{bochkovskiy2020yolov4} and SSD\cite{liu2016ssd}) solve the trade-off between the accuracy and speed by realizing a more efficient network structure. A summary of these algorithms' performance is presented in Fig.\ref{fig:detection_per}.

The applications of object detection methods in traffic video analysis mostly relate to understanding the objects surrounding the road users, such as vehicles, plates, and traffic signs. Most research works in this area adopt one of the aforementioned algorithms for object detection, as summarized in Table \ref{tab:objectdetection}. The metrics used for object detection, recognition, and image segmentation are outlined in Table \ref{tab:somemetrics}.  
We note that some works (such as \cite{6778050,8019461,kim2017deep,selmi2017deep,masood2017license}) tend to fine-tune the CNN framework according to the task requirement, and the recent mainstream traffic works begin to deploy the R-CNN series, YOLO series, and SSD widely. It means these algorithms can stand the test of practice, but it does not mean that the other algorithms are not favorable. It also can be due to the relatively low complexity of this task, or due to the difficulty of deploying recent detectors in real-world scenarios.


\ar{Pedestrians are often the most vulnerable entities on the road, hence pedestrian detection is usually considered a top-priority component of safety assessment and control systems.
Pedestrian detection, especially in crowded areas, would be more complex than similar object detection tasks. Indeed, early CNN-based frameworks underperformed on the task due to the uncertain dense distribution and high dynamics of pedestrians. Some common issues, including the multi-scale problem (different receptive fields with low-information and noisier observations for small-scale pedestrians) and occlusion by crowding (parts of pedestrians are invisible), may challenge these models and decline their accuracy \cite{xiao2021deep}. Some solutions, such as \cite{cai2016unified,li2017scale} addresses the multi-scaling issue by modifying the scale of the region proposals.  Specifically, MS-CNN\cite{cai2016unified} has multiple layers to output features with different receptive fields at the region-proposal sub-network. SAF R-CNN \cite{li2017scale} uses multiple sub-networks to learn the large-scale features and small-scale features, respectively. 
Likewise, \cite{lin2018graininess} uses an attention mechanism by applying a fine-grained attention mask to focus on differently scaled pedestrians. SSA-CA \cite{zhou2019ssa} performs detection after semantic segmentation.}

\ar{Straightforward methods (such as \cite{tian2015deep,xie2021psc}) are proposed to solve the occlusion by enhancing the model to learn patterns with different occlusions. High-level semantic features are employed in \cite{wang2018pcn,liu2019high} to help the occlusion detection. The other explicit solutions \cite{chi2020pedhunter} use data augmentation by introducing extra annotation of the visible parts in the training phase, which substantially enhances the detection ability even under high occlusions. From the author's perspective, the work in \cite{luo2020whether} is very creative, which uses a Cycle-consistent Adversarial Network (Cycle-GAN) \cite{zhu2017unpaired} to translate front view images into bird view images, then solve the detection problem by decomposing it into three subtasks of pedestrian localization, scale prediction, and classification. 
It achieves an outstanding performance of 6.37 Miss Rate (MR) on Caltech (reasonable) and 9.3 MR on CityPersons (reasonable) datasets.}

\ar{We listed some recent works for pedestrian detection in Table \ref{tab:ped_dect}.  These methods for pedestrian detection are often evaluated on Caltech \cite{dollar2011pedestrian} and CityPersons \cite{dollar2011pedestrian} datasets, which both include person-to-person occlusion caused by crowding and other objects. It is notable that these two datasets are designed specifically for pedestrian detection. A more complicated scenario of detecting vehicles and pedestrians simultaneously, is often validated using auto-driving datasets, such as KITTI \cite{Geiger2012CVPR} (see Table \ref{tab:trafficdataset-continue}).}
\begin{figure}[h]
\begin{center}
\centerline{\includegraphics[width=0.8\columnwidth]{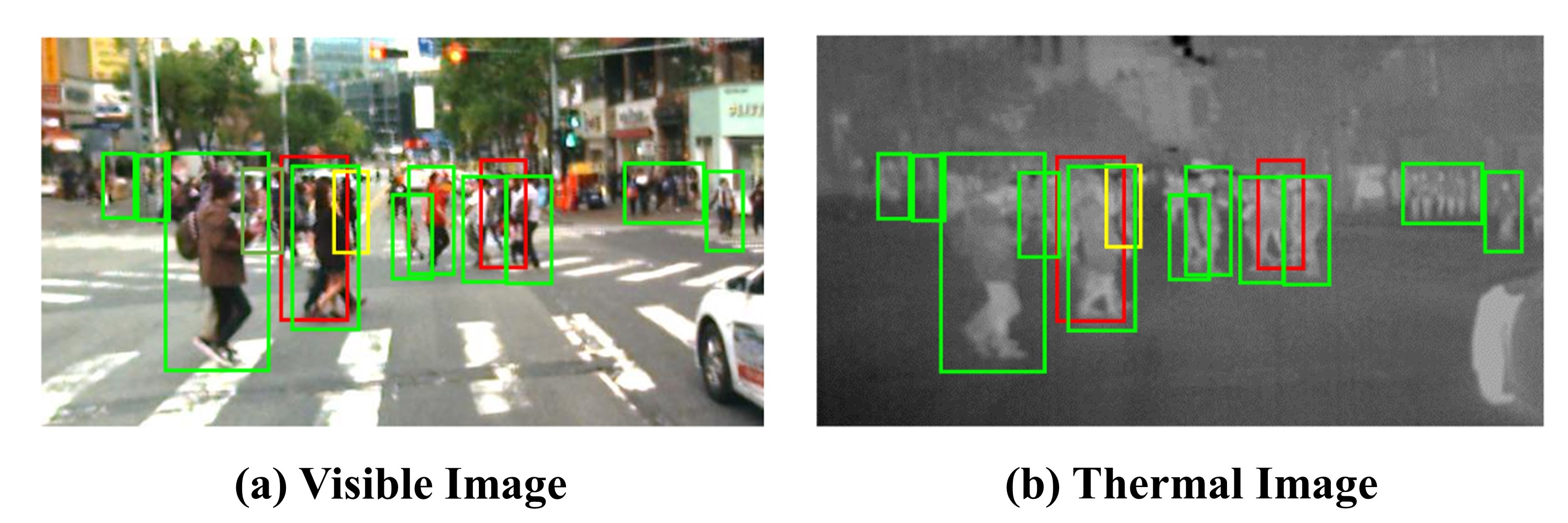}}
\caption{\ar{A dual feel camera feed (visible and thermal image pair) for multi-spectral pedestrian detection. Images are from \cite{hwang2015multispectral}.}} 
\label{fig:multi_spec}
\end{center}
\end{figure}
\ar{It is noteworthy that some handcrafted features, such as optical flow features, are complementary to deep convolutional features, which can further boost the models' ability \cite{hu2017pushing}.} 

\ar{Another interesting problem is pedestrian detection from multi-spectral images (such as RGB true-color + thermal camera feeds, as shown in Fig. \ref{fig:multi_spec}. This approach can be advantageous because infrared images provide more informative and robust features under low illuminations and extreme weather conditions. These works often require \textit{feature fusion} 
to boost the overall performance. Feature fusion can be implemented in different stages, ranging from low-level (near the input layer) to high-level (near the output layer). The authors of \cite{liu2016multispectral} have shown that fusion at the middle-level (the authors name it \textit{halfway fusion}) often achieves the best results. More recently, attention mechanisms, including channel attention and semantic attention, are applied for feature fusion \cite{zhang2019cross} to allow the network automatically learn the importance of different features. Alternatively, an illumination-aware weighting mechanism is used in \cite{li2019illumination} to learn the variation of illumination and adaptively mix the features of the visible (RGB) and thermal images. Some minor issues such as modality imbalance and weak alignment between different channels should be taken care of when fusing multiple camera feeds  \cite{zhou2020improving,zhang2019weakly}. 
A summary of recent implementations is presented in Table \ref{tab:ped_dect}. KAIST \cite{hwang2015multispectral} and CVC-14 \cite{gonzalez2016pedestrian} datasets are often used as the benchmark validation dataset for these works. More datasets for pedestrian detection are provided in Table. \ref{tab:trafficdataset-continue}.}

\begin{table*}[htbp]
\centering
\caption{Examples of object detection tasks for traffic analysis. If the dataset is not indicated, it means the current work uses the dataset generated by its authors. $\triangle$' means that this work uses fine-grained detection problem. '*' Means this work is based on YOLOv5. It is worth mentioning that YOLOv5 may not be considered as a member of YOLO family.}
\resizebox{\textwidth}{!}{
\begin{tabular}{llll} 
\toprule
\textbf{Task}                                                               & \textbf{Methods}                                                             & \textbf{Paper: [Ref] Authors (year)}                                                                & \textbf{Performance: Accuracy [Dataset]}                                                                                                                                     \\ 
\midrule

Vehicle Detection                                                  & R-CNN family                                                        & \begin{tabular}[c]{@{}l@{}} \cite{espinosa2017vehicle}Espinosa, et al.(2017)\\ \cite{wang2017evolving}Wang, et al.(2017)\\ \cite{soin2017moving}Soin, et al.(2017)\\ \cite{zhang2017application}Zhang, et al.(2017)\\ \cite{peppa2018urban}Peppa, et al.(2018)\\ \cite{yu2017model}Yu et al. (2017)$\triangle$  \end{tabular}      & \begin{tabular}[c]{@{}l@{}}70\%\\84.43\% on DETRAC dataset \cite{wen2015ua} \\100\%\\96.5\%\\97\%\\98\%\end{tabular}                                                          \\ 
\cmidrule{2-4}
                                                                   & YOLO family                                                         & \begin{tabular}[c]{@{}l@{}} \cite{sang2018improved}Sang, et al.(2018)$\triangle$\\ \cite{kim2019multi}Kim, et al.(2019) \\ \cite{kasper2021detecting} Kasper-Eulaers et al.(2021)* \\ \cite{nayak2019vision} Nayak et al.(2019)$\triangle$\end{tabular}                      & \begin{tabular}[c]{@{}l@{}}99.51\% \\85.29\%on UA-DETRAC\cite{wen2015ua} \\ 93\% car, 63\% truck front, 52\% truck back (winter condition) \\ 99.73\%\end{tabular}                                                                      \\ 
\cmidrule{2-4}
                                                                   & SSD                                                                 & \begin{tabular}[c]{@{}l@{}} \cite{zhang2019vehicle}Zhang, et al.(2019)\\ \cite{chen2020inception}Chen, et al.(2020)\\ \cite{cao2020front}Cao, et al.(2020)\\ \cite{peppa2018urban}Peppa, et al.(2018)\end{tabular}            & \begin{tabular}[c]{@{}l@{}}77.94\% mAP on UA-DETRAC\cite{wen2015ua}\\84.5\%on KITTI \\92.18\% on KITTI\\98.2\%\end{tabular}                                              \\ 
\cmidrule{2-4}
                                                                   & \begin{tabular}[c]{@{}l@{}}CNN \\ CNN    \end{tabular}                                                                & \begin{tabular}[c]{@{}l@{}}\cite{6778050}Chen, et al.(2014)   \\ \cite{zhou2016dave}Zhou et al.(2016)$\triangle$  \end{tabular}                                                                & \begin{tabular}[c]{@{}l@{}} 99.7\%   \\ 62.85\% on UTS(self), 64.44\% PASCAL VOC2007\cite{everingham2010pascal}, and 79.41\% on LISA 2010\cite{sivaraman2010general}  \end{tabular}                                                                                                                                          \\ 
\midrule

Plate Detection                                                    & R-CNN family                                                        & \cite{lee2016real}Lee, et al.(2016)                                                                   & 99.94\%                                                                                                                                           \\ 
\cmidrule{2-4}
                                                                   & YOLO family                                                         & \begin{tabular}[c]{@{}l@{}}\cite{kessentini2019two}Kessentini, et al.(2019) \\ \cite{khazaee2020real}Khazaee, et al.(2020)\\ \cite{xie2018new}Xie, et al.(2018)\\ \cite{chen2019automatic}Chen, et al.(2019)\end{tabular}        & \begin{tabular}[c]{@{}l@{}}97.67\% on GAP-LP dataset\cite{GAP-LP} 91.46\% on Radar\\97.9\%\\98.32\% on UCSD 97.38\% on PKU\cite{yuan2016robust}\\98.22\% on AOLP\end{tabular}                              \\ 
\cmidrule{2-4}
                                                                   & SSD                                                                 & \begin{tabular}[c]{@{}l@{}} \cite{ren2020implementation}Rene, et al.(2020)\\ \cite{hu2020mobilenet}Hu, et al.(2020) \\ \cite{danilenko2020license}Danilenko, et al.(2020)\end{tabular}                & \begin{tabular}[c]{@{}l@{}}92\%\\90.12\%\\94+\%\end{tabular}                                                                                          \\ 
\cmidrule{2-4}
                                                                   & \begin{tabular}[c]{@{}l@{}}CNN\\Pre-pocessing+CNN\\CNN\end{tabular} & \begin{tabular}[c]{@{}l@{}} \cite{kim2017deep}Kim, et al.(2017) \\ \cite{selmi2017deep}Selmi, et al.(2017) \\ \cite{masood2017license}Masood, et al.(2017)\end{tabular}                & \begin{tabular}[c]{@{}l@{}}98.39\% on Caltech\cite{CALTECH}\\94.8\% on Caltech\cite{CALTECH}\\99.09\% on US\cite{US_dataset} 99.64\% on EU\cite{US_dataset}\end{tabular}                                                    \\ 
\midrule
\begin{tabular}[c]{@{}l@{}}Traffic Sign \\ Detection \end{tabular} & R-CNN family                                                        & \begin{tabular}[c]{@{}l@{}} \cite{qian2016road}Qian,  et al.(2016)\\ \cite{shao2019improved}Shao, et al.(2019) \\ \cite{zhang2020cascaded}Zhang, et al.(2020)\\ \cite{zuo2017traffic}Zuo, et al.(2017)\\ \cite{wu2019traffic}Wu, et al.(2019)\\ \cite{peng2016traffic}Peng, et al.(2016)\end{tabular} & \begin{tabular}[c]{@{}l@{}}85.58\%\\69.56\% on GTSDB+CTSD\cite{yang2015towards}\\98.7\% on GTSDB\\34.49\% mAP on CCF2016\\mAP 91.75\% GTSDB\\mAP 90\% on GTSDB\end{tabular}  \\ 
\cmidrule{2-4}
                                                                   & YOLO family                                                         & \begin{tabular}[c]{@{}l@{}} \cite{zhang2017real}Zhang, et al.(2017)\\ \cite{Tai_2020}Tai, et al.(2020)\\ \cite{liu2021real}Liu et al.(2021)* \\ \cite{qin2021traffic} Qin et al.(2021)*\end{tabular}                     & \begin{tabular}[c]{@{}l@{}}96.69\% on CTSD and GTSDB\\99.1\% mAP\\ 97.2\% mAP\\ 94.3 \% mAP \end{tabular}                                                                     \\ 
\cmidrule{2-4}
                                                                   & SSD                                                                 & \begin{tabular}[c]{@{}l@{}} \cite{gao2019traffic}Gao, et al.(2019)\\ \cite{you2020traffic}You, et al.(2020)\end{tabular}                     & \begin{tabular}[c]{@{}l@{}}91.0\% on GTSDB 75\% on TT100K\end{tabular}                                                                                 \\ 
\cmidrule{2-4}
                                                                   & \begin{tabular}[c]{@{}l@{}}MSER+SVM+CNN\\FullyConv\\CNN\\CNN\end{tabular}                 & \begin{tabular}[c]{@{}l@{}}\cite{yang2015towards}Yang, et al.(2015)\\\cite{zhu2016traffic}Zhu, et al.(2016)\\\cite{wu2013traffic}Wu, et al.(2013)\\ \cite{shustanov2017cnn}Shustanov, et al.(2017)\end{tabular}                  & \begin{tabular}[c]{@{}l@{}} 98.24\% GTSDB 98.77\% CTSD\\91\% TT100K\\AUC 99.73\% “danger” , 97.62\% "mandatory" on GTSDB\\99.94\% on GTSDB\end{tabular}                                 \\
\bottomrule
\end{tabular}
}
\label{tab:objectdetection}
\end{table*}

\begin{table*}[htbp]
\centering
\caption{\ar{Examples of pedestrian detection methods with their miss rate.}}
\label{tab:ped_dect}
\resizebox{0.9\textwidth}{!}{%
\blr
\begin{tabular}{clll} \toprule
\multicolumn{1}{l}{\textbf{Task}} & \textbf{Method} & \textbf{Paper: [Ref] Authors (year)} & \textbf{Performance: Accuracy [Dataset} \\ \midrule
\multirow{10}{*}{Single-spectral} & Faster R-CNN & \cite{cai2016unified} Cai et al. (2016) & 10\% MR on Caltech \\
 & Fast R-CNN & \cite{li2017scale} Li et al. (2017) & 9.32\% MR on Caltech \\
 & CNN+Attention & \cite{lin2018graininess} Lin et al. (2018) & 7.84\% MR on Caltech \\
 & Faster R-CNN & \cite{zhang2018occlusion} Zhang et al. (2018) & 4.1\% MR on Caltech;11\% MR on CityPersons; \\
 & CNN+LSTM & \cite{wang2018pcn} Wang et al. (2018) & 8.4\% MR on Caltech \\
 & FCN & \cite{liu2019high} (2019) & 3.8\% MR on Caltech; 11.4\% MR on CityPersons \\
 & Faster R-CNN+ Attention & \cite{zhou2019ssa} (2019) & 6.27\% MR on Caltech; \\
 & Cycle-GAN & \cite{luo2020whether} (2020) & 6.37\%MR on Caltech; 9.3\% MR on CityPersons \\
 & CNN+GCN & \cite{xie2021psc} Xie et al. (2021) & 6.4\% MR on Caltech; 9.3\% MR on CityPerson \\
 & FCN+pattern-parameter matching & \cite{liu2021adaptive} et al. (2021) & 3.3\% MR on Caltech; 10.4\% MR on CityPersons \\ \midrule
\multirow{9}{*}{Multi-spectral} & Faster R-CNN+ Fusion & \cite{liu2016multispectral} Liu et al. (2016) & 25.73\% MR on KAIST \\
 & Faster R-CNN+ Fusion & \cite{li2019illumination} Li et al. (2019) & 15.73\% MR on KAIST \\
 & CNN+ ATT+ Fusion & IATDNN[34] \cite{guan2019fusion} Guan et al. (2019) & 14.93\% MR on KAIST \\
 & CNN+Attention+ Fusion & \cite{zhang2019cross} Zhang et al. (2019) & 14.12\% MR on KAIST \\
 & Faster R-CNN+ Fusion & AR-CNN[35]\cite{zhang2019weakly} Zhang et al. (2019) & 9.34\% MR on KAIST \\
 & SSD+ATT+ Fusion & MBNet 20\cite{zhou2020improving} Zhou et al. (2020) & 8.13\% MR on KAIST \\
 & CNN+memeory learning & \cite{kim2021robust} Kim et al. (2021) & 8.26\% MR on KAIST; 23\% MR on CVC-14 \\
 & CNN+ATT+Fusion & \cite{kim2021uncertainty} Kim et al. (2021) & 7.89\% MR on KAIST; 18.7\% MR on CVC-14 \\
 & YOLOv5+ATT & \cite{jiang2022attention} Jiang et al. (2022) & 7.85\% MR on KAIST \\ \bottomrule
\end{tabular}%
}
\end{table*}

\begin{table*}[h]
\centering
\caption{\ar{Evaluation metrics for different vision-based traffic analysis tasks (detection, classification, etc.)}}
\resizebox{0.7\textwidth}{!}{
\begin{tabular}{ll} 
\toprule
\textbf{Metric}                                   & \textbf{Definition}                                                                                                                                                                                                                                                                                                                                        \\ 
\midrule
Accuracy                                 & $\frac{\text{Correct predictions}}{\text{Total predictions}}$                                                                                                                                                                                                                                                                                                   \\ 
\midrule
Recall (R)                               & $\frac{TP}{TP+FN}$ ~~~ (*)                                                                                                                                                                                                                                                                                                                                \\ 
\midrule
Precision (P)                            & $\frac{TP}{TP+FP}$                                                                                                                                                                                                                                                                                                                                \\ 
\midrule
F1 score                                 & $2\times\frac{P\times R}{P+ R}$                                                                                                                                                                                                                                                                                                                   \\ 
\midrule
Pixel Accuracy (PA)                      & $\frac{TP+TN}{TP+TN+FP+FN}$                                                                                                                                                                                                                                                                                                                       \\ 
\midrule
\ar{Miss Rate (MR)}                              & \ar{$\frac{FN}{TP+FN}$}                                                                                                                                                                                                                                                                                                                                \\  \midrule
Normalized error (mean)                  & \begin{tabular}[c]{@{}l@{}} Normalized error=$\frac{|d-d_{gt}|}{d_{gt}}$. \\ $d$ is the calculated distance and \\ $d_{gt}$ is the ground truth of distance. \end{tabular}                                                                                                                                                                         \\ 
\midrule
Average Precision (AP)                   & $AP=\int_0^1 P(R)dR$ P-R CURVE (for one class)                                                                                                                                                                                                                                                                                                    \\ 
\midrule
mean Average Precision (mAP)             & $mean(AP_i)$                                                                                                                                                                                                                                                                                                                                      \\ 
\midrule
Intersection over Union (IoU)            & \begin{tabular}[c]{@{}l@{}}$\frac{B_p\cap B_{gt}}{B_p\cup B_{gt}}$\\ $B_{gt}$: ground truth bounding box\\ $B_{p}$: predicted bounding box \end{tabular}                                                                                                                                                                                          \\ 
\midrule
Disparity accuracy\cite{menze2015object}                      &  $disparity_{inlier}\%$. The disparity $\leq$ 3px/5\% is true                                                                                                                                                                                                                                                                                                                                         \\ 
\midrule
Stixel-wise percentage accuracy          & Similar to PA                                                                                                                                                                                                                                                                                                                                     \\ 
\midrule
Identification F1-Score ($ID:F1$)           & $ID:F1=\frac{2ID:TP}{2ID:TP+ID:FP+ID:FN}$                                                                                                                                                                                                                                                                                                              \\ 
\midrule
multiple object tracking accuracy (MOTA) & \begin{tabular}[c]{@{}l@{}}$MOTA=1-\frac{FN+FP+\Phi}{T}$\\ $\Phi$ denotes the number of fragmentations \end{tabular}                                                                                                                                                                                                                              \\ 
\midrule
Multi-camera Tracking Accuracy (MCTA)    & \begin{tabular}[c]{@{}l@{}}$MCTA= \underbrace{\frac{2PR}{P+R}}_{F_1} \underbrace{(1-\frac{M^w}{T^w})}_{\text{within camera}} \underbrace{(1-\frac{M^h}{T^h})}_{\text{handover}}$\\ $M^w$: within-camera identity mismatches \\ $T^w$: true within-camera detections\\ $M^h$: handover mismatches\\ $T^h$: true handover detections \end{tabular}  \\ 
\midrule
RMSE                                     & 
$RMSE=\sqrt{\sum_{i=1}^N(y_i-\hat{y}_i)^2/N}$                                                                                                                                                                                                                                                                                                           \\
\bottomrule
\end{tabular}
}

\label{tab:somemetrics}
\end{table*}

\newpage
\subsection{Visual Tracking}\label{sec:tracking}


Visual tracking refers to capturing the movement of specific objects by processing video frames. Traffic video processing by autonomous vehicles or traffic monitoring systems is perhaps the most popular application of visual tracking. With the advent of AVs and technology-assisted driving systems, this research area has become a hot topic. 
The mainstream algorithms can roughly be categorized into Correlation Filter-based Trackers (CFT) and non-CFT methods \cite{fiaz2019handcrafted}. The two challenging issues are the natural difficulty of reliable visual tracking and the lack of well-annotated 
datasets, especially for driving safety analysis.

Visual tracking methods can be categorized into discriminative methods \cite{hare2015struck,wang2010discriminative,kalal2010pn,wang2014object} and generative methods \cite{kwon2011tracking,sevilla2012distribution,liu2014visual,belagiannis2012segmentation,kwak2011learning} from the modeling standpoint. Generative tracking methods comprise the following sequential steps: (i) extract the target features to learn the appearance model, which represents the target, (ii) search through the image area to find areas that best match the model, using pattern matching. The target information carried by generative models is often richer than that of the discriminative methods. Also, generative models make it is easier to meet the evaluation criteria and the real-time requirements of target tracking when processing a massive amount of data. The key components of generative methods include target representation and target modeling. The limitations of this approach include (i) background information of the image is not fully utilized, and (ii) the appearance of the target in different video frames may include substantial randomness and diversity, which affects the stability of the model.

Discriminative methods turn the tracking problem into a classification problem, and rely on training a classifier to distinguish between the target and the background. The target area is considered the positive sample in the current frame, and the background area is the negative sample. 
Discriminative methods turn the tracking problem into a classification problem, which can simultaneously utilize the information from the target and background. ML methods can be employed to train a binary classifier to distinguish between the target (considered as the positive sample) and the background (considered as the negative samples), and can be updated online as more video frames are accumulated. The trained classifier is used to find the optimal area in the next frame. Discriminative methods are often more robust than generative models when facing appearance and environmental changes.


Correlation Filter (CF) methods and DL methods are often considered discriminant methods. In recent years, tracking methods based on correlation filtering \cite{bolme2010visual,bolme2009average,henriques2012exploiting,henriques2014high,kiani2017learning,li2014scale} have gained a lot of attention from computer vision researchers because of their fast speed and reasonable performance. 

The correlation filter is initialized by the given target in the first frame of the input. 
A classifier is trained by regressing the input features to the target Gaussian distribution. The response peak in the predicted distribution is found in the follow-up tracking step to locate the target's position. 
Correlation filtering methods, when combined with deep features and CNN architectures (e.g., R-CNN series \cite{bewley2016simple,yu2016poi,wojke2017simple}, SSD-based methods \cite{zhao2018multi}) exhibit outstanding performances and hence have gradually become a dominant approach in this field. 

In addition to CF-based trackers, some other DL frameworks, even without using correlation filters, can also achieve an excellent performance. 
More elegant methods, including  \cite{lu2017online,fang2018recurrent,maksai2018eliminating,zhu2018online,ma2018trajectory,sadeghian2017tracking} tried to directly employ sequential learning models such as Long Short Term Memory (LSTM) or Recurrent Neural Networks (RNN) in their network structures after CNN-based feature extractors to capture the temporal information of video frames that represent object motions. 

\begin{figure}[H]
\begin{center}
\centerline{\includegraphics[width=0.8\columnwidth]{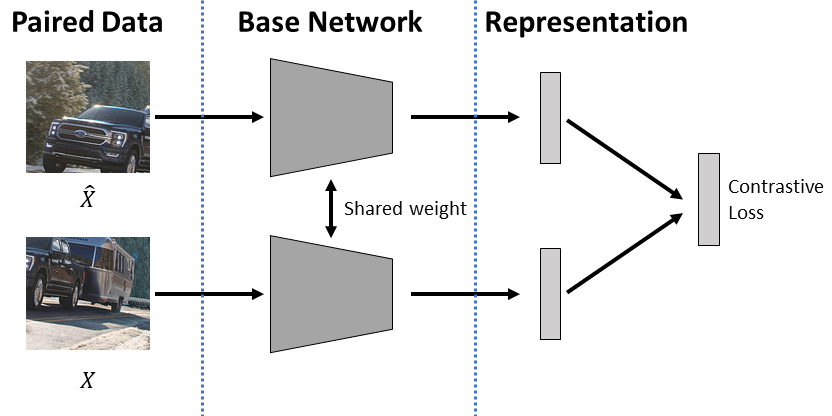}}
\caption{\ar{The architecture of basic Siamese neural network for similarity training.}}
\label{fig:SiameseNet}
\end{center}
\end{figure}

\ar{Siamese Network structure (shown in Fig. \ref{fig:SiameseNet}) offers a paradigm shift for visual tracking.} 
The core idea of Siamese networks \cite{kim2016similarity,wang2016joint,leal2016learning,son2017multi,zhou2018online,wang2019fast,li2019siamrpn++} is training twin networks to identify the similarity between two different images, such as the same object in consecutive video frames. 
These methods address both the similarity knowledge learning and the real-time operation requirements with acceptable tracking accuracy.  


Many contemporary traffic monitoring systems still tend to use conventional tracking methods such as filtering and convolution methods due to their efficiency, low complexity, and stability. For examples, methods based on Kalman filtering \cite{dinh2017development,kaur2016vehicle,o2010rear,teoh2012reliability,khalkhali2020vehicle}, Gaussian Mixture Models (GMM) \cite{chen2012vehicle,dinh2017development}, Hidden Markov Models (HMM) \cite{jazayeri2011vehicle}, and SIFT-based methods\cite{luvizon2016video,yang2016research} are used in traffic monitoring systems. However, using DL-based methods is gaining traction in recent years to perform traffic monitoring tasks due to the emergence of powerful and low-cost processing platforms, making DL methods more affordable and near real-time.

Hybrid methods that enable tracking by detection are another possibility. For instance, a DL method can be used for fast and accurate object detection, followed by a second estimator based on conventional methods, Kalman filtering, and Kanade–Lucas–Tomasi (KLT) feature tracker for tracking purposes. \ar{This approach simplifies the processing job for object detectors designed for specific tasks, such as multi-spectral pedestrian detection. }

A summary of some important implementations is presented in Table \ref{tab:trackingontraffic}. \ar{Note that datasets KITTI, and MOT 15 Challenges \cite{leal2015motchallenge} include pedestrians, in addition to vehicles. Several studies, including \cite{scheidegger2018mono,zou2019multi,fernandez2019real}, are specifically designed to improve the safety of humans.}

\begin{table*}[h]
\centering
\caption{Some traffic-related visual tracking method that use deep learning. The default dataset is generated by the author of each work. The default performance metric is accuracy unless specified otherwise.}
\resizebox{0.9\textwidth}{!}{
\begin{tabular}{lll} 
\toprule
\textbf{Paper: [Ref] Authors (year)} & \textbf{Methods}                        & \textbf{Performance}                                                                                                             \\ 
\midrule
\cite{qiu2018kestrel}Qiu, et al.(2018)     & YOLO+KLT                       & 76.4\% Recall 88.2\% Precision                                                                                               \\ 
\midrule
\cite{lopez2019boosting}L{\'o}pez-Sastre, et al.(2019)    & Faster R-CNN                   & \begin{tabular}[c]{@{}l@{}}30.5AP on subset:M-30 66.2\%AP on subset:M-30-HD~\\38.1\%AP on subset:Urban1 of GRAM-RTM\end{tabular}  \\ 

\midrule
\cite{scheidegger2018mono}Scheidegger, et al.(2018)     & CNN+PMBM filter\cite{garcia2018poisson}                & 80.04\% MOTA on KITTI                                                                                                     \\ 
\midrule
\cite{zou2019multi}Zou, et al.(2019)     & Siamese network+SPP+MDP        & \begin{tabular}[c]{@{}l@{}}75.29\% MOTA campus~\\76.06\% MOTAurban~\\78.14\% MOTA highway~\\on KITTI\end{tabular}                \\ 
\midrule
\cite{li2019spatio}Li, et al.(2019)     & FPN+tracking loss              & 83.2\% IDF1 on nivida AI city                                                                                                 \\ 
\midrule
\cite{nikodem2020multi}Nikodem, et al.(2020)     & hourglass                      & \begin{tabular}[c]{@{}l@{}}MOTA 97+\%\\MCTA 91+\%\end{tabular}                                                              \\ 
\midrule
\cite{zhao2018autonomous}zhao, et al.(2018)     & SSD+dual Kalman filters        & Shown in the form of figures                                                                                            \\ 
\midrule
\cite{wang2019orientation}Wang, et al.(2019)     & YOLOv3+SIFT                    & 81.3\% MOTA                                                                                                                  \\ 
\midrule
\cite{usmankhujaev2020real}Usmankhujaev, et al.(2020)    & yoloV3+kalman filter           & \begin{tabular}[c]{@{}l@{}}100\% Daytime~Wrong cases detection\\89.83\% Daytime(flipped) Wrong cases detection~\\86.11\% Nighttime(flipped) Wrong cases detection\end{tabular}                 \\ 
\midrule
\cite{kwan2018compressive}Kwan, et al.(2018)    & ResNet                         & For compressive measurements                                                                                            \\ 
\midrule
\cite{fernandez2019real}Fern{\'a}ndez-Sanjurjo,et al.(2019)    & CNN detector+DCF+Kalman filter & 86.96\% MOTA on MOT 15\cite{leal2015motchallenge}                                                                                                    \\
\bottomrule
\end{tabular}
}
\label{tab:trackingontraffic}
\end{table*}

\subsection{Semantic Segmentation}\label{sec:Semantic Seg}

\begin{figure}[H]
\begin{center}
\centerline{\includegraphics[width=0.99\columnwidth]{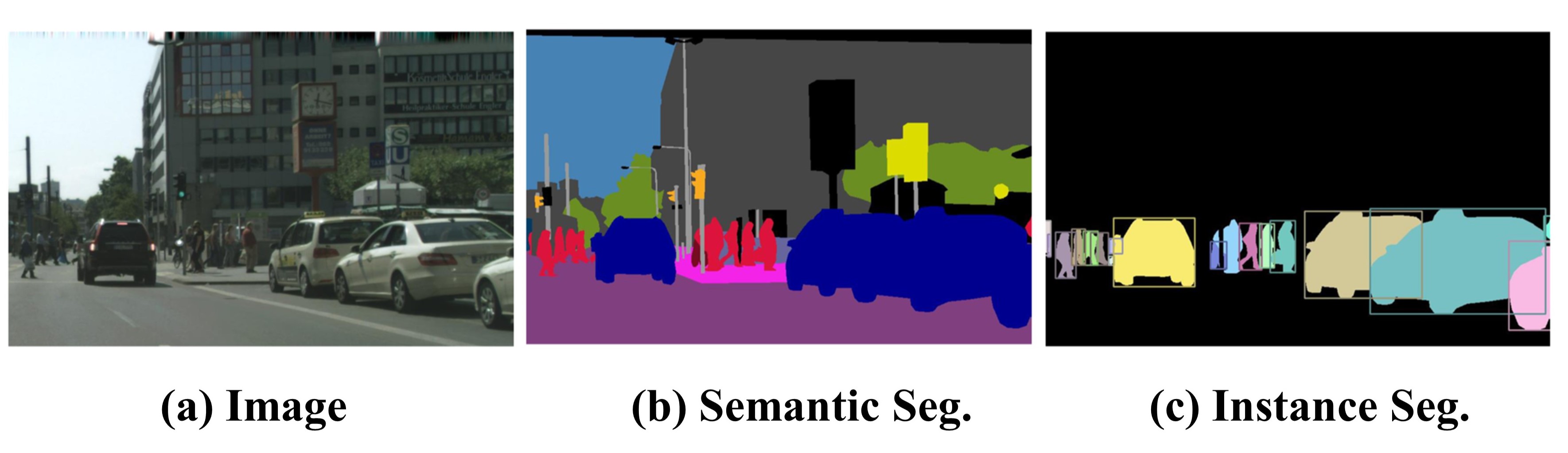}}  
\caption{A example of different types of segmentation. Source from \cite{kirillov2019panoptic}.} 
\label{fig:diffSeg}  
\end{center}
\end{figure}

Semantic segmentation, where objects of different types are separated, can be considered the heart of many video processing tasks. For instance, vehicle detection, vehicle tracking, and environment perception in a crowded environment with interlaced and overlapping objects can be powered by semantic segmentation when regular segmentation methods fail to separate objects from complex backgrounds. \ar{An example of semantic segmentation is shown in Fig. \ref{fig:diffSeg}(b).} The purpose of semantic segmentation is to label each pixel of an image to represent different categories (e.g., cars, pedestrians, roadside infrastructures, traffic signs, etc.). For instance, semantic segmentation can be employed by an autonomous vehicle for background modeling, identifying road boundaries and free spaces, and detecting lane markings and traffic signs. Semantic segmentation can also be used by an external traffic monitoring system for analyzing the behaviors of human-driven and self-driving vehicles in specific zones and times. To avoid reliance on massive data collection and expensive annotations, semi-supervised and weakly-supervised learning methods \cite{wei2018revisiting,hong2015decoupled,souly2017semi,hung2018adversarial} are developed for low-cost implementation with reasonable performance. 


Early works tended to deploy existing classification algorithms at the patch level for semantic segmentation. Since 2014, Fully Convolutional Network (FCN)\cite{long2015fully} was introduced that allows spatially dense prediction tasks by translating famous DL architectures such as AlexNet, VGG net, and GoogLeNet into fully convolutional network architectures.  
Afterward, many upgraded architectures such as U-net\cite{UNET} and SegNet\cite{badrinarayanan2017segnet} were proposed, which build upon the concept of FCN and utilize auto-encoder architectures for semantic segmentation with small training datasets using data augmentation methods. This architecture is further updated to multi-stage auto-encoder networks in \cite{cheng2019spgnet}.
In some work\cite{yu2015multi,chen2017deeplab,paszke2016enet,chen2017rethinking}, atrous/dilated convolution architectures are used to keep spatial resolution and expand the receptive fields. The authors of \cite{peng2017large} expanded the receptive fields by employing large kernel with its proposed global convolution.

\begin{figure}[H]
\begin{center}
\centerline{\includegraphics[width=0.6\columnwidth]{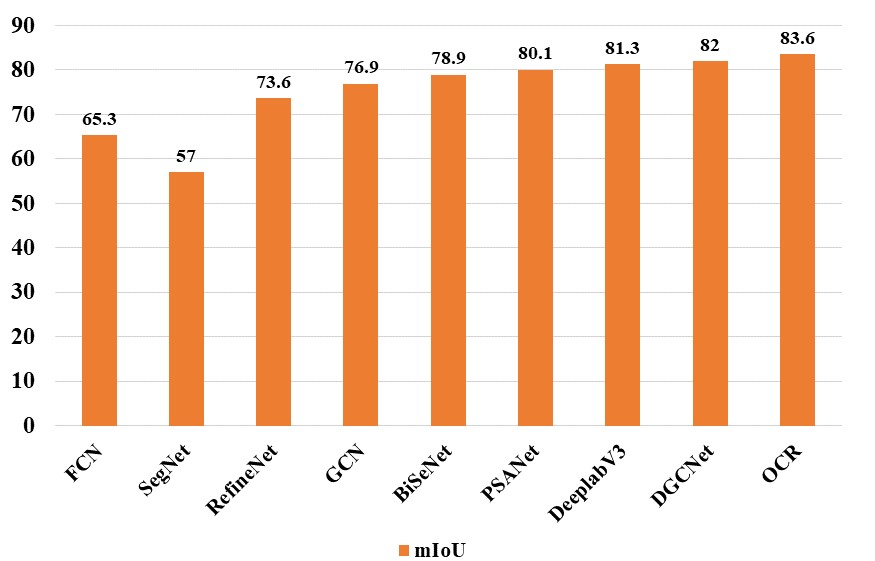}}
\caption{Semantic segmentation models perform on Cityscapes.}
\label{fig:semantic_perf}
\end{center}
\end{figure}

In \cite{liu2015parsenet,pinheiro2016learning,lin2017refinenet}, a feature fusion method is deployed that allows the framework to learn global features merged with more local features. For these methods, Conditional Random Fields can be used to enhance the output. Other works \cite{visin2016reseg,li2016lstm} are based on a Recurrent Neural Network (RNN) structure that can better tackle sequence-related tasks. Additional models including\cite{zhang2019dual} based on graph convolutional network, \cite{zhao2017pyramid} based on pyramid pooling model, and \cite{yuan2020object} based on learning relations between the object region and pixels have competitive performance. Furthermore,  \cite{qi2017pointnet,qi2017pointnet++,huang2016point} focused on semantic segmentation with 3D point cloud data, which have great potentials for autonomous driving and traffic safety analysis. More specifically, the methods \cite{qi2017pointnet,qi2017pointnet++} are evaluated by indoor scene dataset, while \cite{huang2016point} is evaluated by urban scene captured by scanners, something potentially more relevant for traffic analysis. The performance of some models is shown in Fig. \ref{fig:semantic_perf}.

The applications of semantic segmentation frequently appear in the world of AVs. Some examples include (i) scene understanding, which involves understanding the traffic environment with road users, (ii) free space estimation to determine the available spaces on the road that a vehicle is allowed to use with no collisions, and (iii) Stixel representation, which assigns each pixel with a 3D depth information. A summary of related works is presented in Table \ref{tab:semanticwork}. 

\begin{table*}[ht]
\centering
\caption{Some semantic segmentation works in the traffic field.}
\resizebox{\textwidth}{!}{
\begin{tabular}{llll} 
\toprule
\textbf{Task}                  & \textbf{Paper: [Ref] Authors (year)}                                                           & \textbf{Methods}                                                                                                                                                & \textbf{Performance}                                                                                                                                                           \\ 
\midrule
Scene understanding   & \begin{tabular}[c]{@{}l@{}}\cite{romera2017erfnet}Romera, et al(2017)\\ \cite{lyu2019esnet}Lyu et al.(2019)\\  \cite{deng2017cnn}Deng, et al.(2017) \\ \cite{saez2018cnn}Sáez, et al.(2018)\\ \cite{kendall2015bayesian}Kendall, et al.\end{tabular}      & \begin{tabular}[c]{@{}l@{}}Autoencoder\\ Edge Detection Network+Fusion\\ CNN+pyramid pooling module\\ Autoencoder\\ Bayesian Autoencoder \end{tabular} & \begin{tabular}[c]{@{}l@{}}pixel accuracy$>$95\% Cityscapes\\63.2\% mIoU Cityscapes\\54.5\% mIoU Cityscapes\\59.3\% mIoU Cityscapes\\63.1\% mIoU CamVid\end{tabular}            \\ 
\midrule
Free space estimation & \begin{tabular}[c]{@{}l@{}} \cite{ohgushi2020road}Ohgushi, et al.(2020)\\ \cite{hua2019small}Hua, et al.(2019)\\\\ \cite{levi2015stixelnet}Levi, et al.(2015)\\ \\ \cite{deepika2017obstacle}Deepika, et al.(2017)\end{tabular} & \begin{tabular}[c]{@{}l@{}}Autoencoder\\ Autoencoder+fusion+optical flow\\ \\ CNN\\ \\ SegNet \end{tabular}                                          & \begin{tabular}[c]{@{}l@{}}21.9\% mIoU\\Path planning accuracy\\0.15m Indoor and 0.27m Outdoor\\0.87 AUC~\\89.12\% maxF SEGMENTATION on KITTI\\0.9667 IoU\end{tabular}  \\ 
\midrule
Stixel representation & \begin{tabular}[c]{@{}l@{}}\cite{schneider2016semantic}Schneider,  et al.(2016)\\\cite{cordts2017stixel}Cordts,  et al. (2017)\end{tabular}                  & \begin{tabular}[c]{@{}l@{}}FCN+SGM\\ FCN+SGM+graphical model \end{tabular}                                                                             & \begin{tabular}[c]{@{}l@{}}7.8 Disparity Error on KITTI 15.2\% on Ladicky\\83.1\% exact Disparity accuracy\end{tabular}                                                   \\
\bottomrule
\end{tabular}}

\label{tab:semanticwork}
\end{table*}

\subsection{Instance Segmentation}\label{sec:Instance Segmentation Problem definition}
Instance segmentation, which deals with detecting and delineating distinct objects of interest in images and video frames, can be considered as one of the most difficult tasks in computer vision \ar{(Fig. \ref{fig:diffSeg}(c))}. It goes one step beyond semantic segmentation and not only labels the pixels based on their object categories, but also distinguishes between different object instances of the same type. This is of crucial importance for traffic imagery analysis in dense zones, where different objects (e.g., vehicles, pedestrians) have to be identified and located in video frames for optimal decision making by AVs, or to extract safety metrics by monitoring systems. 
In contrast to semantic segmentation, instance segmentation only needs to find the edge contour of the object of interest with no need for bounding boxes, hence it can realize a more accurate object detection when assessing the behaviors of vehicles.
Building a reliable and real-time method, for instance, segmentation, especially for crowded zones and under highly distorted traffic videos (e.g., in rainy and cloudy weather conditions), can be challenging.

Instance segmentation can be divided into one-stage and two-stage methods. Two-stage methods often require generating region or object proposals followed by a classification-based segmentation performed over the features extracted from the selected regions or bounding boxes around object proposals. To generate region proposal, \cite{hariharan2014simultaneous,dai2016MNC,he2017mask} predict a bounding box for each instance, while \cite{pinheiro2015learning,pinheiro2016learning,dai2016instance,li2017fully}  develop pixel-wise coarse segmentation masks. 

Since instance segmentation partitions the image by masking the detected objects and associating each pixel with a distinct object, some two-stage detectors such as Faster R-CNN \cite{ren2015faster}) can execute instance segmentation task after some post-processing, e.g., by adding a branch for mask predictions.

\begin{figure}[H]
\begin{center}
\centerline{\includegraphics[width=1\columnwidth]{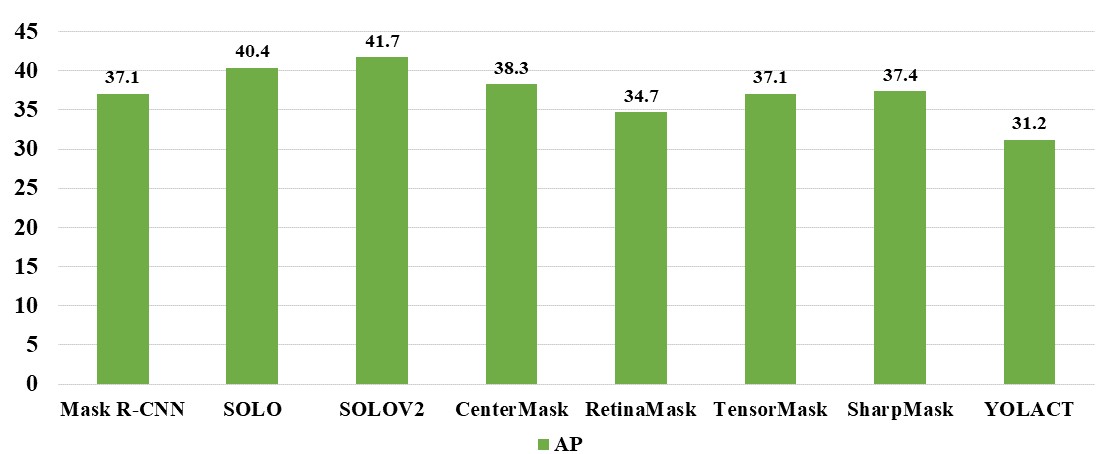}}
\caption{Instance segmentation models perform on MS COCO test-dev }
\label{fig:instance_perf}
\end{center}
\end{figure}

One-stage instance segmentation methods do not utilize a separate stage for generating region proposals; rather, they apply the segmentation directly to the original images. Some methods \cite{bolya2019yolact,bolya2019yolact++,wang2019solo,wang2020solov2} inspired by one-stage detectors (such as YOLO\cite{redmon2017yolo9000}) directly predict bounding boxes. However, these anchor-based methods heavily rely on predefined anchors, which may be affected by many factors such as the predefined anchor boxes' aspect ratio and scales. Another approach is developing anchor-free methods using dense prediction or centerpoint/keypoints. For instance, \cite{chen2020blendmask,lee2020centermask,xie2020polarmask} relied on FCOS\cite{tian2019fcos} as their dense prediction detector, while ExtremeNet\cite{zhou2019bottom} is a keypoint-based detector that can roughly perform the segmentation task. The difference between the accuracy of one-stage and two-stage methods is not as great as one may expect. Indeed, recent one-stage methods such as CenterMask\cite{lee2020centermask} can perform real-time inference with accuracy as high as two-stage methods, or even better. The performance of some instance segmentation models is shown in Fig.\ref{fig:instance_perf}.

In the context of traffic analysis, instance segmentation can be used not only to identify and locate vehicles but also to obtain detailed information about the vehicle, such as a vehicle's class, number of axles, 3D bounding box, dimension, etc. It can be used as part of the processing to identify road lanes, traffic volume, etc. For instance, one may consider each lane as an instance. Some famous methods that utilize instance segmentation for traffic image analysis are presented in Table \ref{tab:instancework}.

\begin{table*}[ht]
\centering
\caption{Some instance segmentation methods used for traffic imagery analysis. If the dataset is not indicated, it means that a proprietary dataset generated by the authors is used.}
\resizebox{\textwidth}{!}{

\begin{tabular}{llll} 
\toprule
\textbf{Task }                                                                               & \textbf{[REF] authors (Year)}                                                     & \textbf{Methods}                                                                                  &\textbf{Performance}                                                                                                                                                                                                                   \\ 
\midrule
\begin{tabular}[c]{@{}l@{}}Obtain comprehensive\\ vehicle information \end{tabular} & \begin{tabular}[c]{@{}l@{}}\cite{zhang2015monocular}Zhang, et al.(2015)\\~\\ \cite{mou2018vehicle}Mou, et al.(2018)\\ \cite{zhang2020traffic}Zhang, et al.(2020)\\ \cite{huang2018measuring}Huang, et al.(2018) \end{tabular} & \begin{tabular}[c]{@{}l@{}}CNN+MRF\\ ~\\ ResFCN\\ Mask R-CNN\\ Mask R-CNN \end{tabular} & \begin{tabular}[c]{@{}l@{}}59.0\% object recall,\\83.1\% of the randomly sampled foreground pixel pairs ordered on KITTI~\\95.87\% F1 score on ISPRS\\97\% vehicle types accuracy\\1.333 front 4.698 side-way KITTI\end{tabular}  \\ 
\midrule
Lane detection                                                                      & \begin{tabular}[c]{@{}l@{}} \cite{neven2018towards}Neven, et al.(2018)\\\cite{roberts2018dataset}Roberts, et al.(2018)\\ ~ \end{tabular}        & \begin{tabular}[c]{@{}l@{}}multi-task CNN\\ SegNet \\~\end{tabular}                     & \begin{tabular}[c]{@{}l@{}}96.4\% accruacy\\IoU Mapillary 82.9\% CityScapes 85.2\%\\KITTI 83.8\% author's dataset 95\%\end{tabular}                                                                                                   \\ 
\midrule
3D reconstruction                                                                   & \cite{hadi2020semantic}Hadi, et al.(2020)                                                         & Mask R-CNN                                                                               & \begin{tabular}[c]{@{}l@{}}AP@50\% 8.862\% CityScapes 43.949\%\\IDD\cite{varma2019idd} 5.643\% WildDash\cite{zendel2018wilddash}\end{tabular}                                                                                                                                    \\
\bottomrule
\end{tabular}
}

\label{tab:instancework}
\end{table*}




\subsection{Video-based Event Recognition } \label{sec:event_detect}

Video-based event recognition extends the role of DL methods from object detection and identification paradigms into a more intricate problem of understanding events. It provides endless possibilities to explore the interactions among objects in an interactive environment rather than focusing on disjoint object-based tasks. Indeed, without event detection and analysis, the majority of video information remains unexploited and underutilized. Note that many safety metrics relate to the interactions of vehicles with one another and with the environment. For instance, traffic sign interpretation can be recast as an image-based object classification problem, while more intricate violations of traffic safety such as unsafe lane changing behavior without signaling require modeling interactions between vehicles and their surrounding environments. 
Such challenging problems are still in their infancy stages and apparently require heavy investment by the research community. In essence, the event recognition problem is also related to another well-investigated problem of video-based human action recognition, and similar tools and algorithms can be adopted here.

An alternative method of event analysis is using conventional analysis approaches by extracting object-based information and manually feeding them into statistical and reasoning models such as Hidden Markov Models (HMM) for safety analysis. However, with the recent advances in developing powerful DL methods, they enable a more automated and direct way of evaluating the behaviors of involved vehicles. The most naive way of event recognition can be realized by extracting static features from video frames using methods like SIFT \cite{lowe2004distinctive}, SURF \cite{bay2006surf}, the Local Binary Patterns (LBP) \cite{ojala2002multiresolution,tang2013combining}, HOG detector \cite{dalal2005histograms}, Binary Robust Invariant Scalable Keypoints (BRISK) \cite{leutenegger2011brisk}, Features from Accelerated Segment Test (FAST) \cite{rosten2006machine,viswanathan2009features} and GIST (a very low-dimensional scene representation) \cite{friedman1979framing,oliva2001modeling}, and then performing object detection followed by a time-series analysis for even recognition. 
Some other methods combine the feature extraction and time-series analysis stage into directly extracting temporal features using methods like motion spatio-temporal features (Motion SIFT (MoSIFT) \cite{chen2009mosift}, Spatio-Temporal Interest Points (STIP) \cite{laptev2005space}, and Dense trajectories \cite{wang2011action,wang2013action}), then perform the classification task. 

More contemporary event detection algorithms benefit from DL methods to automate this process, and a dominant method is directly applying a 3D convolutional network to process non-anomalous frames \cite{hasan2016learning}. Another approach is deploying CNN to extract spatial features and then performing sequential analysis using methods such as  RNN/LSTM \cite{feng2016deep,jiang2016automatic,wang2018abnormal,chong2017abnormal,medel2016anomaly,liu2019lstm,feng2020spatio} to preserve the temporal features.

Most of these methods use supervised learning methods to detect a set of predefined events. Therefore, developing generic methods for understanding safety risks from driving profiles and tackling unseen types of safety violations has a long way ahead.

Event recognition is of particular importance for traffic safety analysis since it can be used for detecting abnormal events and traffic violations and their associations with crash rates \cite{giannakeris2018speed}, car behavior analysis \cite{zhou2016spatial,franklin2020traffic}, and pedestrians' crossing identification \cite{xu2015learning,medel2016anomaly,xu2017detecting}. 
Recently, some works\cite{anno2019gan,ravanbakhsh2019training,nguyen2020anomaly} try to predict anomaly actions using Generative Adversarial Networks (GAN). The core idea is predicting the future video frames for a normal user with rational behavior from the history of normal sequences to identify severe abnormalities by comparing the observed video frame against the anticipated one.

For vehicle-level analysis, when the goal is detecting plain and simple events, conventional methods achieve a reasonable performance. For instance, \cite{akoz2014traffic} uses HMM to detect traffic abnormality, and \cite{kaviani2015automatic} deployed a topic model to recognize crashes from surveillance videos. However, using DL methods can be used to analyze more complex events and achieve higher performance records. 
Some generic platforms like Rekall \cite{fu2019rekall}, provide a query-based approach to translate the challenging task of event recognition into a sequence of object detection and classification problems.

We summarize important DL-based event recognition methods in Table \ref{tab:abnormaldetect}, and present some popular datasets for event recognition tasks in Table \ref{tab:trafficdataset-continue}. One observation is that unsupervised, semi-supervised, and self-supervised models are becoming more prevalent in recent works \cite{giannakeris2018speed,yao2019unsupervised,nguyen2020anomaly} to mitigate the costly and tedious job of video annotation and simplifies volume video processing. 

\begin{table*}[h]
\centering
\caption{Some recent DL works for video-based traffic abnormal event detection. Unless otherwise specified, the non-indicated numbers is the accuracy.}
\resizebox{0.9\textwidth}{!}{%
\begin{tabular}{lll} 
\toprule
\textbf{Paper: [Ref] Authors (year)} & \textbf{Methods}                                                                  & \textbf{Performance}                                         \\ 
\midrule
\cite{nguyen2020anomaly}Nguyen, et al.(2020)     & GAN                                                                      & F1 score 0.9412 AI City Challenge 2019              \\ 
\midrule
\cite{yao2019unsupervised}Yao, et al.(2019)     & \begin{tabular}[c]{@{}l@{}}Auto Encoder-Decoder\\ with GRU \end{tabular} & 60.1\% AUC on A3D Dataset                             \\ 
\midrule
\cite{tian2019automatic}Tian, et al.(2019)     & YOLO-CA                                                                  & 90.02\% AP CAD-CVIS                                   \\ 
\midrule
\cite{kim2020pre}Kim, et al.(2020)     & 3D conv                                                                  & 82\%                                               \\ 
\midrule
\cite{ijjina2019computer}Ijjina, et al.(2019)     & 3D conv                                                                  & 71\%                                                  \\ 
\midrule
\cite{shah2018cadp}Shah, et al.(2018)      & \begin{tabular}[c]{@{}l@{}}DSA-RNN\cite{chan2016anticipating}\\ Faster R-CNN \end{tabular}          & 47.25\% on CADP                                       \\ 
\midrule
\cite{srinivasan2020novel}Srinivasan, et al.(2020)      & Detection Transformer\cite{carion2020end}                                                    & 78.2\% on CADP                                        \\ 
\midrule
\cite{suzuki2018anticipating}Suzuki, et al.(2018)     & QRNN+DeCAF\cite{donahue2014decaf}                                                                & 99.1\% mAP                                            \\ 
\midrule
\cite{giannakeris2018speed}Giannakeris, et al.(2018)     & Faster RCNN                                                              & F1-score 0.33 RMSE 227 on NVIDIA CITY Track 2       \\ 
\midrule
\cite{arceda2018fast}Arceda, et al.(2018)    & YOLO+ViF\cite{hassner2012violent}+SVM                                                             & 89\%                                                  \\ 
\midrule
\cite{biradar2019challenges}Biradar, et al.(2019)   & YOLOv2+ CNN                                                              & F1-score 0.3838 RMSE 93.61 on NVDIA CITY Track 3    \\ 
\midrule
\cite{xu2018dual}Xu, et al.(2018)    & Mask-RCNN+ ResCNN                                                        & F1-score 0.8649 RMSE 3.6152 on NVIDIA CITY Track 2  \\ 
\midrule
\cite{doshi2020fast}Doshi, et al.(2020)    & YOLO+KNN+K-means                                                         & F1-score 0.5926 RMSE 8.2386 on NVIDIA CITY track 4  \\ 
\midrule
\cite{zhou2016spatial}Zhou, et al.(2016)    & 3D Conv                                                                   & 95.2\% on U-turn dataset                              \\ 
\midrule
\cite{franklin2020traffic}Franklin, et al.(2020)    & YOLOv3                                                                   & 100\% and 95.34\% for input video 1 and video 2         \\
\bottomrule
\end{tabular}
}

\label{tab:abnormaldetect}
\end{table*}

\ar{These pure data-driven methods often focus on detection problems. Alternatively, other automated video-based road safety analysis frameworks exist that use ML/DL methods to extract information about the road users from RSU videos and perform safety analysis using domain knowledge. An important class of such studies is extracting surrogate safety measures. For example, the authors of \cite{st2015large} collected large-scale data in about 40 roundabout weaving zones. They track the vehicles using the Kanade–Lucas–Tomasi (KLT)-based tracker \cite{saunier2006feature}, then extract a set of surrogate measures, such as speed, individual interaction measurements, and Time-To-Collision (TTC), for each vehicle. Performing correlation analysis between the surrogate measures and complementary data (such as design geometry and environment attributes) reveals insightful relations about traffic safety. \ar{We also note that \cite{fu2019investigating,noh2022analyzing} focus on the analysis of pedestrian-vehicle interactions by studying the secondary interaction and collision risk predictions. These works are worthy of high attention. Although these sorts of analyses require extracting more complicated patterns from traffic videos, they can be helpful for enhancing the real-world traffic safety.} 
A summary of such frameworks is presented in Table \ref{tab:automated_video}. We note that most of these frameworks use conventional tracking algorithms, which yield reasonable performance. 
However, more advanced hybrid detection and tracking algorithms can be developed where DL networks are used for fast and accurate object detection, followed by a second estimator based on the conventional methods, as discussed at the end of Section \ref{sec:tracking}). For instance, one may combine YOLOv5 and DeepSort \cite{wojke2017simple} for precise vehicle tracking \cite{chen2022network}}.

\begin{table*}[h]
\centering
\caption{\ar{Automated video-based road safety analysis frameworks.}}
\label{tab:automated_video}
\resizebox{\textwidth}{!}{%
\blr
\begin{tabular}{llll} \toprule
\textbf{REF]   authors (Year)} & \multicolumn{1}{c}{\textbf{\begin{tabular}[c]{@{}c@{}}Video Process\\      Method\end{tabular}}} & \multicolumn{1}{c}{\textbf{\begin{tabular}[c]{@{}c@{}}Analysis \\      Method\end{tabular}}} & \textbf{Description} \\ \midrule
\cite{st2015large} St-Aubin et al. (2015) & KLT &  & Dataset Collection \\ \midrule
\cite{zangenehpour2016signalized}   Zangenehpour et al. (2016) & \begin{tabular}[c]{@{}l@{}}KLT for tracking;\\      Bayesian classification for seg.\cite{zangenehpour2015automated};\end{tabular} & \begin{tabular}[c]{@{}l@{}}Regression Analysis;\\      Rank correlation analysis;\end{tabular} & Cyclist and motor-vehicle interactions \\ \midrule
\cite{lu2016video} Lu et al. (2016) & KLT & \begin{tabular}[c]{@{}l@{}}Sensitivity analysis;\\      Golden section search\end{tabular} & Calibrate car-following parameters in VISSIM simulation; \\ \midrule
\cite{mohamed2018impact} Mohamed et al.   (2018) & \begin{tabular}[c]{@{}l@{}}KLT;\\      Kinematic method\\       Motion pattern matching (MPM) by   \\      the longest common sub-sequence (LCSS)\end{tabular} & \begin{tabular}[c]{@{}l@{}}Distribution/frequency  analysis;\\      Clustering\end{tabular} & \begin{tabular}[c]{@{}l@{}}Study on  impact of motion prediction methods\\       on safety indicators\end{tabular} \\ \midrule
\cite{xu2018automated} Xu et al. (2018) & \begin{tabular}[c]{@{}l@{}}YOLOv2;\\      NMS\end{tabular} &  & Database building for pedestrian   analysis \\ \midrule
\cite{battiato2018board} Battiato et al.   (2018) & \begin{tabular}[c]{@{}l@{}}Kernel-based object tracking   \cite{comaniciu2003kernel};\\      HOG\end{tabular} & Bayesian classification & Risk analysis from on-board video data \\ \midrule
\cite{fu2019investigating} Fu et al.   (2019) & BriskLUMINA 0.1 \cite{transoft_solutions} & \begin{tabular}[c]{@{}l@{}}Statistic Analysis;\\      Correlation analysis\end{tabular} & Secondary Pedestrian-vehicle    interactions \\ \midrule
\cite{xie2019mining} Xie et al. (2019) & \begin{tabular}[c]{@{}l@{}}KLT;\\      rPCA;\\      Dirichlet process Gaussian mixture model (DPGMM)\end{tabular} & HMM & model the rear-end conflicts \\ \midrule
\cite{chen2020conflict} Chen et al.   (2020) & \begin{tabular}[c]{@{}l@{}}HOG;\\      SVM;\\      Kalman Filter;\\      Adaptive Gaussian Mixture model\end{tabular} & Distribution analysis & \begin{tabular}[c]{@{}l@{}}Safety space analysis;\\       Data collected by UAV\end{tabular} \\ \midrule
\cite{yang2021proactive} Yang et al.   (2021) & \begin{tabular}[c]{@{}l@{}}KLT;\\      Kalman Filter\end{tabular} & Functional Data Analysis & \begin{tabular}[c]{@{}l@{}}Anomalies detection;\\       Data collected by UAV\end{tabular} \\ \midrule
\cite{noh2022analyzing} Noh et al. (2022) & \begin{tabular}[c]{@{}l@{}}Mask R-CNN;\\      DeepSort\end{tabular} & LSTM & Pedestrian-vehicle collision risk   prediction \\ \midrule
\cite{chen2022network} Chen et al. (2022) & \begin{tabular}[c]{@{}l@{}}YOLOv5;\\      DeepSort\end{tabular} & \begin{tabular}[c]{@{}l@{}}Regression Analysis;\\      Rank correlation analysis;\\      Coalition game theory\end{tabular} & Traffic risk analysis \\ \bottomrule
\end{tabular}%
}
\end{table*}

\subsection{Sensor Information Processing} \label{sec:sensor}
It is notable that there are several studies and datasets devoted to sensor information analysis. Different types of commonly used sensors were provided previously in section \ref{sec:hardware}. Of particular and increasing interest is the point cloud mapping collected from LiDARs, not just because they offer more accurate distance measurement, but also they are considered as separate and independent sensing of the environment to that of videos to ensure accurate perception thus road safety. 3D point cloud mapping from LiDAR is different from grid-based 2D images, thus different treatment strategies are pursued. LiDAR point cloud semantic segmentation works use deep neural networks, initially treating point cloud as construct graph \cite{simonovsky2017dynamic}, followed by the development of multi-layer perceptrons to learn from raw cloud data directly \cite{qi2017pointnet,qi2017pointnet++}. More recently, the spherical projection has been employed to map LiDAR sequential scans to depth images, and improved segmentation \cite{wu2018squeezeseg,wu2019squeezesegv2,milioto2019rangenet++,xiang2019novel}. Since the focus of this paper is video-processing for traffic safety analysis, we refer the interested readers to \cite{nellore2016survey,bernas2018survey,won2020intelligent,ahangar2021survey}.

\subsection{Network-level Analysis} \label{sec:network}

Traffic flow problems can be formulated as a network of mobile nodes, and studies on individual crash analysis can be extended to the more complex setup of network-level analysis. 

There exist a few research paradigms that consider network-level analysis. One bold example is the transportation network design and related family of problems. 
Transportation network design belongs to the category of operations research and can be divided into the Road Network Design Problem (RNDP) and Service Network Design Problem (SNDP) according to their features and functions\cite{farahani2013review}. The works \cite{yang1998models,szeto2015sustainable,hosseininasab2015integration} of RNDP aim to optimize the performance of urban networks according to some criteria such as topology, capacities, and flow accessibility. The works \cite{fan2006optimal,yan2013robust,cancela2015mathematical,liu2016capacitated,di2018transportation} of SNDP aim to address the planning of operations for freight transportation carriers, such as station locations, route planning, and operation frequency. 

Traffic prediction studies often employ statistical techniques (such as Kalman filtering \cite{guo2014adaptive}, hidden Markov model \cite{qi2014hidden}, Bayesian interference \cite{wang2014new}) and DL  (such as LSTM, CNN) methods \cite{ma2015long,cui2018deep,du2018hybrid,wang2019traffic} to infer the network state and produce optimal strategies for different conditions. 

Network-level analysis can be used for traffic safety analysis as well. Some key objectives would include finding correlations between traffic flow, safety metrics geo-maps, and crash rates. For example, one may expect a direct relation between the traffic composition (density and variety of vehicles) and the number of crashes at different parts of highways. One may also expect relations between the traffic flow and crash rates of nearby intersections or segments. Network-level analysis can shed light on these highly unexplored research areas.

There exist four general approaches for network-level analysis \cite{mannering2020big} including traditional statistical models (e.g., \cite{part2010highway}), endogeneity/heterogeneity models (e.g., \cite{bhat2014count,mannering2016unobserved}), data-driven methods \cite{arbabzadeh2017data,cheng2018big}, and causal inference models. Furthermore, we note that some models\cite{lord2010extension,aguero2008analysis} exploit previously collected crash data with road information (such as Average Daily Traffic (ADT), lane width, speed limited, shoulder width) to estimate crash frequencies at intersections or segments. Some models \cite{karlaftis2002effects,othman2009identifying} explore risk probabilities by processing geometric features. Recently, new models\cite{yu2013multi,ahmed2013data,xu2014using,yu2020convolutional,sun2015dynamic,sameen2017severity} 
take advantage of advanced sensors techniques and high-performance computation frameworks to infer real-time crash risks and take proactive strategies. Safety analysis frameworks can perform integrative analysis by incorporating different static and dynamic data modalities (i.e., imagery and sensor inputs) from different points of view to comprehend the network status and derive realistic distributions for safety factors. An immediate benefit of such networks would be assessing the contribution of different factors on safety distributions and providing advisory to improve roadways and infrastructure design as well as developing safety enforcement and public education campaigns. 


A key challenge is developing strategies and scheduling policies for data aggregation to provide required modalities for network-level safety analysis at the minimum cost possible. Also, data aggregation is constrained by the utilized networking infrastructure and communication protocols between the vehicles and roadside infrastructure. Study of such networks are out of the scope of this work, and we refer the interested users to \cite{al2017internet,allal2012geocast,boussoufa2018geographic,dhankhar2014vanets,allal2013geocast,cheng2018big,martinez2011survey,luo2004survey,al2017internet}

Recently, the ideas of using data augmentation and physics-informed neural networks (PINN) are proposed to mimic the dynamics of complex systems while mitigating the need for manual annotation of massive datasets. We believe that the power of PINN and data augmentation is not yet fully utilized in this context. There is a great potential to develop surrogate models for traffic flow and risk analysis. The authors also believe that elegantly designed graph neural networks can play an essential role in modeling network-level events and trends.

\section{Sample Problems}  \label{sec:sProblem}
Safety assessment is a qualitative process, which can be approached by a set of specific and objective sub-problems. For instance, the overall safety of a highway section can be assessed by a set of exemplary problems provided in Table \ref{tab:problem}. Cloud-based software can process videos captured by the roadside infrastructure to extract statistical information from the observed events. The results of such analyses can be used to evaluate the highway's safety profile and offer revisions to the traffic management guidelines to the transportation personnel. For instance, frequent roadblocks in specific sections of a highway may require widening the road, revisiting speed limits, prohibiting commercial vehicles, or planning traffic re-routing. 

Table \ref{tab:problem} presents typical processing steps required for each sample problem. As we discussed in this paper, some technical papers offer a solution for some problems, while more work is required to solve some other problems. Also, it is worth mentioning that some steps are necessary for each problem, while some other pre-processing steps such as denoising and video stabilization may or may not be included to balance between the accuracy and complexity.

\begin{table*}[h]
\centering
\caption{List of sample problems to assess the overall safety of a highway segment. The short codes include VS: video stabilization, DN: denosing, SR: super resolution, MOT: multi-object tracking, OD: object detection, IS: instance segmentation, TE: trajectory extraction, SS: semantic segmentation, C: classification. The step in brackets "()" denotes this step is not required but potentially can enhance the performance. The tasks with '*' meaning that these tasks can be solved by obtaining the relationships (distance, velocity) among vehicles and determining a threshold from the relationships.  }
\resizebox{0.7\textwidth}{!}{%
\begin{tabular}{llc}
\toprule
\textbf{Event}                              & \textbf{Processing Steps}     & \textbf{Examples}                          \\ \midrule
Vehicle Stopped on Shoulder                 & (VS)-(DN)-LD-MOT              & \cite{xu2018dual}                          \\ \midrule
Car Crash                                & (VS)-(DN)-OD/MOT-(IS)-C       & \cite{tian2019automatic,nguyen2020anomaly} \\ \midrule
Emergency Vehicle on The Road               & (VS)-(DN)-fine-grained OD     & \cite{nayak2019vision}                     \\ \midrule
Careless/Evasive Lane Change           & (VS)-(DN)-LD-MOT-TE-(smooting)-C &    \cite{chen2017dangerous,ramyar2016identification}       \\ \midrule
Debris in Roadway                      & (VS)-(DN)-LD-(SR)-OD/(SS)          &  \cite{ohgushi2020road,hua2019small,deepika2017obstacle} \\ \midrule
Traffic Blocked (Slowed-down)          & (VS)-(DN)-MOT-(smoothing)-TE-(SE) & \cite{kaltsa2018multiple,wang2019abnormal}              \\ \midrule
Sharp Braking                               & (VS)-(DN)-MOT-(smoothing)-TE  & \cite{jiang2015analysis}                   \\ \midrule
Passing Red Traffic Light              & (VS)-(DN)-(OD for ligthts)-MOT   & \cite{wonghabut2018traffic}                             \\ \midrule
Traffic Composition                         & (VS)-(DN)-OD/MOT              & \cite{dai2019video}                        \\ \midrule
*Vehicle Driving Obviously Excessively Slow & (VS)-(DN)-MOT-TE-(smoothing)  &    -                                        \\ \midrule
*Vehicle Driving Obviously Excessively Fast & (VS)-(DN)-MOT-TE-(smoothing)  &     -                                       \\ \midrule
Vehicle Driving in the Prohibited Area & (VS)-(DN)-LD-OD                  & \cite{nowosielski2016automatic}                         \\ \midrule
Invalid Car in HOV                          & (VS)-(DN)-LD-OD               &     -                                       \\ \midrule
Improper (Careless) Entering the Road  & (VS)-(DN)-LD-MOT-TE-(smooting)-C & \cite{datondji2016survey}                               \\ \midrule
Improper (Careless) Exiting the Road   & (VS)-(DN)-LD-MOT-TE-(smooting)-C & \cite{datondji2016survey}                               \\ \midrule
Zigzag Driving                              & (VS)-(LD)-MOT-TE-(smooting)-C & \cite{chen2017dangerous}                   \\ \midrule
*Distance Violation to Front Vehicle        & (VS)-(DN)-MOT-(smoothing)-C   &       -                                     \\ \midrule
*Distance Violation to Side Vehicle         & (VS)-(DN)-MOT-(smoothing)-C   &       -                                     \\ \midrule
Violation of The Lines                      & (VS)-(DN)-LD-MOT-(smoothing)-TE    & \cite{nowosielski2016automatic}            \\ \midrule
Pedestrian on The Road                      & (VS)-(DN)-OD                  &      -                                      \\ \midrule
Bike/Motorcycle on The Road                 & (VS)-(DN)-OD                  &       -                                     \\ \midrule
Trailer [Oversized Car] on The Road         & (VS)-(DN)-OD                  &       -                                     \\ \bottomrule
\end{tabular}%
}

\label{tab:problem}
\end{table*}

Here, we provide alternative processing pipelines for zigzag driving as an exemplary application, along with a comparative analysis on the accuracy and complexity of each method. 
Zigzag driving on highways is a single-vehicle-involved event defined as frequent shifts to left and right in a short period of time $t_s$. Figs. \ref{fig:zigzag}(c)-(e) show the three bases of zigzag driving during the short period $t_s$, in comparison with the normal driving (including the normal lane change shown in Fig. \ref{fig:zigzag}(a), and the normal straight driving shown in Fig \ref{fig:zigzag}(b)). These zigzag driving incidents include i) zigzag driving only once without lane change (Fig. \ref{fig:zigzag}(c)). ii) zigzag driving once with lane change (Fig. \ref{fig:zigzag}(d)), and iii) zigzag driving once with multiple lane changes (Fig. \ref{fig:zigzag}(e)). The other cases of zigzag driving shown in Figs. \ref{fig:zigzag}(f)-(h) can be decomposed into three base cases. 
To solve such an easy problem, one may take different approaches, including:

\begin{figure}[]
\begin{center}
\centerline{\includegraphics[width=0.4\columnwidth]{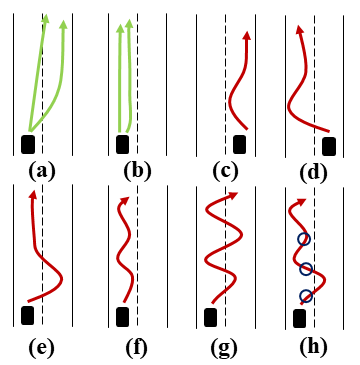}}
\caption{(a)-(d) are the trajectories with normal driving. (a) shows the normal lane changing, and (b) shows the normal straight driving. (c)-(e) are the three bases of zigzag driving. (c) is zigzag driving once without lane changing. (d) is zigzag driving once with lane changing once. (g) is zigzag driving once with lane changing twice. (f)-(h) are some examples of more aggressive and frequent zigzag driving. The blue circles in (h) denote the sampling points by \textit{HQ-sampling} with a big interval $t_0$}  
\label{fig:zigzag}
\end{center}
\end{figure}

\begin{enumerate}
\item \textit{MOT-TP}: This solution uses Multi-Object Tracking (MOT) to extract cars' motion trajectories. The overall process is as follows: Several sub-sequences of each trajectory are generated by using a sliding window with size $t_s$ (in order to save processing time, the stride can be set greater than 1). Then a sub-window with size $L$ is used to extract possible peaks and trough points to be processed by Non-Maximum Suppression (NMS) to obtain the turning points. When the number of the turning points is greater than the threshold $n_z$, this sequence will be determined as zigzag driving. Theoretically, \textit{MOT-TP} is able to detect all cases of zigzag driving, regardless of the road pattern and without the need for detecting road lanes. The potential drawback of this method is that normal driving on sharp road curves may be considered zigzag driving.

\item \textit{LD-Markings}: First, the lanes are detected by instance segmentation. Then, similar to the MOT-TP method, the motion trajectories are extracted by MOT. The zigzag driving is determined by comparing the trajectory sub-sequences with the lane markings. The trajectory is determined as zigzag driving when the number of lane changes is greater than a threshold $n_z$ within a given time interval. The accuracy of this method varies based on the performance of the utilized algorithms for trajectory extraction and lane detection. It is not suitable for dirt and unlaned roads. It also misses zigzag driving with no lane changes, such as the examples in Figs.  \ref{fig:zigzag}(f,h). 

\item \textit{Trj-Cls}: The Trajectory classification method skips modeling the road and the surrounding environment; instead, it directly applies supervised or unsupervised classification to the extracted trajectories to identify zigzag driving events. Similar to the \textit{LD-markings}, the accuracy of this method depends on the accuracy of MOT used for trajectory extraction. It is flexible and can utilize different classification methods. However, this data-driven method requires a relatively large manually annotated dataset to maintain reasonably high classification accuracy. 

\item \textit{HQ-sampling}: This method can be viewed as a simplified version of \textit{LD-Markings}. It uses high-resolution data to accurately identify the plate license numbers (by Optical Character Recognition (OCR)). This solution can track each vehicle by employing plate license numbers to replace the tracking algorithms. Instead of using video, this high-quality data often is in the form of images, to be captured continuously at intervals $t_0$, which may lead to inaccurate trajectory extraction. Similar to the \textit{LD-markings} method, \textit{HQ-sampling} determines zigzag driving by comparing the detected number of lane changes with the threshold $n_z$. Its accuracy depends on the performance of the OCR, the lane detection algorithm, and the utilized sampling interval $t_0$. The main shortcoming of this method is high complexity for short intervals, compromised accuracy for long intervals. This method may also undercount the lane changes if the sampling interval is selected large (e.g., it only counts zero lane changes in Fig \ref{fig:zigzag}(h) with the three sample points). 


\end{enumerate}

\begin{table}[hb]
\centering
\caption{Comparison of the solutions to detect zigzag driving. \\SW denotes sliding window used to search the peak and trough points.}
\resizebox{0.4\columnwidth}{!}{%
\begin{tabular}{cccc}
\toprule
\textbf{Method} & \textbf{\begin{tabular}[c]{@{}c@{}}Complexity of\\      Inference\end{tabular}} &\textbf{\begin{tabular}[c]{@{}c@{}}Sources of\\ Complexity\end{tabular}} & \textbf{Accuracy} \\ \midrule
\textit{MOT-TP}      & M & MOT+SW & H \\ \midrule
\textit{LD-markings} & L & LD+MOT & M \\ \midrule
\textit{Trj-Cls}     & H & MOT+ML & H \\ \midrule
\textit{HQ-sampling} & M/H & LD+OCR & L/M \\ \bottomrule
\end{tabular}%
}\label{tab:Zigzag}
\end{table}

Table \ref{tab:Zigzag} compares these four solutions in terms of accuracy and computation complexity. Note that the comparison is based on the assumption that the MOT algorithm performs well and a powerful trajectory smoothing method is deployed. Additionally, roadside video cameras are often fixed, meaning that lane detection is performed only once per scene and does not substantially affect the computation load. 

As seen, there exist many alternative solutions for such a simple problem. Therefore, a thorough understanding of methods can help the researchers design the most effective processing pipeline for the problem at hand.
It is worth mentioning that other trajectory-based tasks often have similar processing steps of (VS)-(DN)-LD-MOT-TE-(smoothing)-C, as shown in Table \ref{tab:problem}. It means that some parts of processing pipelines can be shared among different problems, or transferable inference models can be used for multiple tasks to lower the cost.

\section{\ar{New Trends in Deep Learning}} \label{sec:new-trends}

\ar{In this section, we review new trends in deep learning that are expected to substantially influence the field of video-based traffic safety analysis.}


\subsection{\ar{Computer Vision by Transformers}}\label{sec:transformer}

\ar{We reviewed various DL methods for vision-based traffic video analysis in Section \ref{sec:video-proc}. Most DL networks include Fully Connected (FC) layers, Convolutional Neural Networks (CNN), or Recurrent Neural Networks (RNNs) as their backbone. This journey witnessed several milestones such as the emerge of LeNet \cite{lecun1989backpropagation} in 1989, AlexNet \cite{krizhevsky2012imagenet} in 2012, VGG-Net \cite{simonyan2014very} in 2014, and GoogleNet\cite{szegedy2015going} in 2015. ResNet \cite{he2016deep} was amongst the most surprising developments that beat Human perception with a 3.57\% top-5 error rate in the ImageNet competition. Parallel to CV, DL methods are developed for sequential learning, mainly in the field of Natural Language Processing (NLP). For years, sequential learning was dominated by Gated Recurrent Units (GRUs) \cite{cho2014properties} and Long Short-term Memory network (LSTM) \cite{hochreiter1997long}, until recently when a revolutionary framework known as Transformer is introduced by Google researchers \cite{vaswani2017attention}. It has an encoder-decoder architecture and uses multi-head self-attention modules to capture longer internal dependencies in addition to the input-output dependencies in sequential data.}
\ar{Position embedding enables parallelizing the training process and capturing dependencies beyond the sequential relations. Technical details and different variants of transformers are discussed in \cite{vaswani2017attention,han2022survey}.}

\ar{Transformers are proven to be successful in capturing short-term and long-term dependencies in both NLP and CV tasks. The most classic work that uses Transformer architecture is \cite{devlin2018bert} from Google, which outperforms the competitors in 11 NLP tasks. The emerge of Transformers is considered the end of the LSTM era in NLP, and can even challenge the position of CNN in CV tasks in the next few years. For instance, Google's implementation of Transformers \cite{dosovitskiy2020image} achieved an unprecedented accuracy of 88.55\% on ImageNet, taking advantage of transfer learning. They pre-trained the network on a very large dataset (e.g., JFT-300M), then fine-tuned if for ImageNet.} 
\ar{Similarly, the authors of \cite{yuan2021tokens} achieved an 83.3\% top-1 accuracy and beat the ResNet50 with a similar number of parameters. They showed the Transformers' capability on downstream tasks, such as detection and semantic segmentation.
We expect to see more Transformer-based safety analysis frameworks in years to come, not only in CV tasks, but also for processing sequential data, such as trajectory extraction, tracing, and individual and collective behavior modeling of pedestrians, self-driving cars, and AVs. }

\subsection{\ar{Federated Learning}}

\ar{One of the challenges in traffic safety analysis is processing massive volumes of RSU videos. Several approaches, such as cloud computing, edge/fog computing, down-sampling, resolution reduction, importance sampling, and event-triggered on-demand processing methods, are developed by researchers to tackle this problem while not missing key safety events. One key research direction is developing optimal strategies for data and model sharing so that more meaningful information is extracted by collective processing of the aggregated data in RSUs. One of these methods is using Federated Learning (FL).}

\ar{Federated learning was first introduced by Google \cite{mcmahan2017communication} in 2017 as a decentralized collaborative technique. Federated learning aims to address several practical issues in practical scenarios, such as vehicular edge computing. Some of these issues include: i) the data is often in silos for privacy problems; ii) the communication capacity is limited; iii) data is unbalanced in different local nodes, and iv) data not be independent and identically distributed (non-i.i.d), for example, in-board cameras in two vehicles from different regions can both record specific types of roads, but the roads in different regions often have different styles. Federated learning is categorized into three classes based on data partitioning \cite{zhang2021survey}: i) horizontal federated learning, which has more overlapped features with fewer overlapped users; ii) vertical federated learning, which has less overlapped features but with more overlapped users; and iii) federated transfer learning, with non-overlapping features and users. To train an FL model, the local nodes only train on their local data and upload the encrypted learned weights or gradients to the central server without sharing data. Then the central server aggregates this information to update the global model and release the model to the nodes for the next round of learning. Two baselines algorithms, known as \texttt{FedSGD} and \texttt{FedAvg}, are implemented in \cite{mcmahan2017communication}. Specifically, in each round of \texttt{FedSGD}, each local client updates once, and then the global model is updated based on the weighted average of the updated gradient from the clients. A little different from \texttt{FedSGD}, \texttt{FedAvg} performs more aggressively, in which the local client takes more steps for each central server update. The experimental result shows that \texttt{FedAvg} imposes less communication load compared to \texttt{FedSGD} to achieve the same accuracy. The authors of \cite{li2019convergence} proved the convergence of \texttt{FedAvg} under strong convexity and smoothness assumptions. In past years, some works have already used FL in smart transportation, such as a privacy-preserving traffic prediction introduced in \cite{liu2020privacy,elbir2020federated}, and FL-based Vehicle-to-Vehicle (V2V) and Vehicle-To-Everything (V2X) low-latency communication introduced in \cite{samarakoon2018federated,zhang2019deep}. We refer the interested reader to \cite{nguyen2021federated} for a more interesting and broader range of applications of FL.}


\ar{It is noteworthy that FL still has some unresolved issues. Firstly, its reliability and robustness under uncertain conditions are not yet well established. Specifically, as discussed before, the center server has no access to the entirety of the local data and is unaware of local update progress. It means the aggregated model is much vulnerable to defending the introduced backdoor functionality if one or some nodes are malicious \cite{bagdasaryan2020backdoor}. Additionally, most practical FL networks consist of heterogeneous local nodes with varying learning ability and communication rates, which can challenge the overall robustness and performance of the system.  
After solving its issues, FL would be a promising solution for traffic safety analysis, as it provides a powerful tool to extract and analyze massive data volumes while protecting the privacy of road users. RSUs and in-vehicle processors can act as central and local processors of a large-scale FL system.}

\subsection{\ar{Adversarial Learning For Safety Analysis}} \label{sec:AML} 
\begin{figure}[]
\begin{center}
\centerline{\includegraphics[width=0.8\columnwidth]{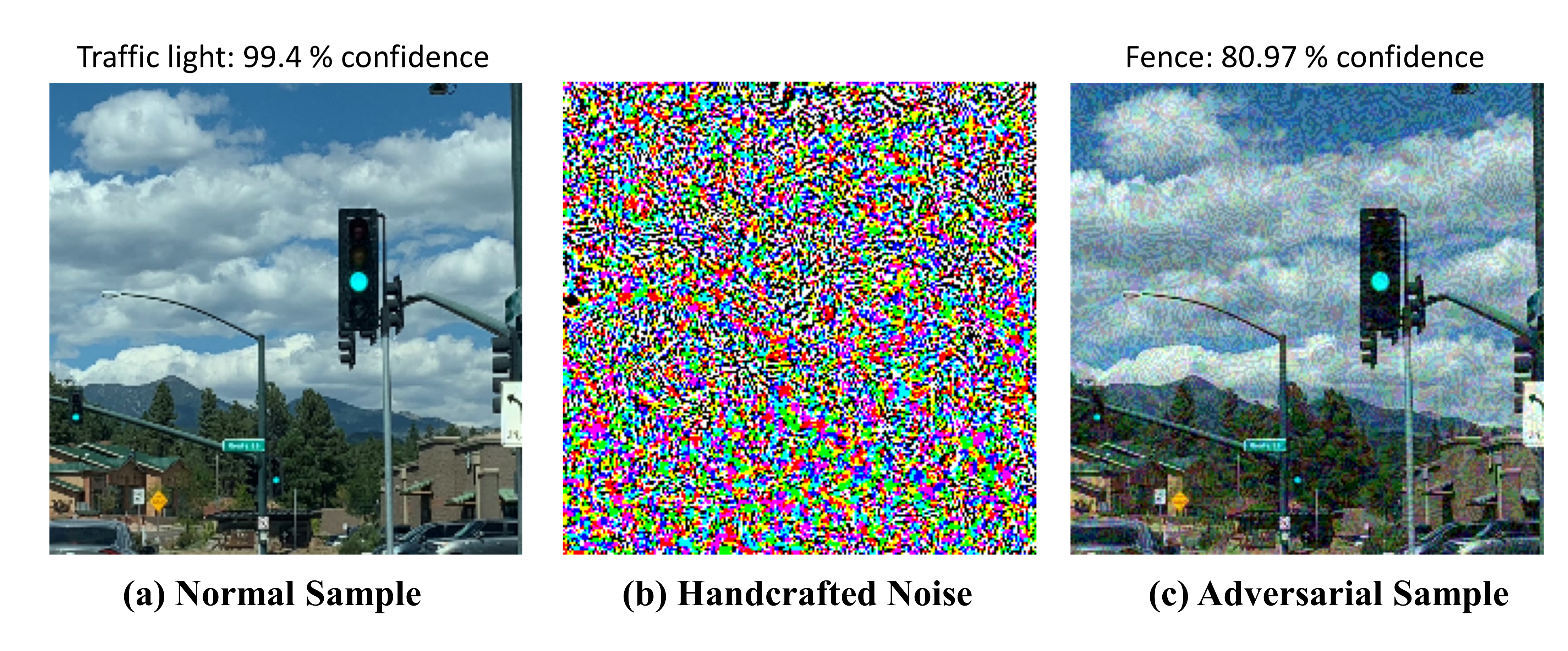}}
\caption{\ar{An example of adversarial samples generated by Fast Gradient Signed Method (FGSM) \cite{goodfellow2014explaining}. The detector is MobileNetV2 trained on ImageNet. (a) is the normal sample, classified as "traffic light" by MobileNetV2, (b) is the hand crafted noise (adversarial attack), and (c) is the adversarial sample generated by adding hand crafted noise to the normal sample, classified as "fence" by MobileNetV2.}}  
\label{fig:adv_sample}
\end{center}
\end{figure}

\ar{Despite the demonstrated power of DL methods in learning from massive data, the community is increasingly concerned about the security and reliability of neural networks. One reason for this concern is that small perturbations can easily fool a big DL network \cite{szegedy2013intriguing}, which can lead to misleading results in traffic safety analysis and operation, especially for self-driving cars.} \ar{Fig. \ref{fig:adv_sample} demonstrates an example that the adversarial sample generated by Fast Gradient Signed Method (FGSM) \cite{goodfellow2014explaining}, which easily fools MobileNetV2 to detect a traffic light as a fence. }

\ar{Adversarial machine learning aims to alleviate these issues by exploring the vulnerabilities of ML algorithms and defending against potential attacks. The attacks can be divided into two main categories: evasion/exploratory attacks and Poisoning/causative attacks \cite{szegedy2013intriguing,madry2017towards}. The former cannot tamper with the model but feed carefully designed adversarially perturbed examples to the network in the testing phase. These data (sometimes imperceptible) can significantly change the feature representations and result in fatal misclassifications. In the latter case, the attacker injects malicious data samples into the training dataset to misguide the model training. It is a legitimate concern if one anticipates that adversarial samples are inserted in the training dataset using software backdoors \cite{chen2017targeted}. 
Alternatively, one way poison the training process is by explicitly attempting to corrupt the trained model. For instance, poisoning data samples can be added to cause gradient vanishing \cite{huang2021unlearnable}. }

\ar{Indeed, one would suppose that the attacker and defender play a two-player game, where the defender expects to design a robust model against the adversarial samples. 
It augments the data by adding the adversarial samples in the training phase, and then solves a min-max problem. Another strategy is to try to recognize and filter out the adversarial examples \cite{feinman2017detecting}. Since many attacks are implemented by tracking the gradient, some defenders (such as \cite{papernot2016distillation}) attend to hide or change the gradient information. Some other defenders design robust optimization based on regularization methods against attacks \cite{cisse2017parseval}.}

\ar{Finally, we remind that these attacks or defense methods are often designed for specific or known situations. However, real-world applications are often exposed to unknown attacks. This concern can be even more critical when developing traffic-related frameworks. 
The driving safety analysis requires further investments in developing robust models using inverse learning. The models should undergo heavy inspections before being utilized in traffic control systems.}

\subsection{\ar{Meta-Learning}}\label{sec:meta}
\ar{Meta-Learning, aka learn-to-learn, aims to learn the learning process from other machine learning or deep learning tasks, and then apply this learned experience to guide the design and training process for a new task. It can generally learn the new task faster than the manual training; therefore facilitates designing a novel algorithm or framework merely by data-driven methods \cite{vanschoren2018meta}. To be more specific, in general learning, the optimizer only learns on the current dataset, and the objective function is defined only for the current task. However, meta-learning focuses on the elements of the learning phase (such as parameter initialization, optimizer, architecture, etc.) and learns how to generalize these elements across tasks and how to perform better on new tasks. Formally, the problem is presented  by \cite{hospedales2020meta} as:
\begin{align}\label{eq:meta}
&\underset{\omega}{ \min } \sum \mathcal{L}^{\text {meta }}\left(\theta^{*(i)}(\omega), \omega, \mathcal{D}^{(i)}\right), \\ \nonumber
&\text { s.t.~~~ }  \theta^{*(i)}(\omega) =\arg \underset{\theta}{\min} \mathcal{L}^{\text {task }}\left(\theta, \omega, \mathcal{D}^{(i)}\right),
\end{align}
where $\omega$ denotes the meta prior to be learned, $\theta^{*(i)}(\omega)$ denotes the learned parameters on the dataset $\mathcal{D}^{(i)}$ for task $i$ under the meta prior $\omega$. $\mathcal{L}^{\text {meta }}$ and $\mathcal{L}^{\text {task }}$ denote the loss of meta-learning and the task itself, respectively. Meta-learning can be trained using a non-parametric setup by measuring the distance between the target and trained samples, and predicting the sample label based on the closest trained samples. A differentiable  $ \mathcal{L}^{\text {meta}}$ can be solved by gradient descent. }

\ar{It is noteworthy that this concept is somewhat related to transfer learning but is also slightly different from it. Specifically, transfer learning aims to learn the regular and trainable parameters, such as the weights of the networks from other datasets or tasks, then fine-tunes the pre-trained model for a new task, while meta-learning tries to learn the hyper-parameters from other tasks and then utilizes them to learn a new task from scratch. Both transfer learning and meta learning can help develop more accurate driving safety analysis, taking advantage of similar datasets.}



\subsection{\ar{Unsupervised Spatiotemporal Representation Learning}}\label{sec:unsuperST}

\ar{Video data plays an essential role in traffic-related tasks; however, supervised learning can be prohibitively expensive in some traffic scenarios, such as driving anomaly detection, due to the high cost of data annotation. Under such limitations, unsupervised and self-supervised learning methods would be an appealing way of processing traffic video.}

\ar{One approach to developing unsupervised learning is taking advantage of representation learning to translate raw input into more workable representations. An exemplary scenario would be representing video frames in a latent space that shows a substantial distinction between safe and unsafe traffic volumes.}

\ar{Generally, the goal of unsupervised learning is learning data in a latent and often a lower-dimensional space, which offers temporal invariance and simplifies the downstream analysis. In 2021, He et al. \cite{feichtenhofer2021large} investigated the performance of four popular image-based unsupervised frameworks (Momentum Contrast (MoCo) \cite{he2020momentum}, a Simple framework for Contrastive Learning of visual Representations (SimCLR) \cite{chen2020simple}, Bootstrap Your Own Latent (BYOL) \cite{grill2020bootstrap}, and Swapping Assignments between multiple Views of the Same image (SwAV) \cite{caron2020unsupervised}) for the spatiotemporal representation learning. The authors named the spatiotemporal invariant as \textit{temporally-persistent}. Specifically, they anticipated that the high-level representation of the same video would be similar in different clips. These four frameworks take advantage of contrastive learning, which aims to maximize the similarity of positive samples and minimize the similarity of negative samples. Generally, the negative samples are the sampled clips from other videos. In MoCo, the authors introduced a creative method known as momentum coding, which constructs a dynamic dictionary to manage the representation of samples. This allows to quickly build and update the large and consistent dictionary to store much representation with a lower memory cost, which hugely enhances the ability of contrastive learning. BYOL and SwAV are considered versions of SimCLR and MoCo without using negative samples, respectively. They showed that these extended image-based frameworks could learn the spatiotemporal representation of video. A more interesting result is the competitive performance of these frameworks when combined with downstream tasks, compared to supervised learning in many tasks. These results provide a limitless possibility to deploy the unsupervised frameworks to industry and real applications. }


\section{Traffic Datasets}   \label{sec:dataset}

\begin{figure*}[ht]
\begin{center}
\centerline{\includegraphics[width=0.6\textwidth]{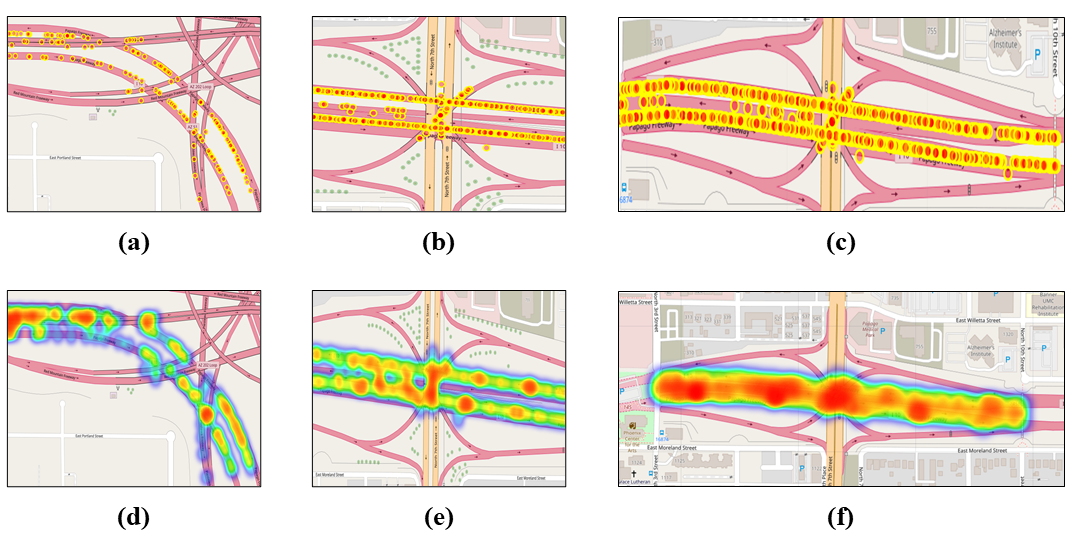}}
\caption{The visualization of some exemplary crash data. Each point in (a)-(c) denotes a crash, and (d)-(f) are the corresponding heatmaps.} 
\label{fig:crash_data}
\end{center}
\end{figure*}

\begin{table*}[h]
\centering\caption{Some popular traffic-related datasets. }
\resizebox{\textwidth}{!}{%
\begin{tabular}{llll}
\toprule
\textbf{Category} &
  \textbf{Dataset} &
  \textbf{Size} &
  \textbf{Features} \\ \midrule

\multirow{6}{*}{\begin{tabular}[c]{@{}l@{}}Trajectory\\ analysis\end{tabular}} &
  MIT traffic dataset(2011)\cite{wang2011automatic} &
  \begin{tabular}[c]{@{}l@{}}around 162, 000 frames\\ from a 90 minutes long video sequence (30 fps)\end{tabular} &
  \begin{tabular}[c]{@{}l@{}}Both pedestrian and vehicle movements;\\ with occlusions and varying illumination conditions\end{tabular} \\ \cmidrule{2-4} 
 &
  MIT trajectory dataset(2008)\cite{grimson2008trajectory} &
  577 radar tracks, 45 453 video tracks &
  Multiple single camera view \\ \cmidrule{2-4} 
 &
  CVRR trajectory analysis datasets (2011) \cite{morris2011trajectory}  &
  \begin{tabular}[c]{@{}l@{}}4 scenes: a simulated intersection (19 activities),\\ a real highway (8 activities)\end{tabular} &
  \begin{tabular}[c]{@{}l@{}}Units are pixels;\\ only contain spatial information\end{tabular} \\ \cmidrule{2-4} 
 &
  i-Lids–advanced vehicle detection challenge\cite{ilidsadvanced_2007} &
  35000 images. 7 sequences with 25 fps. &
  \begin{tabular}[c]{@{}l@{}}Consider daytime;\\ various locations in the UK\end{tabular} \\ \cmidrule{2-4} 
 &
  NGSIM(2007)\cite{NGSIM}  &
  45 mins+45 mins+ 30 mins&
  \begin{tabular}[c]{@{}l@{}}Three sub-datasets;\\ US-101 and I-80 for freeway scenario;\\ Lankershim Boulevard Dataset for arterial corridors scenario;\\ each dataset is segmented with different congestion conditions \end{tabular} \\ \cmidrule{2-4} 
 
  &
  QMUL junction dataset(2012) \cite{hospedales2012video} &
  1 hour video &
  Busy traffic at a junction \\ \midrule
\multirow{7}{*}{\begin{tabular}[c]{@{}l@{}}Auto-driving\\ tasks\end{tabular}} &
  CamVid(2009) \cite{brostow2009semantic} &
  700 images, 32 classes, 10 min HD video with 30fps &
  \begin{tabular}[c]{@{}l@{}}Support pixel-wise segmentation tasks;\\ provide 3D camera pose\end{tabular} \\ \cmidrule{2-4} 
 &
  Cityscapes (2016)\cite{cordts2016cityscapes} &
  \begin{tabular}[c]{@{}l@{}}5000 pixel-level annotated images\\  and 20000 coarse annotated images (weak-label)\end{tabular} &
  \begin{tabular}[c]{@{}l@{}}complex urban street scenes;\\  record from 50 cities;\end{tabular} \\ \cmidrule{2-4} 
 &
  Mapillary Vistas Dataset(2017)\cite{neuhold2017mapillary} &
  \begin{tabular}[c]{@{}l@{}}25 000 high-resolution images; \\ 66 classes for objection\\ 37 instance-specific labels\end{tabular} &
  \begin{tabular}[c]{@{}l@{}}Various conditions regarding weather, \\ season and daytime\end{tabular} \\ \cmidrule{2-4} 
 &
  SYNTHIA(2016) \cite{ros2016synthia} &
  \begin{tabular}[c]{@{}l@{}}13,400 frames of the city (real);\\ \\ four video sequences of  ~50,000 frames (simulated);\\ 13 classes\end{tabular} &
  \begin{tabular}[c]{@{}l@{}}Scene diversity;\\ multiple seasons;\\ lighting conditions and weather\end{tabular} \\ \cmidrule{2-4} 
 &
  KITTI Vision Benchmark Suite (2012)\cite{Geiger2012CVPR}&
  7481 training images and 7518 test images; &
  \begin{tabular}[c]{@{}l@{}}Up to 15 cars and 30 pedestrians per image;\\ tasks of interest are stereo, optical flow, \\ visual odometry, 3D object detection, and 3D tracking\end{tabular} \\ \cmidrule{2-4} 
 &
  Berkeley Deep Drive (2018)\cite{yu2018bdd100k}  &
  over 100K videos &
  \begin{tabular}[c]{@{}l@{}}Diversity of annotations;\\ diversity of geography, environment, and weather\end{tabular} \\ \cmidrule{2-4} 
 &
  IDD (2019) \cite{varma2019idd}&
  10,004 images; 34 classes; 182 drive sequences &
  \begin{tabular}[c]{@{}l@{}}Unstructured environments;\\ four-level label hierarchy; possibility for new tasks: \\ domain adaptation, \\ few-shot learning and behavior prediction\end{tabular} \\ \midrule
\multirow{6}{*}{\begin{tabular}[c]{@{}l@{}}Traffic sign\\ detection\end{tabular}} &
  TLR (2009)\cite{de2009real} &
  \begin{tabular}[c]{@{}l@{}}11 179 frames (8min 49sec, @25FPS);\\ 9 168 instances of traffic lights\end{tabular} &
  \begin{tabular}[c]{@{}l@{}}Hand-labeled;\\ tasks for traffic light detection and traffic sign detection\end{tabular} \\ \cmidrule{2-4} 
 &
  LISA (2012)\cite{mogelmose2012vision}&
  \begin{tabular}[c]{@{}l@{}}6600 frames with ~7800 annotated objects;\\ 49 classes of signs\end{tabular} &
  \begin{tabular}[c]{@{}l@{}}Recorded in the US;\\ Images from various camera types\end{tabular} \\ \cmidrule{2-4} 
 &
  GTSDB(2013)\cite{houben2013detection} &
  \begin{tabular}[c]{@{}l@{}}900 images \\ divided into 600 training images (846 traffic signs) \\ and 300 evaluation images (360 traffic signs)\end{tabular} &
  \begin{tabular}[c]{@{}l@{}}Signs are divided into prohibitory, danger, mandatory, \\ and other signs\end{tabular} \\ \cmidrule{2-4} 
 &
  BelgianTS(2014)\cite{timofte2014multi} &
  \begin{tabular}[c]{@{}l@{}}13,444 traffic sign annotations in 9,006 still images \\ corresponding to 4,565\\ physically distinct traffic signs.\end{tabular} &
  Available for 2D and 3D testing \\ \cmidrule{2-4} 
 &
  TT100K (2016)\cite{zhu2016traffic}&
  \begin{tabular}[c]{@{}l@{}}100000 images containing 30000 traffic-sign instances;\\  128 sign classes\end{tabular} &
  Diversity of annotation \\ \cmidrule{2-4} 
 &
  BSTL (2017)\cite{behrendt2017deep}&
  \begin{tabular}[c]{@{}l@{}}13427 images (5093training+8334test~15fps);\\ 24000 annotated traffic lights\end{tabular} &
  \begin{tabular}[c]{@{}l@{}}Cover a decent variety of road scenes and\\  typical difficulties\end{tabular} \\ \midrule
\multirow{4}{*}{Plate detection} &
  ReId (2017)\cite{ReId} &
  9.5hours from 8 locations &
  \begin{tabular}[c]{@{}l@{}}Varied daytime;\\ multi-plates on a frame;\\ cameras placed on bridges above highways;\\ relatively small objects\end{tabular} \\ \cmidrule{2-4} 
 &
  UCSD car dataset(2007) \cite{UCSDCARS} &
  \begin{tabular}[c]{@{}l@{}}10 hours video;\\ 878 cars tracked from the moment;\\ 291 still cars recorded in parking lot\end{tabular} &
  Label with make, model, license plate location, plate texts \\ \cmidrule{2-4} 
 &
  UFPR-ALPR Dataset(2018)\cite{laroca2018robust_UFPR} &
  \begin{tabular}[c]{@{}l@{}}4,500 fully annotated images from 150 vehicles;\\ 30 fps\\ >30,000 license characters;\end{tabular} &
  \begin{tabular}[c]{@{}l@{}}Different license color;\\ both vehicle and camera are moving\end{tabular} \\ \cmidrule{2-4} 
 &
  SSIG Dataset(2016)\cite{gonccalves2016benchmark_SSIG} &
  \begin{tabular}[c]{@{}l@{}}2000 Brazilian license plates; \\ 14,000 alphanumeric symbols\end{tabular} &
  \begin{tabular}[c]{@{}l@{}}Static camera view;\\ character are segmented\end{tabular} \\  \bottomrule
\end{tabular}
}
\label{tab:trafficdataset}
\end{table*}

\begin{table*}[h]
\centering\caption{Some popular traffic-related datasets (continue).}
\resizebox{\textwidth}{!}{%
\begin{tabular}{llll}
\toprule
\textbf{Category} &
  \textbf{Dataset} &
  \textbf{Size} &
  \textbf{Features} \\ \midrule
\multirow{2}{*}{\begin{tabular}[c]{@{}l@{}}Naturalistic\\ Study\end{tabular}} &
  The 100-Car Naturalistic Driving Study(2006)\cite{dingus2006100} &
  \begin{tabular}[c]{@{}l@{}}~2,000,000 vehicle miles;\\ ~43,000 hours of data\end{tabular} &
  \begin{tabular}[c]{@{}l@{}}Contains extreme cases of driving behavior;\\ Event-based database\end{tabular} \\ \cmidrule{2-4} 
 &
  SHRP 2 NDS dataset(2012)\cite{the_shrp_2_2012}  &
  more than 2 PB of continuous naturalistic driving data &
  \begin{tabular}[c]{@{}l@{}}Multi-view video outputs;\\ collected during a 3-y period from \\ more than 3,500 participants, aged 16–98\end{tabular} \\ \midrule

\multirow{4}{*}{UAV} &
  MOR-UAV(2020)\cite{MOR-UAV} &
  \begin{tabular}[c]{@{}l@{}}89,783 moving object instances;\\  10,948 various scenarios;\end{tabular} &
  \begin{tabular}[c]{@{}l@{}}Tasks for moving and non-moving objects;\\ varied conditions of the environment;\\ labeled by aixs-aligned bounding boxes\end{tabular} \\ \cmidrule{2-4} 
 &
  Stanford Drone Dataset(2016)\cite{robicquet2016learning} &
  \begin{tabular}[c]{@{}l@{}}>100 top-view scenes;\\ 20,000 targets\\ annotated trajectories and id\end{tabular} &
  \begin{tabular}[c]{@{}l@{}}Real-world scenario-based;\\ varied types of targets\end{tabular} \\ \cmidrule{2-4} 
 &
  The highD dataset(2018)\cite{krajewski2018highd} &
  \begin{tabular}[c]{@{}l@{}}16.5hous of measurements for 110,000 vehicles from 6\\ locations.\end{tabular} &
  \begin{tabular}[c]{@{}l@{}}Real-world scenario-based;\\ focus on highways\\ contain naturalistic behavior of road users\end{tabular} \\ \cmidrule{2-4} 
 &
  The inD dataset(2019)\cite{bock2019ind} &
  \begin{tabular}[c]{@{}l@{}}contains more than 11500 road users;\\ 10 hours measurement\end{tabular} &
  \begin{tabular}[c]{@{}l@{}}Successor of highD,\\ focus on intersections\end{tabular} \\ \midrule
  \multirow{4}{*}{\begin{tabular}[c]{@{}l@{}}Video-based \\ Traffic events\\ (Crash)\end{tabular}} &
  UCF-Crimes(2018) \cite{sultani2018real} &
  1900 sequences, 128hours &
  \begin{tabular}[c]{@{}l@{}}Includes realistic anomalies such as fighting, road crash;\\ No spatial annotation\end{tabular} \\ \cmidrule{2-4} 
 &
  DAD(2016) \cite{chan2016anticipating} &
  1730 sequences, 2.4 hours &
  \begin{tabular}[c]{@{}l@{}}Spatio-temporal annotation\\ Video captured by dashcam\end{tabular} \\ \cmidrule{2-4} 
 &
  CADP(2018)\cite{shah2018cadp} &
  1416 sequences, 5.2 hours &
  \begin{tabular}[c]{@{}l@{}}Include multi-crash sequences\\ Captured by CCTV; Include varied environment\end{tabular} \\
  \cmidrule{2-4}
  &
  NVIDIA AI CITY challenge\cite{ai_city_challenge} &
  \begin{tabular}[c]{@{}l@{}}Track 1: 31 videos (about 9 hours in total)\\ Track 2: 56,277 images,\\ 36,935 come from 333 object identities\\ Track 3: 215.03 minutes videos \\ 300K bounding boxes\\ Track 4: 200 15mins videos 30fps\end{tabular} &
  \begin{tabular}[c]{@{}l@{}}Track 1: vehicle counting\\ Track 2: vehicle Re-id\\ Track 3: vehicle tracking\\ Track 4: Abnormal detection\\ Anomalies can be due to \\ car crashes or stalled vehicles\end{tabular} \\ \midrule
  
\multirow{2}{*}{\begin{tabular}[c]{@{}l@{}}Fine-grained\\ Vehicle\\ Classification\end{tabular}} &
  Standford Cars\cite{krause20133d} &
  16,185 images of 196 classes of cars &
  \begin{tabular}[c]{@{}l@{}}The cars images are taken from many angles;\\ classes typically based on Make, Model, Year\end{tabular} \\ \cmidrule{2-4} 
 &
  COMPCARS\cite{yang2015large} &
  \begin{tabular}[c]{@{}l@{}}Web-nature: 136,726 (entire car)+27,618 (car parts),\\ 163 car makes with 1,716 car models;\\ Surveillance-nature: 5000 car images\end{tabular} &
  \begin{tabular}[c]{@{}l@{}}Web-nature data are labeled with \\ bounding boxes and viewpoints;\\ the surveillance-nature images are in the \\ front view\end{tabular} \\  \midrule
  \blr
 \multirow{5}{*}{Pedestrian Detection} & \blr Caltech \cite{dollar2011pedestrian} (2009) & \blr\begin{tabular}[c]{@{}l@{}}61k images with 192k pedestrians   to train;\\   \blr   56k images with 155k pedestrians to test;\\      1273 unique pedestrians\end{tabular} &\blr \begin{tabular}[c]{@{}l@{}}Objects with different   occlusion;\\      Objects with multi-scale;\\      Sequential images\end{tabular} \\ \cmidrule{2-4} 
 &\blr CityPersons \cite{zhang2017citypersons}   (2016) &\blr \begin{tabular}[c]{@{}l@{}}5k images with 35k   pedestrians;\\      19654 unique pedestrians\end{tabular} &\blr \begin{tabular}[c]{@{}l@{}}Built upon Cityscapes;\\      Record in different city with different weather;\\      Objects with different occlusion;\end{tabular} \\ \cmidrule{2-4} 
 &\blr KAIST \cite{hwang2015multispectral}   (2015) &\blr \begin{tabular}[c]{@{}l@{}}95,328 annotated image pairs   (thermal and visible),\\       103,128 pedestrian annotation,\\      1182 unique pedestrians\end{tabular} &\blr \begin{tabular}[c]{@{}l@{}}Multi-spectral;\\      Include partial occlusion and heavy occlusion;\\      Include day and night;\\      Sequential images\end{tabular} \\ \cmidrule{2-4} 
 &\blr CVC-14 \cite{gonzalez2016pedestrian}   (2016) &\blr \begin{tabular}[c]{@{}l@{}}5051 grayscale visible and   thermal frame pairs;\\      9,319 pedestrian annotation (visible);\\      8,750  pedestrian annotation   (thermal);\end{tabular} &\blr \begin{tabular}[c]{@{}l@{}}Multi-spectral;\\      Include day and night;\\      Sequential images;\\      Annotation include mandatory and unclear persons\end{tabular} \\ \cmidrule{2-4} 
 &\blr FLIR \cite{flir_dataset} ($\sim$2018) &\blr 9,711 thermal and 9,233 RGB   images ; &\blr \begin{tabular}[c]{@{}l@{}}Multi-spectral;\\      15 Classes;\\      Sequential images;\\      Many pairs are unaligned;\end{tabular} \\ \bottomrule
\end{tabular} 
}
\label{tab:trafficdataset-continue}
\end{table*}

Table \ref{tab:trafficdataset} and Table \ref{tab:trafficdataset-continue} provide a relatively complete list of datasets that can be used for different aspects of video-based analysis for different CV-based traffic tasks. Data for trajectory analysis is often extracted from still cameras or drone-captured videos and used to understand microscopic-level (such as car-following models) to macroscopic-level traffic flow (such as traffic wave models) \cite{li2020trajectory}. Safety analysis of human-driven vehicles closely relates to video understanding tasks such as scene recognition, object detection, instance and semantic segmentation, developed for auto-driving vehicles. Traffic sign detection is another difficult problem due to using different variants for the same signs, uncontrolled and varying light conditions, the impact of weather conditions (rain, snow, fog, etc.), and the unpredictable traffic environment, especially when real-time processing is needed.

Naturalistic Driving Studies often record and process the driver's behavior, cognition, and perception of the surrounding environment without influencing or distracting the driver. This kind of data is generally large-scale since it requires long-term observations with multiple observed objects.
The plate detection task involves drawing a bounding box around every detectable plate. A sub-task of plate detection is Optical Character Recognition (OCR), which requires understanding and validating engraved license plate numbers. Another twist to this problem is the angle of view. While most images are taken by roadside cameras, drone-based imagery datasets provide top-view images that solve some problems but poses new image processing challenges like human detection and small object detection.  

It is notable that general Computer Vision (CV) datasets, presented in Table \ref{tab:CVdatasets}, can be used to approximately evaluate the performance of developed algorithms if the desired traffic-specific dataset is not readily available. Here, we also want to mention some popular CV datasets (shown in Table \ref{tab:CVdatasets}) since the most recent state-of-the-art algorithms deploy them as their evaluation criteria.

\textbf{Crash Data:} Police-reported Crash Data, generally collected and stored by local, regional, state, and/or national government agencies after traffic crashes, can provide official, objective information about crash incidents. Sanitized crash data (e.g., personal information removed) is generally publicly available at the state level either online (e.g., Michigan crash data can be obtained at \cite{michigantrafficcrashfacts}) or by a public records request. National level crash data in the US is available from the National Highway Traffic Safety Administration (NHTSA) via the National Automotive Sampling System  General Estimates System (NASS-GES)\cite{GES} which provides a national sample of police-reported crashes and the Fatality Analysis Reporting System (FARS)\cite{FARS} which provides details on nationwide fatal crashes. Crash data sets provide a wealth of information related to crash circumstances (including sequence of crash events and crash type), environmental conditions at the time of the crash, roadway characteristics at the crash scene, vehicle/road user information, and crash involved person-level characteristics (e.g., injury severity, age, gender, impairment, etc.…), among other information. The Model Minimum Uniform Crash Criteria (MMUCC)\cite{mmucc} provides a voluntary guideline for agencies with respect to the minimum data elements that should be included in crash databases and includes a description of 115 recommended data elements related to the incident, vehicle, person, roadway, and other categories. Crash data can be used to revise traffic and/or roadway plans, develop safety countermeasures, and explore associations between crashes and traffic safety violations or other non-crash safety metrics. It can also be used to incorporate prior knowledge when analyzing traffic safety events. 


\textbf{Missing Datasets}: There is a critical need to develop new datasets for traffic analysis that cover the underexamined aspects.  
As mentioned later in section \ref{sec:humancondition}, some studies consider driving safety from the behavioral science perspective. For instance, eye motion tracking can be used to gauge the driver's attention and detect distraction episodes \cite{sodhi2002road,crundall2011visual,dukic2012older,kiefer2012towards,yang2012can}.
Very few datasets exist to facilitate such research. Among these, we found two datasets (The 100-Car Naturalistic Driving Study\cite{dingus2006100} and SHRP 2 NDS dataset\cite{the_shrp_2_2012}) that offered a relatively large number of samples for driver behavior analysis. 
However, we noted that in these datasets, the activities for one vehicle in a specific scene are non-repeating, meaning that these datasets cannot provide driver-specific information. The authors of this study are currently working to develop a small dataset for driver-specific anomaly detection by collecting aerial imagery from test drivers' behavior on specific scenarios multiple times. This dataset would enable profiling drivers based on their reaction to traffic conditions and use it to find abnormal behaviors that can be indicators of driving issues and potential crash risks.  
Also, very few datasets record traffic events from different perspectives. Datasets that can offer roadside imagery, along with aerial imagery and car-mounted cameras for synchronous analysis of traffic views, can open new research directions, particularly for multi-modal video analysis.The visualization of some exemplary crash data from Arizona Department of Transportation (ADOT) is presented in Fig.\ref{fig:crash_data}.

\begin{table*}[h]
\centering\caption{Some popular CV datasets. }
\resizebox{\textwidth}{!}{%
\begin{tabular}{lll}
\toprule
\textbf{Dataset} &
  \textbf{Size} &
  \textbf{Features} \\ \midrule
MS-COCO\cite{lin2014microsoft} &
  91 objects types. 2.5 million labeled instances in 328k images. &
  \begin{tabular}[c]{@{}l@{}}A set of datasets.\\ labeled by per-instance segmentation;\\ more instances per category\end{tabular} \\ \midrule
PASCAL VOC\cite{everingham2010pascal} &
  20 classes. ~11530 images. &
  \begin{tabular}[c]{@{}l@{}}A set of datasets.\\ Provide standard evaluation procedures;\\ contain significant variability in terms of object size, \\ orientation, pose, illumination, position and occlusion\end{tabular} \\ \midrule
ImageNet\cite{deng2009imagenet} &
  \begin{tabular}[c]{@{}l@{}}21841 non-empty synsets, 14,197,122 images. \\ 1,034,908 mages with bounding box annotations.\end{tabular} &
  Organized according to the WordNet hierarchy. \\ \bottomrule
\end{tabular}%
}

\label{tab:CVdatasets}
\end{table*}

\section{Safety Metrics } \label{sec:safety-metrics}
An essential objective of driving safety analysis is extracting \textit{operational safety metrics}, which are quantifiable measures extracted from traffic videos (or other data sources) that determine the relative risk of an event that may lead to a crash. 
Some important safety metrics that are used to analyze car crashes include: 
\begin{enumerate}
  \item \textbf{Temporal-based indicators:} Time to Collision (TTC), Extended Time to Collision (Time Exposed Time-to-Collision (TET), Time Integrated Time-to-Collision(TIT)\cite{minderhoud2001extended}), Modified TTC (MTTC), Crash Index (CI), Time-to-Accident (TA), Time Headway (H), and Post-Encroachment Time (PET). 
  
  \item \textbf{Distance-based indicators:} Potential Index for Collision with Urgent Deceleration (PICUD), Proportion of stopping Distance (PSD), Margin to Collision (MTC), Difference of Space Distance and Stopping Distance (DSS), Time Integrated DSS (TIDSS), and Unsafe Density (UD);
  
  \item \textbf{Deceleration-based indicators:} Deceleration Rate to Avoid a Crash (DRAC), Crash Potential Index(CPI), and Criticality Index Function (CIF).
\end{enumerate}

Reviewing different safety metric is out of the scope of this paper and we refer the interested readers to recent papers on safety metrics  \cite{mahmud2017application,wishart2020driving,elli2021evaluation}. As part of the Institute of Automated Mobility (IAM) project, the authors of this paper are working toward extending safety metrics into network-level metrics and developing a taxonomy of metrics for safety metrics for AVs based on  the level of access required of ADS data \cite{chen2022network}.
 



\section{Miscellaneous Points} \label{sec:points}

This section discusses some general facts about the driving safety analysis and reviews closely related research directions. We conclude by mentioning key areas and emerging topics that require further investigations.

\subsection{Study From Human Condition And Psychology Perspective:} \label{sec:humancondition}
It is notable that understanding and interpreting traffic patterns, especially for human-driven vehicles, involves behavioral and psychological factors. 
Some studies study traffic safety from a deeper perspective of assessing driver's cognition and mental capacity. For instance, \cite{lajunen1995driving,van2006convergent,de2010driver,zhao2012investigation} use inventories or questionnaires to investigate the correlation between driving quality and general personality measures. Driving inventories in this context include the Driving Behavior Inventory (DBI), the Drivers' Skill Inventory (DSI), and Montag Driving Internality and Externality Scales (MDIE).  
Some other studies including \cite{constantinou2011risky,chai2016effect,zhang2016association,ge2017effects,ball2018influence} analyze aggressive driving behaviours affected by the driver personality. The impaired driving performance caused by Synthetic Cannabinoids (SC) is investigated in \cite{tuv2014prevalence} based on the collected blood samples. This study concludes that the SC is not the only key factor causing the reduced driving skills since the drivers often use other potent psychoactive drugs simultaneously. The impact of landscapes on drivers’ mental status is investigated in \cite{jiang2020perceived,antonson2009effect,antonson2014landscape,wang2016impact}.

Some car manufactures have utilized AI systems to assess drivers' cognition systems, for instance, by monitoring eye motion and alerting drivers when distracted attention\cite{mejia_2020}. Other AI systems aim to gauge drivers' emotions and drowsiness by cameras and microphones\cite{affectiva}. 
This line of research mainly relies on sampling techniques and often can offer some empirical results, which is hard to examine and generalize. Therefore, these methods have a long way ahead for being widely adopted by the engineering community. 


\subsection{Relation to AVs}\label{sec:AVlevels}

AI platforms have been utilized in recent years to build AVs. AVs use different technologies such as regular cameras, radar, optical radar, and GPS, along with computer vision and learning methods, to realize autonomous driving. In 2015, Tesla started to commercialize 'Autopilot' features in its cars, and soon afterward, other manufacturers joined the race. Currently, there are over 250 autonomous vehicle companies, including automakers, technology providers, services providers, and tech start-ups, that are taking serious steps to make self-driven or driver-less cars a reality. According to \cite{top5av}, the top five autonomous vehicle companies are Waymo, General Motors' Cruise division, Tesla, Baidu, and Argo AI (Ford Motor).

The official level classification system for autonomous driving is defined by the Society of Automotive Engineers (SAE International) \cite{SAE2014} and approved by the National Highway Traffic Safety Administration (NHTSA). The standard describes the five autonomy levels including ~\cite{national2013us}: 
\begin{itemize}
    \item Level 0 (no automation): The human driver has full authority to operate the car, and can be assisted by warning and protection systems during driving.
    \item Level 1 (driving support/hands on): Provide driving support for one operation of the pay-off reel and acceleration/deceleration through the driving environment, and the human driver operates other driving actions.
    \item Level 2 (partial automation/hands off): The vehicle can fully control the car by acceleration, braking, and steering; however, the driver should keep monitoring the environment and be prepared to intervene immediately at any time.
    \item Level 3 (conditional automation/ eyes off): The unmanned driving system completes all driving operations. According to the system request, the human driver provides an appropriate response.
    \item Level 4 (highly automated/ mind off): The unmanned driving system completes all driving operations. According to system requests, human drivers do not necessarily need to respond to all system requests and limit road and environmental conditions.
    \item Level 5 (fully automated): The unmanned driving system completes all driving operations. Human drivers take over when possible. Drive on all roads and environmental conditions.
\end{itemize}

One of the biggest questions surrounding AVs is: How safe is safe enough? This is a controversial issue. Statistics\cite{mueller2020humanlike} show that AVs are very safe, and AV-caused crashes are rare. It is still hard to get an accurate self-driving car death toll since some AVs are still developing or testing. Tesla claims that their ADS processing time is four times safer than regular cars, while operating in Autopilot mode. Their estimation is one fatality per 320 million miles driven. However, people still have doubts about the safety performance of autonomous vehicles. As reported in~\cite{UberAccident}, a self-driving Uber car (a test vehicle) hit and killed a woman in Arizona partly because it failed to recognize the pedestrian jaywalk, but further investigations found the pedestrian at fault, and Tempe Police called the crash unavoidable since  the pedestrian had been crossing outside of a crosswalk \cite{UberAccident2}.  
Also, there is a cultural barrier to the broad utilization of AVs, especially in developing countries. To solve these challenges, characterizing safety metrics and their thresholds for AVs is of great importance. These analyses, along with the low crash rate of AVs, can provide further relief to society for using AVs. Most AVs calculate safety metrics and use AI methods to react to safety issues appropriately. However, they see the events and the traffic from their own perspective. Developing network-level safety metrics can provide a holistic assessment of the overall traffic safety level when AVs join the regular traffic flows.


\subsection{Relation to Vehicle Insurance Evaluation}
The primary use of auto insurance is to provide financial protection against physical damage or bodily injury resulting from traffic collisions and against liability that could also arise from incidents in a vehicle. In general, the insurance company calculates the insurance premium based on different factors that affect the customer's chance of being involved in a crash. The factors include age, car type, driving history, where the customers live, and other factors. 
Therefore, developing safety models that integrate these factors and predict crash rates can be used by insurance companies for more accurate estimations \cite{singh2019automating}.

Recently, a fully autonomous vehicle insurance pricing~\cite{konrardy2019fully} system was built based on such information. Moreover, a system~\cite{bowne2019methods} that uses vehicle operation data collected via mobile devices for vehicle insurance pricing. Studies on traffic safety metrics could provide more precise and targeted information to these systems to determine a more flexible and reasonable insurance policy for both customers and companies.

\subsection{Crowd-sourcing for Traffic Analysis and Driving Safety}
Noting the wide use of smartphones with accurate positioning systems, crowd-sourcing can be used to collect vehicle motion trajectories, crash incidents, etc., for network-level safety analysis. For instance, \cite{pooja2017early} uses crowd-sourcing to provide an early warning to drivers regarding road safety hazards due to construction work, defective street cuts, bumps, etc., using a cellphone-based App and embedded accelerometer readings. Indeed, part of the navigation software features in GoogleMAP and Waze is based on crowd-sourcing\cite{GoogleCrowdsourced_2013}. 
However, most crowd-sourcing methods share raw data; therefore,  using more elegant DL-enabled safety analysis algorithms can substantially enhance the efficacy of the produces advisory messages.    

\subsection{Vehicular Edge Computing} \label{sec:vec}

When the computation load is beyond the local servers' power, cloud computing is used. However, cloud computing may cause intolerable computation delays and interruptions due to networking delays. For these scenarios, edge emerging is adopted by running computations on servers located at the network edge, to mitigated networking and scheduling delays \cite{khan2019edge}. For instance, the idea of offloading heavy computations tasks by AVs to RSUs is proposed in \cite{liu2019edge}. 
These distributed nodes can share their computation ability that decentralizes the stress of the large network and reduces bandwidth consumption and response time among various servers and end-users. Users are allowed to access the physically closest servers to operate their real-time applications with good Quality of Service (QoS) and low latency\cite{yu2017survey}.

\begin{figure*}[t]
\begin{center}
\centerline{\includegraphics[width=0.7\textwidth]{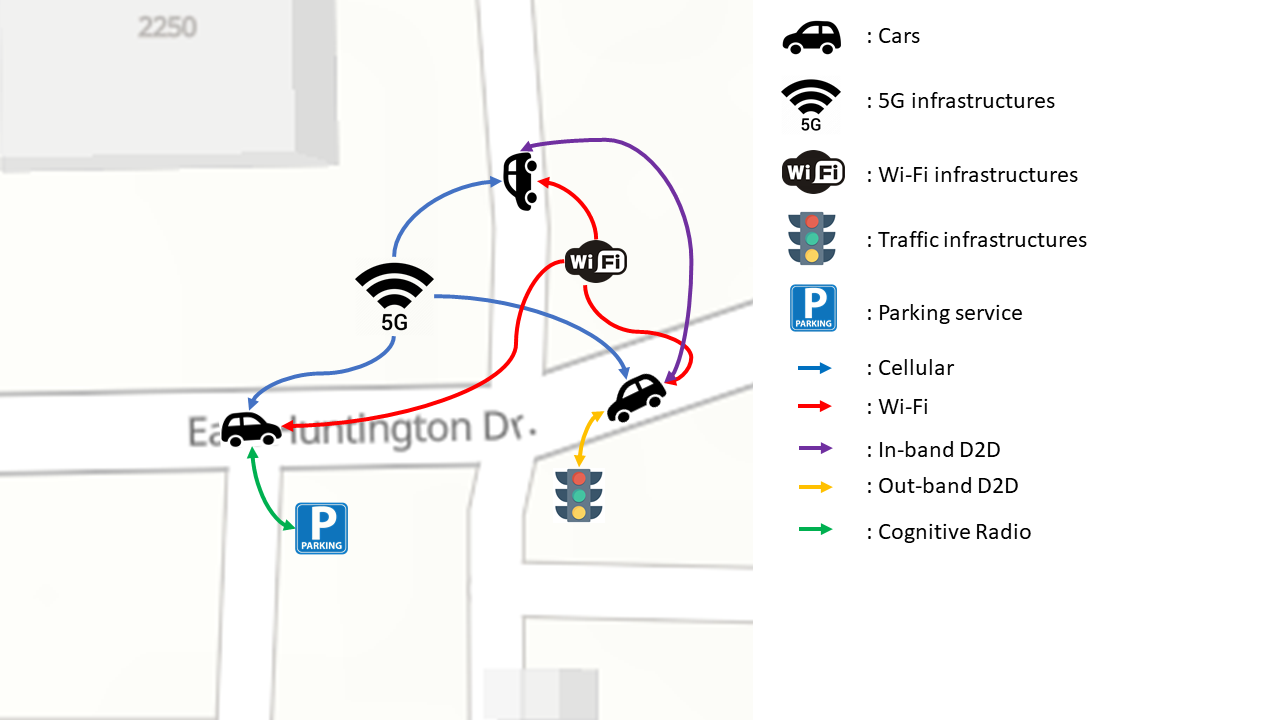}}
\caption{VANETs framework.}
\label{fig:VANETs}
\end{center}
\end{figure*}

Recently, the idea of deploying Vehicular Edge Computing (VEC) in Vehicular Ad Hoc Networks (VANETs) is proposed (Fig. \ref{fig:VANETs}). Conventionally, VEC consists of three layers \cite{raza2019survey}: User Layer, Edge layers, and Cloud Layer. The User Layer is composed of Vehicular Terminals (VTs), mainly as smart vehicles. Terminals perform sensing the environment, Vehicle-to-Vehicle (V2V) and Vehicle-to-Infrastructure (V2I) communications, limited computation tasks, and part of the storage. Edge Layer mainly consisted of Road-Side Units (RSUs), handle caching, computation, offloading, and deliver low-latency diverse service. RSUs are widely distributed on the roads and have more powerful computation capacity and larger storage space. RSUs deploy wireless communication protocols (such as WLAN, 3GPP, 4G, 5G, etc.) to guarantee reliable links that connect the Cloud and User Layers. It should be noted that the majority of data computation and storage is carried out in this layer. Cloud Layer, which consisted of cloud servers, has the greatest computation capacity and the higher storage capacity. This layer is desired to support centralized control, data aggregation, and global management and optimization. These tasks often are complicated and time-consuming; however, latency-sensitive tasks should not be included. To achieve enhanced performance, VECs use several techniques, including Content caching, redundancy (the content the terminal potentially requests), and history content on the edge servers in the proximity of users. This reduces the traffic flow across the whole network and enhances the user experience\cite{hu2017roadside,su2018edge,mahmood2016mobility}. In order to solve computation issues, task Offloading is used to transfer the excessive computations from local servers (potentially in vehicles computers) to the closest RSUs \cite{liu2019edge,liu2018computation,du2018computation}. Software-Defined Networking (SDN)\cite{deng2017latency} separates the control plane and data plane that allows easy switches reconfiguration for flexible network management. 

VEC addresses a wide range of problems for traffic safety. It facilitates real-time collecting and processing data from smart vehicles and infrastructures to optimize navigation to avoid congestion and warn the surrounded vehicles to avoid collisions or other incidents. Also, each participating VEC receives the information and is allowed to adjust its strategy (such as Platoon \cite{jia2015survey}). Moreover, beyond safety-related scenarios \cite{liu2020vehicular}, VEC also make high traffic demand applications possible, such as video streaming\cite{chen2019study,grassi2017parkmaster}, Augmented Reality (AR)\cite{park2013vehicle}, and in-vehicle Infotainment Service (such as online gaming)\cite{cheng2011infotainment}. 
The current challenges and active research problems in VEC include transmission reliability, service capacity, integration, scalability, security and user privacy, 
and economic considerations.









\section{\ar{A Roadmap of Traffic Safety Development in Vehicle Industry}}
\label{sec:roadmap}

\begin{figure*}[hb]
\begin{center}
\centerline{\includegraphics[width=1\textwidth]{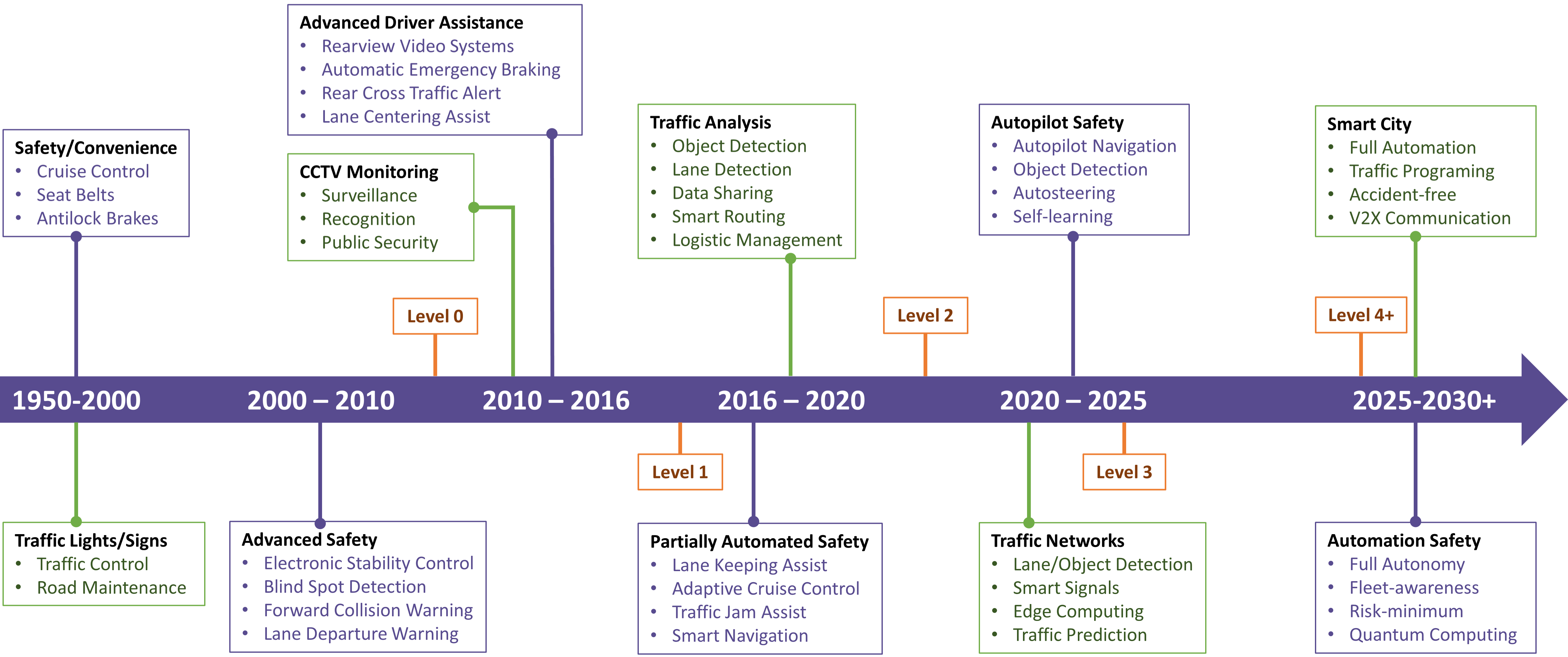}}
\caption{\ar{A roadmap for traffic safety development with vehicle automation level.}}
\label{fig:roadmap}
\end{center}
\end{figure*}

\ar{Since 2010, conventional safety technologies such as \textit{Automatic Emergency Braking} (AEB) and \textit{Electronic Stability Program} (ESP) have become prevalent for most private and commercial cars \cite{hulshof2013autonomous}. Starting in 2016, more vehicle manufacturers such as Toyota, Honda, GM, and BMW have equipped advanced driver-assistance systems, like \textit{Lane Keeping Assist} (LKA) and \textit{Adaptive Cruise Control} (ACC), for more standard models to improve the safety of vehicles \cite{eichelberger2016toyota}. These subsystems can achieve basic steering, acceleration/deceleration, and lane changing under specific circumstances. According to \cite{ostling2019passenger}, these advanced driver assistance technologies have reduced the accident rate by about 25 \% on average. 
More importantly, the single-vehicle and head-on collisions have already been reduced by about 50 \% \cite{sternlund2017effectiveness}, meaning that the future focus of safety features would be on the turning and crossing scenarios. In such scenarios, more advanced driver assistance systems are required to handle the active safety of drivers and passengers with complex external interactions.}

\ar{In 2019, Tesla announced its \textit{Full Self-Driving} (FSD) service on private transportation \cite{tesla2019}. This is the first case of SAE level 3 automation on the mass production level. 
Soon afterward, Google and NVIDIA presented their level 4 autonomous vehicles \cite{waymo2020,nvidia2022jidu,nvidia2022nio}, while GM and Amazon announced their level 5 autonomous shuttles later in 2020 \cite{gm2022origin,zoox2020}. The above companies all presented their mature solutions for self-driving in both private and commercial transportation. Tesla and Waymo also introduced the self-driving semi-truck for commercial and industrial utilization \cite{tesla2022semi,waymo2022via}.}

\ar{Vehicle automation mostly benefited from the vast experience of vehicle production in the last decades, while smart traffic network management systems are still in their infancy. Part of this lag is due to the need for heavy investment in networking, the internet, and road infrastructures. The present traffic network systems utilize features such as Closed-Circuit Television (CCTV)-based traffic monitoring, incident detection, and traffic flow analysis based on GPS information and real-time data analysis \cite{flir2020incident,trafficvision2016,notraffic2020,smartmicro2019}. These safety features have not been widely utilized in suburban areas and require further investment and appreciation from the local departments of transportation \cite{suburban2022}.}

\ar{We conclude this section by presenting key milestones and the future roadmap of the vehicle-based and network-level traffic safety features in Fig. \ref{fig:roadmap}.}

\section{Key Challenges and Open Problems} \label{sec:challenges}

\ar{We reviewed the applications of DL in video-based traffic safety analysis along with the envisioned future directions in Section \ref{sec:new-trends}.} 
Although the use of DL methods for different aspects of driving safety analysis gains more momentum every year, there still exist numerous challenges and issues to be addressed.

\subsection{\ar{Data Domain Drift}}

\ar{Domain data drift is one of the most practical problems when applying deep learning methods in traffic analysis, especially in some custom tasks (i.e., tasks that have specific requirements and/or when proprietary data is utilized).}

\ar{The proposed network often are trained and evaluated on public popular datasets (such as datasets in Table \ref{tab:trafficdataset} and Table \ref{tab:trafficdataset-continue}). Such datasets are usually neat, distortion-free, and well-annotated. This is not the case for most locally collected datasets. Therefore, training DL networks from scratch or utilizing networks that are pre-trained on benchmark datasets for new tasks can yield disappointingly poor performances. 
This performance shift is frequently reported due to the proprietary datasets' domain shift, such as different resolutions, noise levels, altered perspectives and field of view, etc. We experienced the same issue in our network-level metrics analysis project \cite{chen2022network}. In this work, we aimed to extract network-level metrics from surveillance videos. \ar{As part of this problem, we need to detect and trace vehicles; however, using pre-trained benchmark detection algorithms, such as FCOS and RetinaNet, yield poor performance.}}
\ar{Finally, we ended up using the combination of YOLOv5 and DeepSort for tracking vehicles on the road, which gave 
accepted results. This is not specific to our work and often occurs when applying a DL framework to new problems, especially for practical systems.   
Further investigations of more elegantly designed transfer learning approaches or formal ways of fine-tuning pre-trained networks can be a game-changer. } \ar{Another possibility is using meta-learning, which is briefly discussed in Section \ref{sec:meta}, to learn how to learn efficiently and optimally. This may decrease the cost of annotation and allow the fast deployment of DL frameworks in this context. }

\subsection{\ar{Data Processing Labor Cost}} 
 	
Labor cost is another limiting factor for developing learning-based traffic modeling frameworks, \ar{especially for data preparation and annotation.} For instance, the 100-Car naturalistic driving study (2006)\cite{dingus2006100}, one of the most popular traffic datasets, collected data from about 100 cars totally driving approximately 2,000,000 miles and 43,000 hours, and took about four years to complete. We agree that not every problem needs this huge of a dataset; however, the traffic-related work needs this kind of data to create reliable automated analysis frameworks. \ar{Usually, annotation requires a big team of experts to perform annotations and tackle the difference of opinion. This can impose unaffordable costs and time delays for budget-constrained projects. Alternative solutions are using semi-supervised learning to label data, unsupervised learning (such as unsupervised spatiotemporal representation learning discussed in Section \ref{sec:unsuperST})}), and data augmentation methods to mitigate the need for massive annotated datasets.

\subsection{Modeling Complexity} 
Developing data-driven and mathematical frameworks to model traffic flow and safety risks remains a challenging issue. Part of the reasons is the difficulty of modeling the environment, a huge number of factors with interlaced roles, the impact of human factors and cognition, and the relations between different vehicles, which creates a complex system to model. 
Several studies tried to model complicated traffic conditions. Some studies including  \cite{st2013automated,yan2008validating,mullakkal2020probabilistic,chen2017surrogate,saunier2014road} applied conventional methods to create surrogate safety models, while other works such as  \cite{hao2013vehicle,shahdah2014integrated,hoogendoorn2013longitudinal,zheng2014freeway,kuang2015tree,huang2014evaluating,kim2013identifying,yan2008validating,wu2014using} deployed statistical models to analyze traffic data. More recent works use DL for modeling purposes. For instance, DL frameworks are utilized by \cite{lv2014traffic,zhao2017lstm,polson2017deep,yu2017spatio,wu2018hybrid} for traffic prediction, by \cite{mozaffari2019deep} for vehicle behaviour prediction, and by \cite{duan2016efficient,wang2015applications,behrendt2017deep} for traffic classification. Although these works achieve excellent performance, they only modeled some specific scenarios (e.g., intersection, ramp merge, etc.). A general model that can briefly represent real road traffic conditions, perhaps by integrating the existing models, is still considered an open research problem.

\subsection{Algorithm Reliability and Efficiency:} 
Although DL methods have shown superior performance in simple image-based tasks such as object recognition, object tracking, and instance segmentation, they can be prohibitively unreliable when it comes to modeling multi-factor and multi-faceted phenomena that involve extracting complicated tasks by processing videos in real-time. 

It is known that many DL-based algorithms deployed by traffic systems and AVs, such as \cite{girshick2014rich,liu2016ssd} for object detection, \cite{vzbontar2016stereo,luo2016efficient} for stereo matching, \cite{FlowNet,10.1109/cvpr.2017.291} for optical flow. With recent advances in high-computational processing platforms, this issue seems to be mitigated day by day. More and more studies provide evidence for AVs' safety and accuracy of DL-based traffic monitoring systems to alleviate cultural barriers in using DL-powered technology and replacing humans with computers. However, still, some manufacturers like Tesla emphasize their products still require active driver supervision\cite{teslaWeb}. 

For conventional traffic analysis, the use of DL-based algorithms is not critical and does not directly compromise safety. However, the more widespread use of these algorithms can provide a better understanding of traffic safety in general. It can improve traffic risks by offering design hints to transportation infrastructures and real-time warnings to the cars on the road.


\subsection{\ar{Model Explainability}}

\ar{There exists a known performance-explainability trade-off in developing DL frameworks. One may explore the association among features from traffic scenarios (such as the association between crash rate and traffic volume) by statistic learning or simple models (such as linear regression or logistic regression). These methods are generally easy to explain and interpret. On the other hand, data-driven DL methods often show higher capabilities in extracting useful information for massive data; however, they are viewed as \textit{black boxes} with limited \textit{explainability}.} \ar{This hinders extracting interpretable results, conceptual interpretations, and useful design guidelines from the developed models.}

\ar{Currently, developing explainable AI platforms is considered an emerging field where efforts are made to make DL frameworks more explainable and easy to comprehend. For instance, visualizing and translating latent features into more meaningful representations in justification and dialogue system \cite{vassiliades2021argumentation} is gaining higher momentum. Alternatively, one may intend to explain the local parts of the model. It can be addressed by using an interpretable model to imitate the behavior of an uninterpretable model. For example, there are some works that use local interpretable model-agnostic explanations (LIME) \cite{ribeiro2016should} to explain the selected data points in multiple applications, such as time series forecast \cite{schlegel2021ts}, medical imaging \cite{meske2020transparency}. We expect that developing explainable and interpretable models would be a potential paradigm in traffic safety analysis and autonomous driving.}

\subsection{Equipment Support:}
Scarcity of data and the lack of sufficient monitoring infrastructure is another drawback. For instance, in the state of Arizona, more than 441 public cameras 
are used by the ADOT to monitor traffic\cite{az511}, but there are a total of 144,959 miles\cite{fhwa}. This means that a lot of roads have not yet been fully covered by surveillance cameras. 

\subsection{Privacy and Secrecy}
Another barrier of common spread use of DL-algorithm for safety analysis is that traffic video contains personal information such as human face and plate numbers, which raises privacy concerns. We believe that publishing more traffic video repositories with removed Personal Identifiable Information (PII) can substantially accelerate the rate of discovery without compromising people's safety.

\subsection{\ar{Model/Data Safety and Adversarial Learning}}

\ar{Trained DL networks can be susceptible to adversarial attacks that try to disrupt the model's operation by injecting falsely annotated and misleading data points. This issue often refers to a topic known as adversarial learning (more detail, see Section \ref{sec:AML}).}

\ar{Imagine a few crafted adversarial examples that can totally mislead the model to make a wrong decision or introduce backdoors to the system \cite{chakraborty2018adversarial}. 
This issue is not easily imperceptible but significantly dangerous, especially for self-driving vehicles and RSU-based safety control systems. Developing high-standard defense strategies for traffic-related models is a top priority, and inverse learning can play an essential role in this respect.}

\subsection{Naturalistic Driving Data:}
Although some naturalistic driving data were collected for traffic flow and transportation research, such as Next Generation Simulation (NGSIM) data\cite{NGSIM}, open-source naturalistic driving data for safety evaluation of vehicles are rarely reported.  Since for vehicle safety analysis, some quantitative data of each vehicle are required, such as location, speed, acceleration, and heading angles, advanced technology need to be developed and applied to accurately obtain these measurements.  When cameras are applied, computer vision and related to objective identification and tracking algorithms, based on ML or DL, need to be specifically developed, especially towards real-time processing.  Some recent available naturalistic driving data obtained by drones, such as LevelX \cite{krajewski2018highd,bock2019ind}, can provide necessary pre-processed data for safety analysis purposes.  However, these data are processed offline and expensive.

\subsection{\ar{Conflict of Responsibility}} \label{sec:conflict}

\ar{Compared to conventional driver assistance technologies, companies such as Tesla, NVIDIA, and Google prefer to use neural network-based models to handle most of the driving decisions in complex environments.
As mentioned earlier, Tesla has announced the first Full Self-Driving (FSD) service on private transportation in 2019 \cite{tesla2019}. The system includes traffic-aware cruise control, autosteer, auto lane change, autopark, traffic sign control, and navigate on autopilot (Beta). Based on these features, the later Tesla products can be considered SAE level 3 vehicles. 
Almost the same time,  Waymo has launched an SAE-4 taxi service in Phoenix, Arizona \cite{waymo2020}, as one of the first driver-less transportation services in the country. Also, Intel's mobile eye \cite{mobileeye,mobileye2020adas}, and Nvidia's Drive \cite{omniverse2022,nvidia2022selfdrive} systems are other examples of SAE level 4 AVs. 
The GM Cruise Origin \cite{gm2022origin}, and Amazon Zoox \cite{zoox2020}, presented almost fully-autonomous buses, toward developing SAE level 5 AVs.
However, due to the conflict of responsibility between manufacturers and customers, some corporations abstain from declaring higher levels of autonomy. Some suppliers call their service SAE-2 \cite{tesla2021level2}, since the SAE-3 and higher levels of autonomy would require more strict regulations and would incur higher levels of responsibility to the manufacturer when a crash occurs \cite{SAE2014}. 
This might cause inconsistency and interoperability issues between the connected AVs and smart traffic systems in future developments.}

\subsection{\ar{Fusion of AVs and the Future Traffic Network}}

\ar{Despite the rapid growth of both autonomous driving and smart traffic networks, very few works discuss the interoperability and integration of such systems. A recent book \cite{ersoy2020autonomous} reviews the deployment of autonomous shuttles and self-driving technologies in various scenarios. \cite{ranka2020vision} on the other hand, only discussed the applications of smart traffic techniques on AVs. 
A key question of "how to connect self-driving vehicles to the smart traffic network?" remains largely unanswered. The present smart traffic networks mostly rely on active environment sensing, then pass the information to fixed terminals (mobile device, APP) \cite{smartmicro2019}. These features are usually not utilized by current vehicle models, nor equipped by older models. A lawful document is expected to push these safety features to wider use.}

\ar{Overall, developing efficient and sustainable AI-based traffic systems requires universal standards and guidelines through closer cooperation between different entities, including researchers, technology developers, law enforcement authorities, and operational teams}. 
\ar{Besides, the communication between AVs from different manufacturers is limited. The dataset collected by corporations is labor costly, and privacy-sensitive \cite{bloom2017self}. Apparently, large vendors are hesitant to share their datasets to keep their leverage in the competing AV market. This causes difficulty for researchers to obtain useful datasets and may lead to safety concerns \cite{banks2018driver}.}
\ar{Implementing stronger intensives for data sharing, and standardizing the development process can further power this line of research.}


\section{Conclusion}
This paper reviewed DL methods that can be used for different aspects of video-based traffic safety analysis. We reviewed methods, tools, and datasets that are recently developed by the research community and industry. We highlighted key achievements and mentioned areas that need further investigation. For example, we enumerate areas that require more advanced tools and also collecting well-annotated datasets. Some examples include but not limited to the need for developing DL algorithms, tools, and datasets for aerial traffic monitoring systems, more advanced video-based action recognition systems, integrative analysis of multi-modal traffic image and sensor data, extending individual safety metrics into network-level safety metrics, formal ways to develop safety metric distributions, finding associations between network-level safety metrics and crash rate, developing online safety metric extraction tools, and developing end-to-end frameworks to translate safety risks into traffic advisory messages. We also made connections to closely related research areas, including AVs, Crowd-sourcing for traffic analysis, and driver's behavioral patterns and psychological profiles, and the insurance industry. Our hope is that this paper will help computer scientists solve traffic safety problems, particularly areas that need further investigation. 
This paper also aimed to help traffic engineers and personnel to identify and use existing open-source tools for their problems.

\section{Declaration of Competing Interest}
The authors declare that they have no known competing financial interests or personal relationships that could have appeared to influence the work reported in this paper.

\section{Acknowledgment}
We would like to thank Drs. 
Junsuo Qu and Greg Leeming for his insightful comments.
We are grateful to Arizona Commerce Authority, Intel Corporation, State Farm Insurance, Arizona Department of Transportation, and the Institute for Automated Mobility (IAM) for their continued support of this project. 

\bibliographystyle{vancouver} 
\bibliography{references}

\end{document}